\documentclass[journal]{IEEEtran}
\usepackage{wrapfig}
\usepackage{xparse}
\usepackage{xkeyval}
\usepackage{morewrites}
\usepackage{cite}
\usepackage{amsmath,amssymb}
\usepackage{algorithmic}
\usepackage{array}
\usepackage{graphicx}
\usepackage{stfloats}
\usepackage{url}

\usepackage[T1]{fontenc}

\usepackage[colorlinks,linkcolor=black,filecolor=black,urlcolor=black,citecolor=black]{hyperref}
\usepackage{balance}
\usepackage{subcaption}
\usepackage{booktabs}
\usepackage[table,dvipsnames,svgnames,x11names]{xcolor}
\usepackage{caption}    
\usepackage{color}
\definecolor{lightblue}{rgb}{0.8,0.9,1}
\definecolor{lightorange}{rgb}{1,0.9,0.7}
\definecolor{lightgray}{rgb}{0.9,0.9,0.9}
\newwrite\authorbibfile
\AtBeginDocument{%
  \immediate\openout\authorbibfile=\jobname.aub%
}%
\AtEndDocument{%
\immediate\closeout\authorbibfile
\InputIfFileExists{\jobname.aub}{}{}
}%

\makeatletter
\define@key{authorbib}{scale}[1]{%
\def\AuthorbibKVMacroScale{#1}%
}
\define@key{authorbib}{wraplines}[10]{%
\def\AuthorbibKVMacroWraplines{#1}%
}
\define@key{authorbib}{imagewidth}[4cm]{%
\def\AuthorbibKVMacroImagewidth{#1}%
}
\define@key{authorbib}{overhang}[10pt]{%
\def\AuthorbibKVMacroOverhang{#1}%
}
\define@key{authorbib}{imagepos}[l]{%
\def\AuthorbibKVMacroImagepos{#1}%
}
\makeatother

\presetkeys{authorbib}{imagepos=l, imagewidth=4cm, wraplines=8, overhang=20pt}{}
\newlength{\AuthorbibTopSkip}
\newlength{\AuthorbibBottomSkip}
\NewDocumentCommand{\authorbibliography}{+o+m+m+m}{%
  \IfNoValueTF{#1}{%
  }{%
    \setkeys{authorbib}{#1}%
    \immediate\write\authorbibfile{%
      \string\begin{wrapfigure}[\AuthorbibKVMacroWraplines]{\AuthorbibKVMacroImagepos}[\AuthorbibKVMacroOverhang]{\AuthorbibKVMacroImagewidth}^^J
        \string\includegraphics[scale=\AuthorbibKVMacroScale]{#2}^^J
        \string\end{wrapfigure}^^J
    }%
  }%
  \IfNoValueTF{#3}{%
    \typeout{Warning: No author name}%
  }{%
      \immediate\write\authorbibfile{%
      \unexpanded{\vspace{\AuthorbibTopSkip}}^^J
      \string\noindent\relax
      \unexpanded{\textbf{#3}\par}^^J
      \string\noindent\relax
      \unexpanded{#4}^^J%
      \unexpanded{\vspace{\AuthorbibBottomSkip}}^^J
      }%
  }%
}%
\begin{document}
\title{Balancing Emotional Alignment and Semantic Consistency in Image Generation via Reinforcement Learning with Valence-Arousal Anchoring}

\author{Jisheng Dang, Zhenxuan Wang, Bin Li, Ronghao Lin, 
 Bin Hu, ~\IEEEmembership{Fellow, ~IEEE},  Tat-Seng Chua
  
\thanks{This work was supported by the National Natural Science Foundation of China (Grants No. 62227807 and U24B20186). This work was also supported by the Supercomputing Center of Lanzhou University.}
\thanks{Jisheng Dang and Zhenxuan Wang are with School of Information Science \& Engineering, Lanzhou University, Lanzhou 730000, China.  (E-mail: dangjsh@mail2.sysu.edu.cn).}

\thanks{Bin Li is with School of Environmental and Spatial Informatics, China University of Mining and Technology, Xuzhou 221116, China E-mail: tb24160013a41@cumt.edu.cn}

\thanks{Ronghao Lin is with the College of Computer Science and Software Engineering, Shenzhen University, Shenzhen 518060, China (E-mail: linrh@szu.edu.cn)}

\thanks{Bin Hu is with the School of Medical Technology, Beijing Institute of Technology, Beijing 100081, China (E-mail: bh@bit.edu.cn). }

\thanks{Tat-Seng Chua is with the School of Computing, National University of Singapore, Singapore 119077 (E-mail: dcscts@nus.edu.sg). }
\thanks{*Corresponding author: Ronghao Lin and Bin Hu.}}

\maketitle

\begin{abstract}
Continuous emotion control in text-to-image generation requires a model to improve affective alignment without changing the objects, layout, or scene described by the prompt. Existing supervised emotion-injection methods often optimize feature-space proxies and may therefore exhibit \emph{emotion--semantic drift}, in which stronger emotional conditioning is accompanied by unintended content changes. We address this problem with a flow-matching image-generation framework that combines continuous valence--arousal (VA) conditioning, Group Relative Policy Optimization (GRPO), and a neutral semantic anchor. The deterministic probability-flow ODE is converted into a marginal-preserving SDE, yielding non-degenerate transition densities for trajectory sampling and policy-ratio estimation. A frozen CLIP-based VA regressor supplies a terminal reward measuring the distance between the predicted and target VA coordinates, while an image generated from the same prompt under zero VA conditioning provides a feature-space reference for semantic preservation. A reduced denoising schedule is used for online RL sampling, whereas the original schedule is retained at inference. Experiments on 3,300 prompt--emotion combinations show substantially lower valence and arousal errors than the VA-conditioned baseline and an improved CLIPScore relative to EmotiCrafter, with a measurable trade-off in reference-free image quality. The results support anchor-regularized Flow-GRPO as a practical approach to balancing emotional alignment and semantic consistency in continuous-affect image synthesis. Our code is available at \url{https://github.com/ramon-alana/eit-with-anchor-and-grpo}.
\end{abstract}


\section{Introduction}\label{sec:introduction}
Emotion-aware image generation aims to control not only what an image depicts, but also the affect that it conveys. This capability is useful in affective computing, human--computer interaction, personalized content creation, and creative or therapeutic applications~\cite{lee1999the,li2020emotional,zhang2024refashinoning}. Meanwhile, modern text-to-image models~\cite{rombach2022highresolution,esser2024sdxl,podell2023sdxl} provide strong semantic priors for generating complex scenes from natural-language prompts. Combining these two forms of control raises a central question that how can a generator change the perceived emotion of an image without changing the content requested by the user?

This question is nontrivial because visual emotion is not an independent style attribute. Perceived affect arises from the interaction of global appearance cues, such as color, illumination, and atmosphere, with semantic cues, such as objects, actions, and scene context~\cite{zhao2022affective,yang2021stimuli,yang2021solver}. Some appearance cues can be adjusted while preserving the depicted event, whereas changing an object or action may alter both affect and meaning. A useful emotional generator must therefore distinguish affective appearance changes from prompt-inconsistent semantic changes.

The valence--arousal (VA) model~\cite{russell1980circumplex} represents emotion in a continuous two-dimensional space, where valence describes pleasantness and arousal describes activation. Compared with discrete emotion labels, VA coordinates support gradual changes and interpolation between affective states. EmotiCrafter~\cite{dang2025emoticrafter_iccv} introduces this continuous control into text-to-image generation by injecting VA values into textual features. However, the emotion condition and the prompt semantics remain coupled in the learned representation. As the VA condition becomes stronger, the generator may alter objects, scene layout, illumination, or event content instead of changing affect alone.

\begin{figure*}[t]
    \centering
    \begin{subfigure}[b]{0.19\textwidth}
        \centering
        \includegraphics[width=\textwidth]{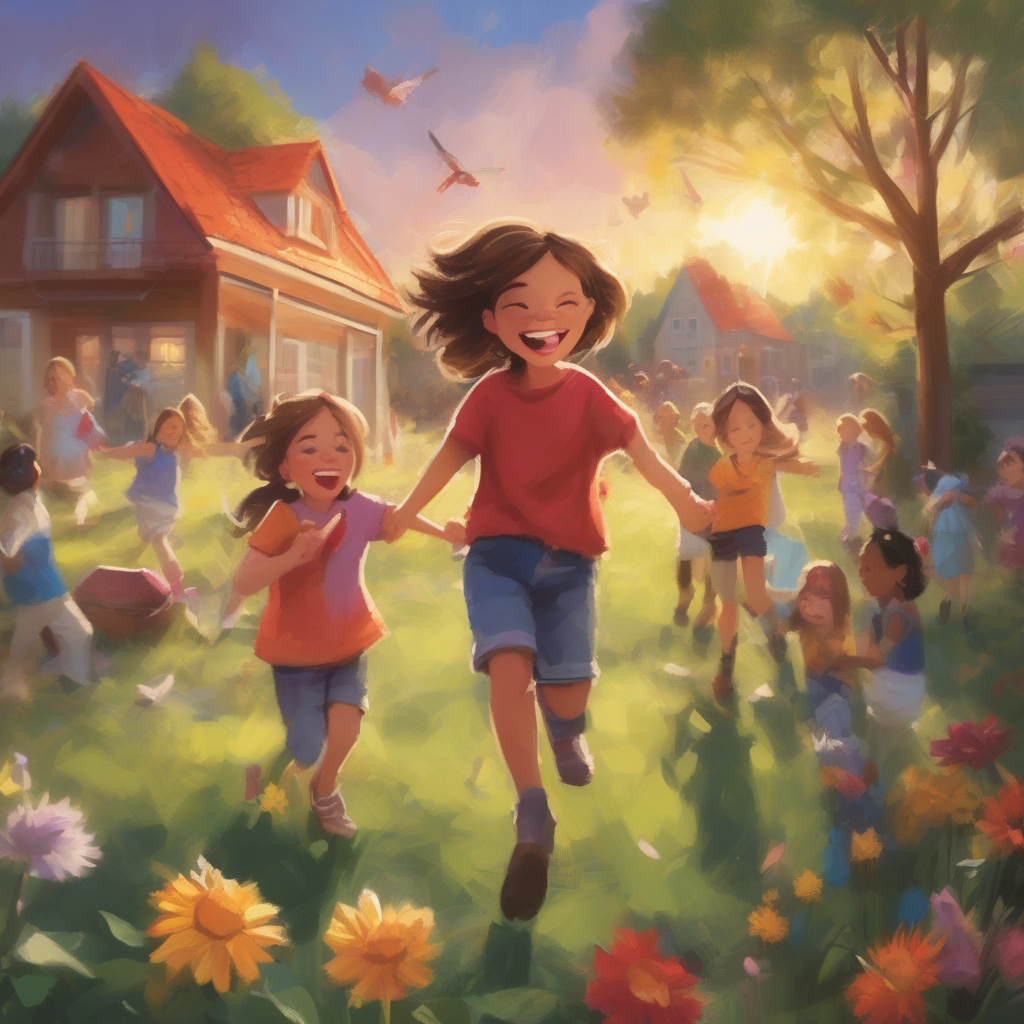}
        \caption{(3,3)}
        \label{fig:baseline_neutral}
    \end{subfigure}
    \hfill
    \begin{subfigure}[b]{0.19\textwidth}
        \centering
        \includegraphics[width=\textwidth]{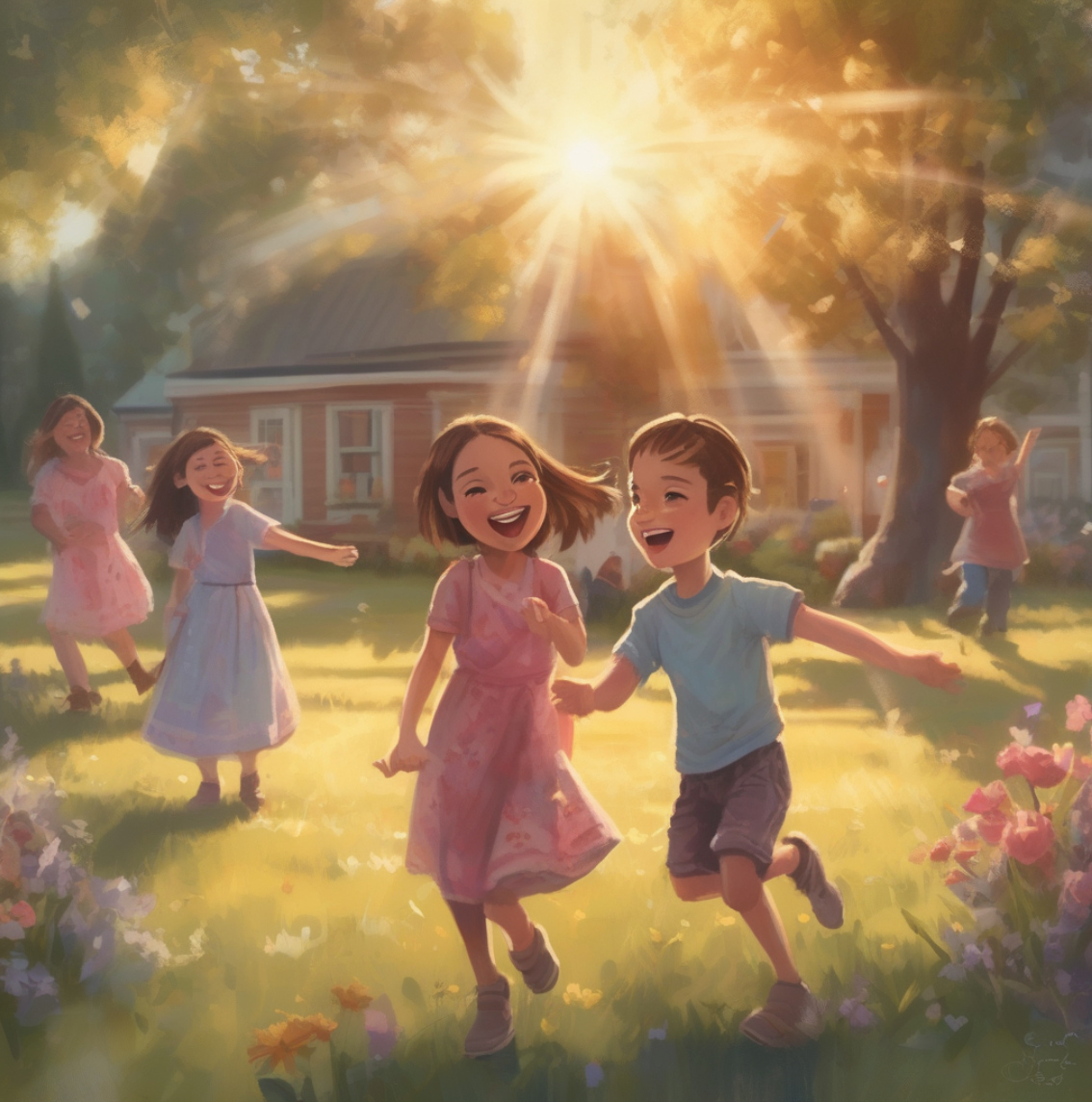}
        \caption{(1.5,1.5)}
        \label{fig:baseline_happy}
    \end{subfigure}
    \hfill
    \begin{subfigure}[b]{0.19\textwidth}
        \centering
        \includegraphics[width=\textwidth]{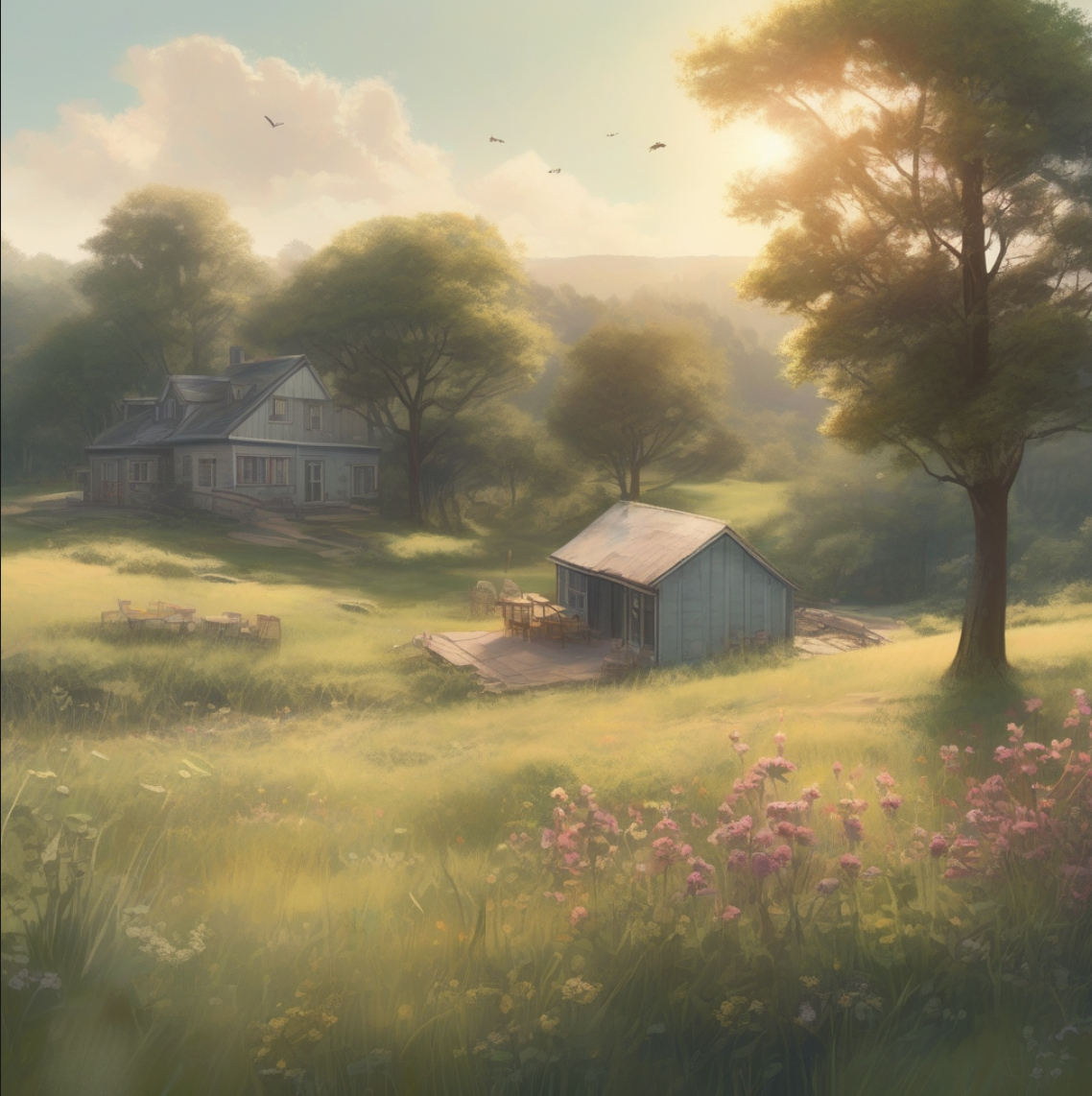}
        \caption{(0,0)}
        \label{fig:baseline_sad}
    \end{subfigure}
    \hfill
    \begin{subfigure}[b]{0.19\textwidth}
        \centering
        \includegraphics[width=\textwidth]{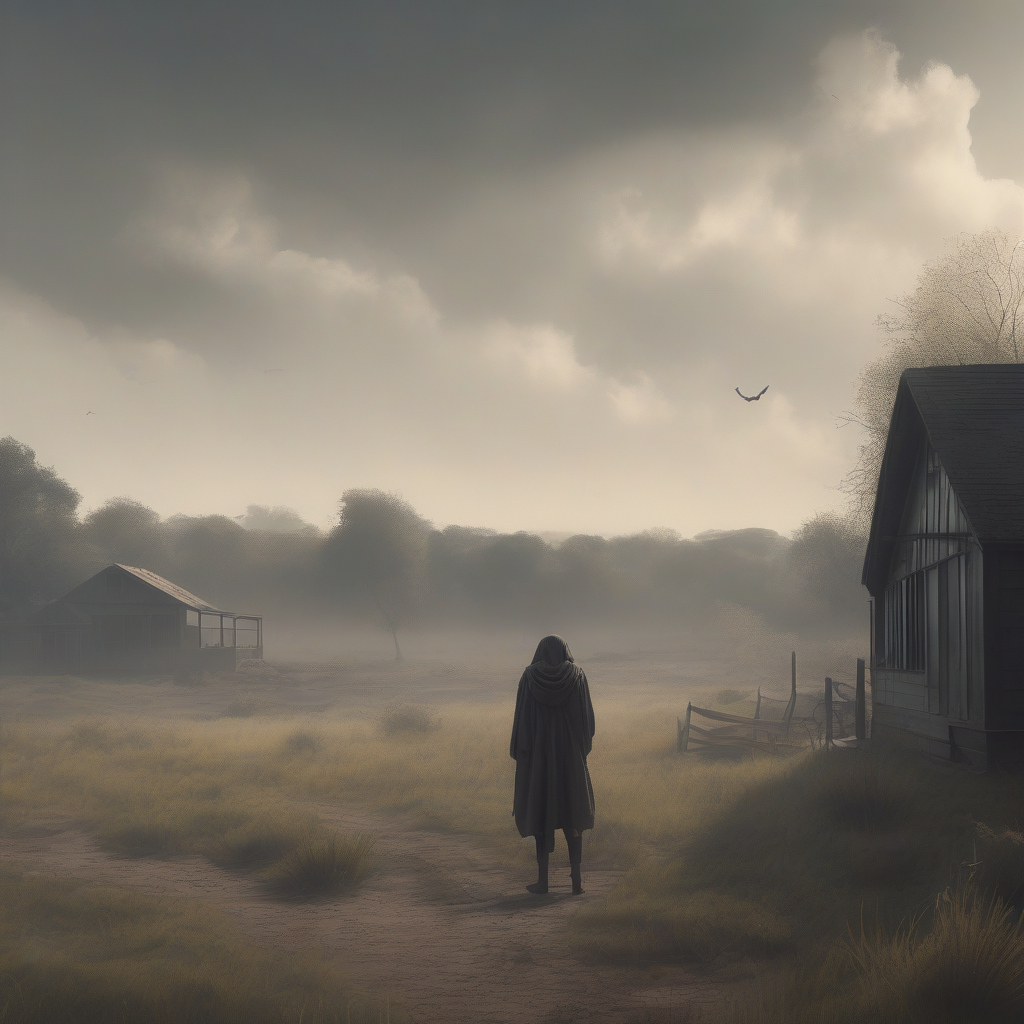}
        \caption{(-1.5,-1.5)}
        \label{fig:baseline_angry}
    \end{subfigure}
    \hfill
    \begin{subfigure}[b]{0.19\textwidth}
        \centering
        \includegraphics[width=\textwidth]{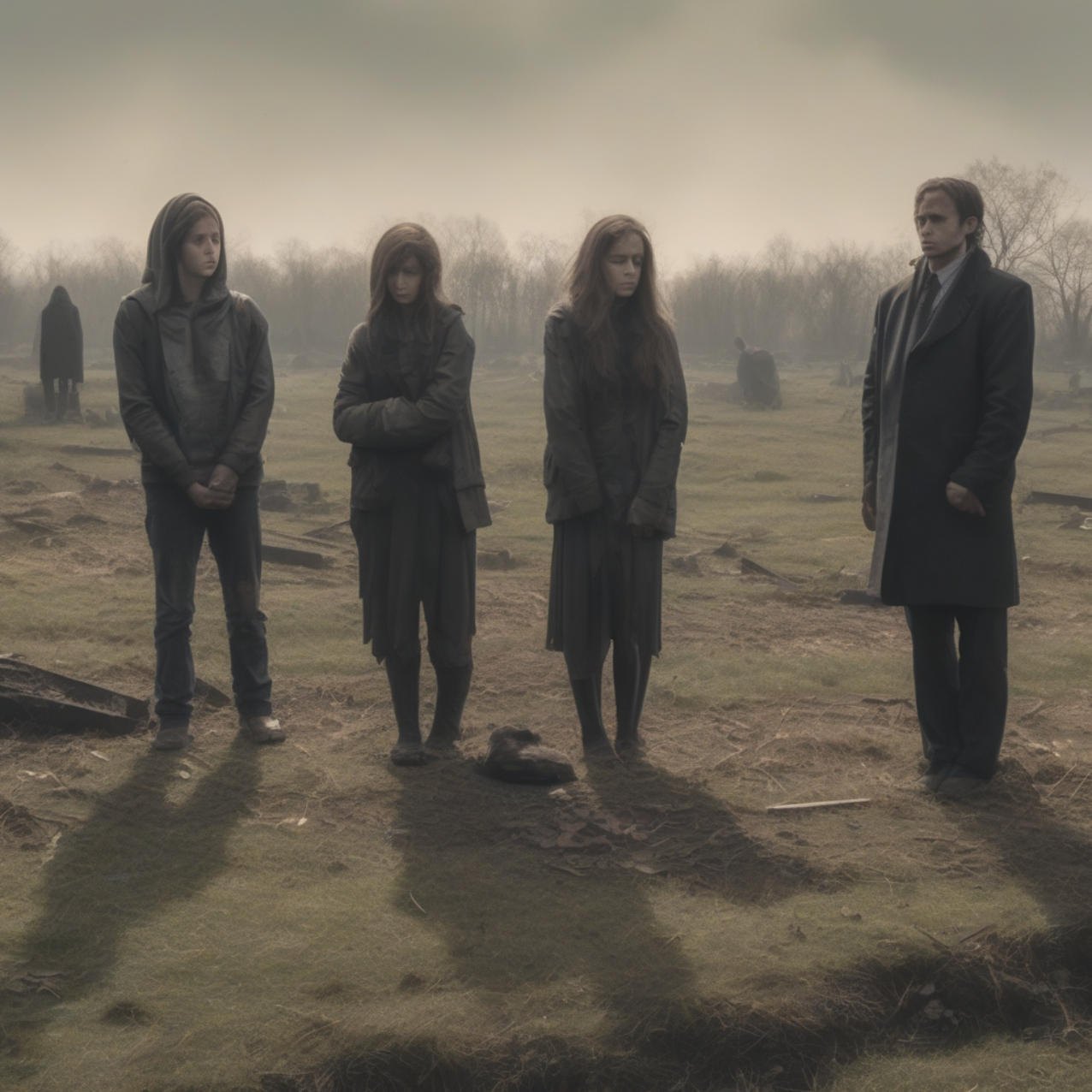}
        \caption{(-3,-3)}
        \label{fig:baseline_surprised}
    \end{subfigure}
    \vspace{1ex}  
    \begin{subfigure}[b]{0.19\textwidth}
        \centering
        \includegraphics[width=\textwidth]{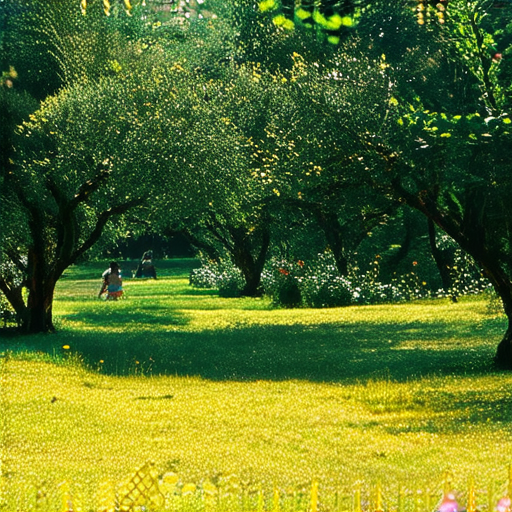}
        \caption{(3,3)}
        \label{fig:ours_neutral}
    \end{subfigure}
    \hfill
    \begin{subfigure}[b]{0.19\textwidth}
        \centering
        \includegraphics[width=\textwidth]{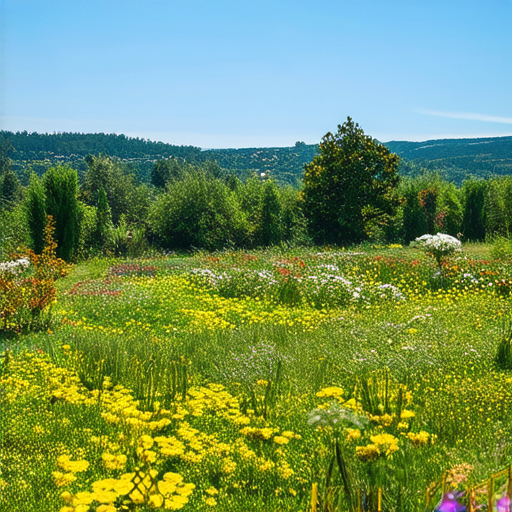}
        \caption{(1.5,1.5)}
        \label{fig:ours_happy}
    \end{subfigure}
    \hfill
    \begin{subfigure}[b]{0.19\textwidth}
        \centering
        \includegraphics[width=\textwidth]{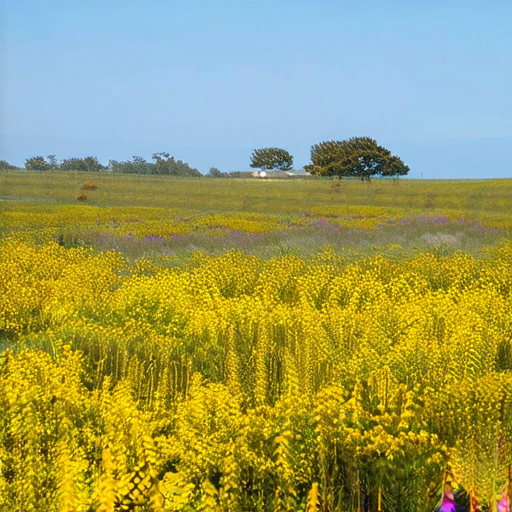}
        \caption{(0,0)}
        \label{fig:ours_sad}
    \end{subfigure}
    \hfill
    \begin{subfigure}[b]{0.19\textwidth}
        \centering
        \includegraphics[width=\textwidth]{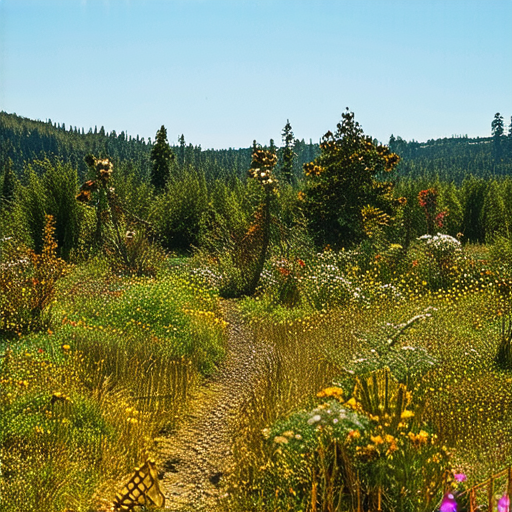}
        \caption{(-1.5,-1.5)}
        \label{fig:ours_angry}
    \end{subfigure}
    \hfill
    \begin{subfigure}[b]{0.19\textwidth}
        \centering
        \includegraphics[width=\textwidth]{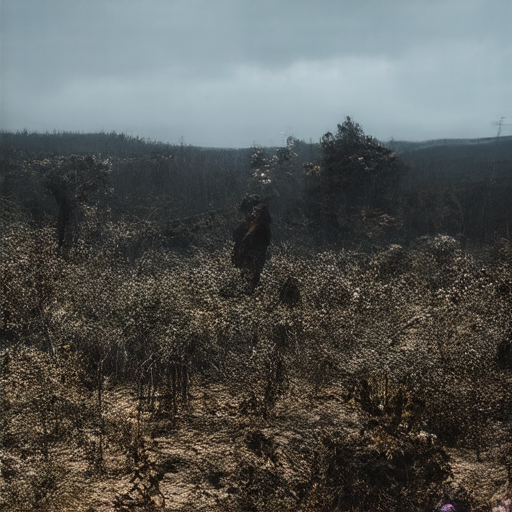}
        \caption{(-3,-3)}
        \label{fig:ours_surprised}
    \end{subfigure}
    \caption{Comparison of emotion-semantic drift across five Valence-Arousal (VA) values for the prompt ``a sunny meadow'', ordered from high (3,3) to low (-3,-3) as shown by the subcaptions. Top row (baseline): the model alters scene structure for non-neutral VA values. At (3,3), the scene becomes a festival. At (1.5,1.5), the composition becomes chaotic. At (0,0), the meadow is preserved. At (-1.5,-1.5), the sky darkens and trees become threatening. At (-3,-3), it turns into night. Bottom row (ours): our method preserves the meadow while modulating lighting and color appropriately for each VA coordinate.}
    \label{fig:drift_compare}
\end{figure*}

We refer to this failure mode as \emph{emotion--semantic drift}. Fig.~\ref{fig:drift_compare} provides an example: for the same prompt, the baseline changes the underlying scene across VA coordinates, whereas the desired behavior is to preserve the meadow and express affect through appearance-related attributes. The problem is therefore not merely to reduce emotion-prediction error; it is to improve continuous emotional alignment while limiting unintended semantic displacement.
Existing supervised emotion-conditioning methods mainly optimize intermediate feature regression, reconstruction, or contrastive objectives~\cite{zhao2022affective,dang2025emoticrafter_iccv}. Such losses do not directly optimize the emotion predicted from the final decoded image. Conversely, optimizing only an image-level emotion score is under-constrained: a generator can obtain a better score by changing the scene itself. These observations motivate two complementary objectives. The first should evaluate continuous emotional alignment on the final image; the second should explicitly discourage prompt-inconsistent changes.

We implement the first objective through reinforcement learning (RL), which can optimize a terminal signal without differentiating through the reward model~\cite{fan2023dpok,black2024training}. A frozen VA predictor measures the distance between the generated image and the target coordinate. Applying policy optimization to a flow-matching generator nevertheless requires care. Its standard sampler follows a deterministic ODE~\cite{lipman2023flow,liu2023rectified}; different initial noise samples provide output diversity, but the transition from a fixed intermediate state is a Dirac distribution and therefore has no ordinary density for the step-wise likelihood ratio used by GRPO. Following Flow-GRPO~\cite{flowgrpo2025}, we use a marginal-preserving SDE during training to obtain stochastic Gaussian transitions with tractable likelihoods.

The emotional reward alone does not address semantic drift. We pair each emotion-conditioned output with a zero-VA output from the same prompt and initial latent. The zero-VA image acts as a content reference, and their distance in a CLIP space forms an anchor regularizer. The complete training flow as follows: EIT module injects target VA coordinate into prompt representation; flow-matching generator samples an image through the stochastic training policy; VA predictor supplies terminal reward to Flow-GRPO; the neutral branch supplies semantic anchor. The generator is then updated with the negative GRPO objective and anchor loss jointly. This coupling between output-level emotional optimization and prompt-paired semantic regularization is the central methodological contribution of the work. The neutral anchor provides sample-specific reference rather than generic prior. Because branches share prompt and initial latent, feature difference is less affected by prompt variation or sampling-induced changes. Moreover, constraint is imposed in frozen visual space rather than pixel level, allowing emotional branch to modify color, lighting, and other cues. Anchor therefore is not intended to force identical images; it biases policy optimization away from semantic shortcuts and reduces overall semantic drift. 

The main contributions are summarized as follows:

\begin{itemize}
\item We formulate continuous emotion-conditioned image generation as joint output-level VA optimization and semantic regularization. This formulation makes emotion--semantic drift an explicit optimization concern rather than treating prompt preservation as an implicit consequence of emotional conditioning.

\item We combine a VA terminal reward with a prompt-paired neutral anchor. Flow-GRPO optimizes the decoded image against a frozen VA predictor, while the anchor penalizes the CLIP-feature displacement between emotion-conditioned and zero-VA outputs generated from the same prompt and initial latent.

\item We provide a parameter-efficient realization on a flow-matching backbone using LoRA and denoising reduction, together with experiments that separately examine the VA reward, anchor term, and backbone. Experiment results show reduced VA error and improved CLIPScore relative to the VA-conditioned baseline, while also exposing a remaining perceptual-quality trade-off.
\end{itemize}

\section{Related Work}

\subsection{Continuous Emotion-Conditioned Generation}

Early emotion-conditioned generators commonly use categorical labels such as happiness, sadness, and anger. EmoGen maps discrete emotion labels to visual content descriptors for domain-free emotional image generation~\cite{yang2024emogen}. Its control variable is therefore drawn from a finite set of categories, whereas our method accepts an arbitrary numerical coordinate in the continuous VA space together with a free-form prompt. Although category-based control is intuitive, it does not directly capture gradual affective transitions or distinguish nearby affective states within the same category.

Dimensional affect models instead represent emotion in a continuous space. The circumplex model characterizes affective states using valence and arousal~\cite{russell1980circumplex}. EmotiCrafter injects a target VA coordinate into the text features of a pretrained text-to-image model, enabling a free-form prompt and continuous emotion values to be specified jointly~\cite{dang2025emoticrafter_iccv}. Visual-emotion analysis has progressed from stimuli-aware feature extraction and scene--object relational reasoning to subjective emotion-distribution learning and cross-modal semantic-guided pretraining~\cite{yang2021stimuli,yang2021solver,yang2022subjectivity,chen2025uniemox}. Recent studies have further enhanced visual-emotion representation through hierarchical feature interaction~\cite{uniemo2025}, structured emotional annotations~\cite{guo2025emoverse}, and multimodal VA recognition~\cite{maven2025}. These studies primarily improve affect understanding and representation, whereas our work focuses on using continuous VA signals to guide image generation while mitigating semantic drift. In particular, they do not directly optimize the VA error of the final image together with a paired constraint that preserves the prompt-conditioned scene.

Existing approaches differ in how emotional supervision and content preservation are implemented. Feature-injection methods condition intermediate representations, while feedback-based methods evaluate decoded outputs. Text--image similarity promotes prompt relevance but does not explicitly preserve content. Our method combines continuous VA conditioning, output-level VA supervision, and a paired neutral reference.
Building on EmotiCrafter, we address semantic drift by supervising the decoded image and introducing a prompt-paired neutral reference. EmoGen is excluded from our VA-error comparison because its categorical control does not directly correspond to the $5\times5$ continuous VA grid; a fair comparison would require a predefined category-to-VA mapping and a common prompt and backbone protocol.

\subsection{Reinforcement Learning in Generative Models}

Many objectives in image generation, including human preference, aesthetics, and task-specific scores, can only be evaluated after an image has been decoded. Preference datasets and learned reward models provide output-level supervision for these objectives~\cite{kirstain2023pickapic,xu2023imagereward}. Policy optimization offers a means of improving such terminal objectives without differentiating through the evaluator. DPOK applies policy gradients with KL regularization to diffusion-model fine-tuning~\cite{fan2023dpok}, while DDPO formulates iterative denoising as a Markov decision process~\cite{black2024training}. Score-based SDE models provide a stochastic-process foundation for diffusion generation~\cite{song2021score}, and continuous-time formulations further connect generative modeling with stochastic control~\cite{zhang2025score}.

PPO introduces clipped policy updates for stable optimization~\cite{schulman2017ppo}. GRPO estimates relative advantages from a group of sampled outputs and avoids the need for a separately trained value model~\cite{shao2024deepseekmath}. DanceGRPO extends group-relative optimization to diffusion-based visual generation~\cite{xue2025dancegrpo}. Flow-GRPO adapts this approach to flow matching by replacing the deterministic training sampler with a marginal-preserving SDE and reducing the number of online denoising steps~\cite{flowgrpo2025}. Related group-relative objectives have also been investigated for masked generative models~\cite{luo2025maskgrpo}.

These methods provide general mechanisms for optimizing decoded-image objectives, but the resulting behavior remains dependent on the task-specific reward and constraints. In emotional synthesis, a VA reward alone may favor semantic shortcuts, whereas an overly strong semantic penalty may suppress perceptible emotional changes. We therefore adopt Flow-GRPO rather than redefining it. Its role is to make the final-image VA signal amenable to optimization in a flow-matching generator, while our task-specific contribution is to couple this policy objective with continuous VA conditioning and a semantic reference matched by prompt and initial latent.

\subsection{Reinforcement Learning in Affective Computing}
Reinforcement learning has also been explored in affective computing to optimize emotion-related objectives using feedback from generated or recognized visual content. In this setting, the supervision is typically obtained from the output rather than imposed solely through the generation process, making it relevant to our use of final-image VA supervision. EmoFeedback2 uses a large vision--language model to evaluate generated images and provide both an emotion-aware score and textual feedback~\cite{jia2025emofeedback2}. Its feedback loop uses the vision--language model to guide subsequent generation. EmoSENSE instead adopts hierarchical fuzzy reinforcement learning: a high-level module associates discrete emotion categories with visual attributes, while a low-level module adjusts the attribute mapping for different emotion--object combinations~\cite{emosense}. These studies demonstrate the value of output-level feedback, but their control variables and optimization signals differ from those of continuous VA-conditioned synthesis.

More specifically, the differences between these approaches and ours lie in three aspects. First, EmoSENSE employ categorical emotion control simlar as EmoGen, whereas our method uses a continuous VA coordinate. Second, EmoFeedback2 obtains learned score-and-text feedback from a vision--language model, whereas our scalar reward is defined as the negative numerical distance between the predicted and target VA coordinates. Third, our optimization incorporates a zero-VA image generated from the same prompt and initial latent as a paired semantic reference; neither categorical control nor affective feedback alone provides this constraint. This distinction motivates the proposed joint objective, and we do not claim that reward optimization or Flow-GRPO itself constitutes a novel contribution. Because the published systems employ different emotional interfaces, backbones, and evaluation protocols, the present experiments do not constitute a direct quantitative comparison with EmoGen, EmoFeedback2, or EmoSENSE.

\section{Method}

\subsection{Problem Formulation}

Let $c$ denote a neutral text prompt and let $\mathbf{y}=(v,a)\in[-3,3]^2$ denote the target valence--arousal coordinate. Given an initial latent $z_T\sim\mathcal{N}(0,I)$, the conditional generator $G_\theta$ produces an image $x_{\mathbf{y}}=G_\theta(z_T,c,\mathbf{y})$. A frozen VA predictor $P_{\mathrm{VA}}$ maps the image to $\hat{\mathbf{y}}=(\hat v,\hat a)$, while a frozen visual encoder $f$ provides a representation used to measure semantic deviation. The objective is to reduce the distance between $\hat{\mathbf{y}}$ and $\mathbf{y}$ without changing the scene layout, object identity, or visual intent specified by $c$.

These two requirements are tightly coupled and, in practice, often conflict with each other. When the emotional signal becomes stronger, the generator may indeed move closer to the desired emotion in the valence-arousal space, but such improvement is frequently achieved at the cost of semantic fidelity. For example, instead of expressing emotion through subtle changes in color tone, lighting, facial expression, or atmosphere, the model may alter the scene composition, replace important objects, or modify the environment in a way that changes the meaning of the original prompt. We refer to this phenomenon as \emph{emotion-semantic drift}. Therefore, emotional image generation should not be viewed as a pure emotion prediction problem or a pure text-to-image generation problem, but rather as a joint optimization problem in which emotional expressiveness must be improved under an explicit semantic preservation constraint.

Based on this observation, we treat the denoising generator as a policy and optimize two complementary terms: a terminal reward for continuous emotion alignment and an anchor regularizer for semantic preservation. The reward is defined on the final decoded image and can therefore be optimized without differentiating through the frozen VA predictor. The anchor term remains differentiable through the generated image and restricts the policy from obtaining a high emotional reward through large content changes. Sections III-C--III-F define the conditioning mechanism, anchor regularizer, stochastic policy, and joint objective, respectively.

\subsection{Framework Overview}
\begin{figure*}[t]
\centering
\includegraphics[width=\textwidth]{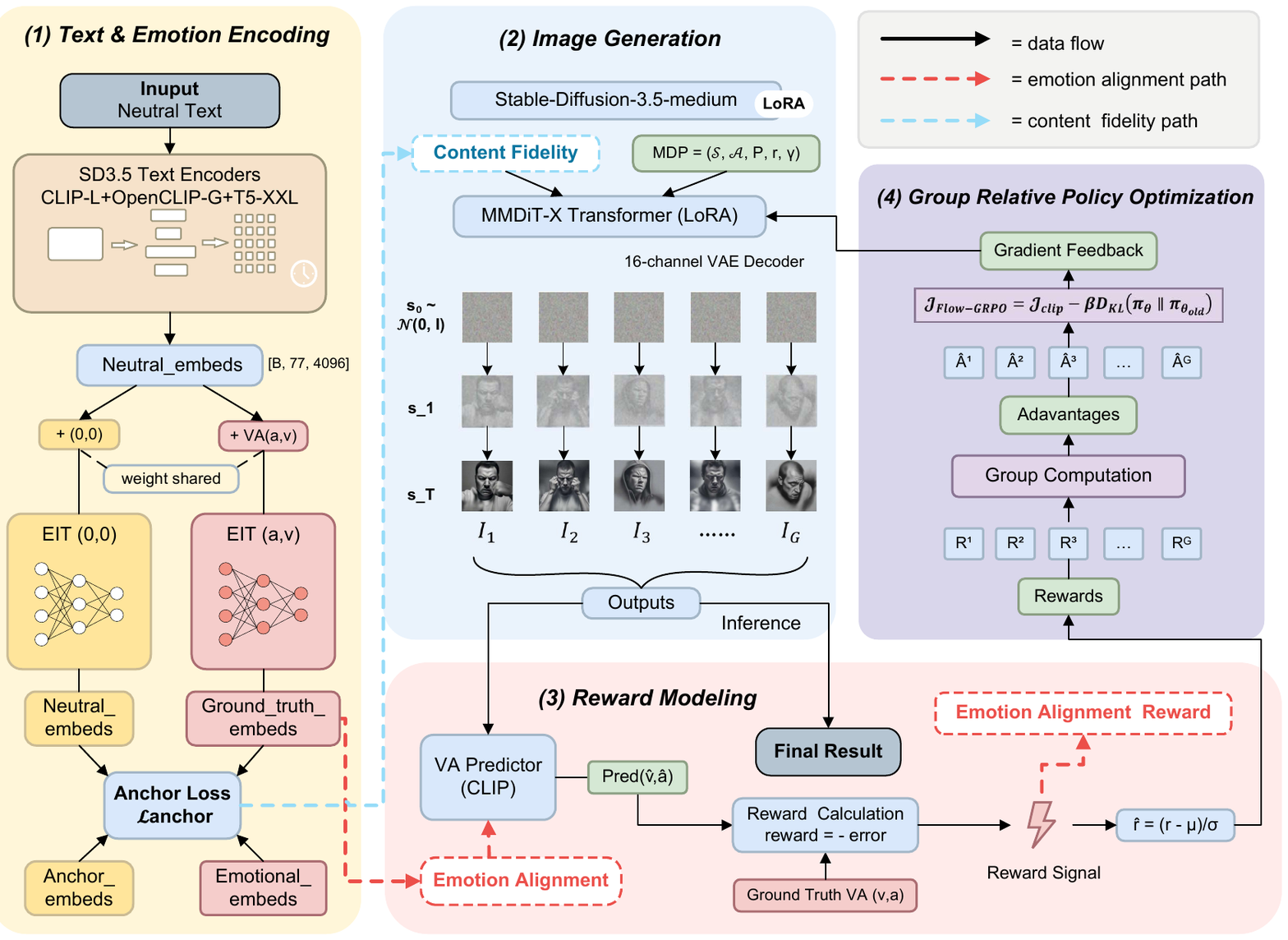}
\caption{Overview of the proposed framework for emotion-aware image generation with dual mechanisms: Flow-GRPO optimization and anchor-based semantic constraint.}
\label{fig:wide}
\end{figure*}

Fig.~\ref{fig:wide} illustrates the overall architecture. The framework jointly operates two complementary mechanisms during training: Flow-GRPO for emotional alignment and an anchor-based constraint for semantic preservation. The former addresses the objective mismatch of supervised emotion injection, while the latter prevents the model from achieving high emotional reward by sacrificing semantic fidelity.
Starting from a neutral text prompt, the model first produces a text representation through the backbone text encoder. The target valence--arousal signal is then injected into this representation by an Emotion Injection Transformer (EIT), producing an emotion-conditioned embedding that guides the flow-matching image generator. The final image is evaluated along two paths. The emotion path applies a frozen VA predictor and returns a terminal reward to GRPO. The semantic path forms a neutral branch by setting the VA condition to $(0,0)$ and compares the decoded emotional and neutral images in the CLIP visual feature space. Thus, the embedding labels in Fig.~\ref{fig:wide} identify the conditioning signals used to generate the two branches; the anchor distance itself is evaluated between visual features of the corresponding decoded images, as defined in Eq.~\eqref{eq:anchor_loss}.

If trained only on emotional reward, the model could exploit shortcuts by altering scene structure to improve the emotion score. Conversely, training only on the anchor would suppress emotional variation. Our framework balances these forces: Flow-GRPO encourages emotionally accurate generation, while the anchor constraint regularizes the optimization trajectory so that the learned policy remains faithful to the original content. In this sense, the entire method can be understood as a constrained emotional generation framework in which emotion alignment is directly optimized under semantic preservation.

\subsection{VA-Conditioned Generation}

A core requirement of our task is continuous emotional control rather than selection from a small set of discrete emotion categories. We therefore inject the target VA coordinate into the text condition while retaining the generative prior of the pre-trained image backbone. Let $E_{\mathrm{text}}$ denote the frozen text-encoding stack of SD3.5M(Stable Diffusion 3.5 Medium). The neutral prompt representation is
\begin{equation}
\mathbf{e}_{0}=E_{\mathrm{text}}(c)\in\mathbb{R}^{L\times d},
\label{eq:neutral_embedding}
\end{equation}
where $L$ is the sequence length after the backbone-specific tokenization and feature assembly, and $d$ is the resulting channel dimension. This notation deliberately abstracts over the multiple text encoders used by SD3.5M and avoids assigning a single encoder-specific token length to the combined condition.

The Emotion Injection Transformer, parameterized by $\phi$, fuses the continuous condition $\mathbf{y}=(v,a)$ with the neutral representation:
\begin{equation}
\begin{aligned}
\mathbf{e}_{\mathbf{y}}
&=\operatorname{EIT}_{\phi}(\mathbf{e}_{0},\mathbf{y}), \\
\mathbf{e}_{\mathrm{anchor}}
&=\operatorname{EIT}_{\phi}(\mathbf{e}_{0},\mathbf{0}).
\end{aligned}
\label{eq:eit_condition}
\end{equation}
The two branches share the EIT parameters and differ only in their VA inputs. The emotional representation $\mathbf{e}_{\mathbf{y}}$ conditions the flow-matching generator, whereas $\mathbf{e}_{\mathrm{anchor}}$ supplies the zero-VA condition used to construct the neutral reference. With $D_{\mathrm{VAE}}$ denoting the frozen decoder, the emotional output is
\begin{equation}
x_{\mathrm{emo}}=D_{\mathrm{VAE}}\!\left(z_0\right),
\qquad
z_0\sim p_{\theta}(z_0\mid z_T,\mathbf{e}_{\mathbf{y}}),
\label{eq:conditional_generation}
\end{equation}
where $z_T\sim\mathcal{N}(0,I)$ and $\theta$ denotes the trainable LoRA parameters of the image generator.

LoRA is inserted into the attention layers of the image generator. The original backbone and VAE weights remain frozen, while the LoRA parameters $\theta$ and EIT parameters $\phi$ are optimized. This parameter-efficient design limits changes to the pre-trained generative prior and reduces the memory required for online trajectory sampling.

\subsection{Anchor-Based Semantic Constraint}

Although continuous emotion injection allows the model to condition generation on target valence-arousal values, this alone is not sufficient to guarantee semantic fidelity. A common failure mode in emotion-conditioned generation is that the model reduces emotional prediction error by making large semantic changes to the image, such as altering the scene type, replacing key objects, or dramatically changing the layout. These changes may improve the score of an emotion predictor, but they do so by violating the intended content of the prompt. To explicitly discourage such behavior, we introduce an anchor-based semantic constraint.

The key idea is to construct, for each prompt, a neutral reference image that represents the prompt content without an explicit emotional displacement. The neutral branch uses the same prompt and the shared EIT under $\mathbf{y}=\mathbf{0}$:
\begin{equation}
x_{\mathrm{anchor}}=G_{\theta}(z_T,c,\mathbf{0}).
\label{eq:anchor_generation}
\end{equation}
During computation of the semantic target, the anchor feature is detached from the gradient graph so that the neutral branch acts as a reference rather than being pulled toward the emotional output. We avoid pixel-level matching because color, illumination, and local appearance are legitimate carriers of emotion. Instead, the anchor loss is evaluated in the visual feature space:
\begin{equation}
\mathcal{L}_{\text{anchor}}
=\left\|f(x_{\text{emo}})
-\operatorname{sg}\!\left[f(x_{\text{anchor}})\right]\right\|_2^2,
\label{eq:anchor_loss}
\end{equation}
where $f(\cdot)$ is the frozen CLIP image encoder and $\operatorname{sg}[\cdot]$ denotes stop-gradient. The loss does not require the two images to be pixel-identical; it penalizes changes in their high-level visual representations and therefore favors preservation of the scene identity while permitting affective changes in tone and appearance.

This design is particularly important in the reinforcement learning setting because the policy may otherwise exploit semantically undesirable shortcuts to reduce the VA error. The balancing coefficient $\lambda$ determines the trade-off: larger values emphasize proximity to the neutral reference, whereas smaller values permit greater movement toward the target emotion.

\subsection{GRPO-based Optimization}

While the anchor term constrains semantic drift, emotion alignment is optimized from a terminal signal on the final decoded image. We model stochastic denoising as a finite-horizon MDP. At timestep $t$, the state is
\begin{equation}
s_t=(c,\mathbf{y},t,z_t),
\end{equation}
and the action is the next denoised latent $a_t=z_{t-\Delta t}$. The corresponding policy is the conditional transition density
\begin{equation}
\pi_\theta(a_t\mid s_t)
=p_\theta(z_{t-\Delta t}\mid z_t,c,\mathbf{y}),
\label{eq:transition_policy}
\end{equation}
where $\Delta t>0$ denotes the magnitude of one reverse-time step. A trajectory is $\tau=(z_T,\ldots,z_0)$, and a nonzero reward is assigned only after decoding $z_0$.

\subsubsection{VA Terminal Reward}
The frozen predictor $P_{\mathrm{VA}}$ consists of a CLIP image encoder~\cite{radford2021learning} followed by the released valence and arousal regression heads used by EmotiCrafter~\cite{dang2025emoticrafter_iccv}. It is not updated during policy optimization. Given
$(\hat v,\hat a)=P_{\mathrm{VA}}(x_{\mathrm{emo}})$, the terminal reward is
\begin{equation}
R^i=-\sqrt{(\hat v^i-v)^2+(\hat a^i-a)^2}.
\label{eq:va_reward}
\end{equation}
Thus, larger rewards correspond to smaller Euclidean errors in the continuous VA space.

\subsubsection{Marginal-preserving Stochastic Policy}
The original rectified-flow sampler follows the deterministic ODE
\begin{equation}
dz_t=v_\theta(z_t,t,\mathbf{e}_{\mathbf{y}})\,dt,
\label{eq:flow_ode}
\end{equation}
where $v_\theta$ is the conditional velocity field. Although different values of $z_T$ generate different images, a fixed $z_t$ has a deterministic successor under Eq.~\eqref{eq:flow_ode}. Its conditional transition is therefore a Dirac measure, so the likelihood ratio required by policy-gradient optimization is not defined as an ordinary probability-density ratio.

Following Flow-GRPO~\cite{flowgrpo2025}, we use the marginal-preserving SDE for rectified flow:
\begin{equation}
\begin{aligned}
dz_t={}&\left[v_\theta(z_t,t,\mathbf{e}_{\mathbf{y}})
+\frac{\sigma_t^2}{2t}\left(z_t+(1-t)
v_\theta(z_t,t,\mathbf{e}_{\mathbf{y}})\right)\right]dt \\
&+\sigma_t\,d\bar w_t,
\end{aligned}
\label{eq:flow_sde}
\end{equation}
where $\bar w_t$ is a standard Wiener process and $\sigma_t$ controls the injected stochasticity. The additional score-dependent drift in Eq.~\eqref{eq:flow_sde}, rather than noise injection alone, is what preserves the marginal density of the original probability path. We use the Flow-GRPO schedule $\sigma_t=\alpha\sqrt{t/(1-t)}$, with noise coefficient $\alpha=0.3$. Euler--Maruyama discretization makes Eq.~\eqref{eq:transition_policy} an isotropic Gaussian whose mean depends on $v_\theta$ and whose covariance is $\sigma_t^2\Delta t I$. Consequently, both the transition likelihood and its ratio to an old or reference policy are tractable.

\subsubsection{Group-relative Policy Objective}
For each condition $(c,\mathbf{y})$, the old policy samples a group of $G$ trajectories. Their terminal rewards are normalized within the group:
\begin{equation}
\widehat A^i=\frac{R^i-\mu_R}{\sigma_R+\delta},
\qquad
\mu_R=\frac{1}{G}\sum_{j=1}^{G}R^j,
\label{eq:group_advantage}
\end{equation}
where $\sigma_R$ is the group reward standard deviation and $\delta$ is a small numerical constant. The step-wise importance ratio is
\begin{equation}
r_t^i(\theta)=
\frac{\pi_\theta(z_{t-\Delta t}^i\mid z_t^i,c,\mathbf{y})}
{\pi_{\theta_{\mathrm{old}}}(z_{t-\Delta t}^i\mid z_t^i,c,\mathbf{y})}.
\label{eq:policy_ratio}
\end{equation}
We denote the quantity maximized by GRPO as
\begin{equation}
\begin{aligned}
\mathcal{J}_{\mathrm{GRPO}}(\theta)
=\frac{1}{GT}\sum_{i=1}^{G}\sum_{t=1}^{T}\Big[&
\min\big(r_t^i\widehat A^i,
\operatorname{clip}(r_t^i,1-\epsilon,\\1+\epsilon)\widehat A^i\big)
&-\beta D_{\mathrm{KL}}(\pi_\theta\|\pi_{\mathrm{ref}})\Big],
\end{aligned}
\label{eq:grpo_objective}
\end{equation}
where $\epsilon$ is the clipping coefficient, $\beta$ controls deviation from the frozen reference policy, and $T$ is the number of sampled denoising transitions. Writing this term as $\mathcal{J}_{\mathrm{GRPO}}$, rather than as a loss, makes its maximization direction explicit.
To reduce the cost of collecting $G$ trajectories for every condition, training uses the denoising-reduction schedule of Flow-GRPO. The reduced schedule is used only for online sampling and policy updates; the original denoising schedule is restored at inference.

\subsection{Overall Objective}

The final minimization objective combines the negative policy objective with the anchor regularizer:
\begin{equation}
\mathcal{L}_{\mathrm{total}}
=-\mathcal{J}_{\mathrm{GRPO}}+\lambda\mathcal{L}_{\mathrm{anchor}}.
\label{eq:overall_objective}
\end{equation}
The sign of the first term reflects that $\mathcal{J}_{\mathrm{GRPO}}$ is maximized, whereas $\mathcal{L}_{\mathrm{total}}$ is minimized. The coefficient $\lambda$ controls the empirical trade-off between VA alignment and proximity to the zero-VA reference. Equation~\eqref{eq:overall_objective} therefore optimizes emotional alignment and semantic preservation jointly rather than applying semantic correction as a post-processing step.
\begin{figure}[t]
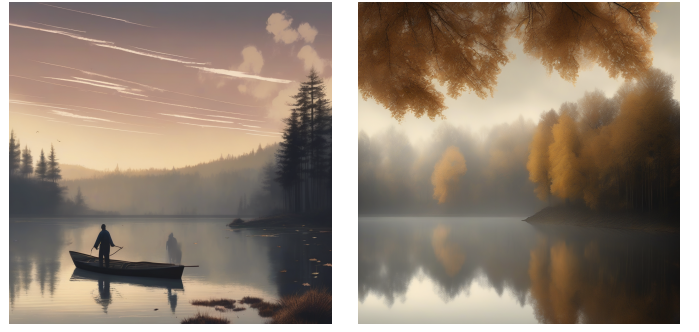

    \centering
    \begin{subfigure}[b]{0.48\linewidth}
        \centering
        \includegraphics[width=\linewidth]{fig_anchor_without.pdf}
        \subcaption{Without anchor constraint}
        \label{fig:anchor_without}
    \end{subfigure}
    \hfill
    \begin{subfigure}[b]{0.48\linewidth}
        \centering
        \includegraphics[width=\linewidth]{fig_anchor_with.pdf}
        \subcaption{With anchor constraint (Ours)}
        \label{fig:anchor_with}
    \end{subfigure}
    \caption{Effect of the anchor constraint on a landscape prompt. Given the prompt ``A tranquil lake surrounded by golden autumn trees, misty morning, no people'' with target emotion (valence=3, arousal=3), the model without an anchor (left) introduces unintended elements such as a person and a boat. 
    }
    \label{fig:anchor_effect}
\end{figure}
Fig.~\ref{fig:anchor_effect} provides a qualitative illustration of the two terms in Eq.~\eqref{eq:overall_objective}: without the anchor, the sampled output introduces prompt-inconsistent elements, whereas the anchored output remains closer to the requested scene while moving toward the target VA condition. The aggregate effect of this regularizer is evaluated by the GRPO-only and full-model comparison in Section~\ref{sec:experiments}.

\section{Experiments}\label{sec:experiments}
This section reports the experimental setup, qualitative comparisons, automatic metrics, component ablations, user study and further discussion. 

\begin{figure*}[t]
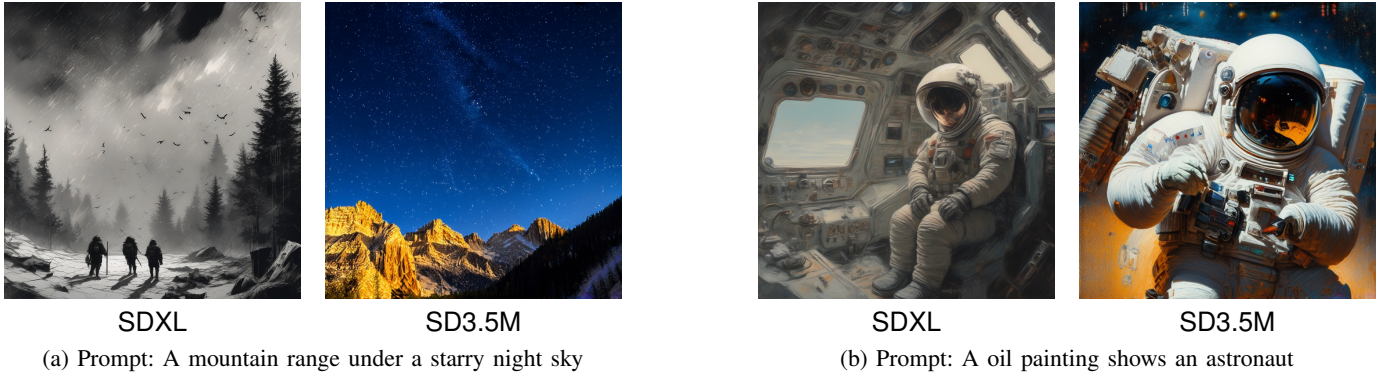

    \centering
    \begin{subfigure}[b]{0.45\textwidth}
        \centering
        \begin{minipage}[b]{0.48\linewidth}
            \centering
            \includegraphics[width=\linewidth]{fig_sdxl_prompt1.pdf}
            \textsf{SDXL}
        \end{minipage}
        \hfill
        \begin{minipage}[b]{0.48\linewidth}
            \centering
            \includegraphics[width=\linewidth]{fig_sd35m_prompt1.pdf}
            \textsf{SD3.5M}
        \end{minipage}
        \\
        \caption*{(a) Prompt: A mountain range under a starry night sky}
        \label{fig:group1}
    \end{subfigure}
    \hfill
    \begin{subfigure}[b]{0.45\textwidth}
        \centering
        \begin{minipage}[b]{0.48\linewidth}
            \centering
            \includegraphics[width=\linewidth]{fig_sdxl_prompt2.pdf}
            \textsf{SDXL}
        \end{minipage}
        \hfill
        \begin{minipage}[b]{0.48\linewidth}
            \centering
            \includegraphics[width=\linewidth]{fig_sd35m_prompt2.pdf}
            \textsf{SD3.5M}
        \end{minipage}
        \\
        \caption*{(b) Prompt: A oil painting shows an astronaut}
        \label{fig:group2}
    \end{subfigure}
    \caption{Comparison of SDXL and SD3.5M under two different prompts. For each prompt, SD3.5M (right) generates images semantically closer to the prompt than SDXL (left), as reflected by higher CLIPScore. However, neither model alone achieves high emotional accuracy, motivating our GRPO+Anchor fine-tuning.}
    \label{fig:backbone}
\end{figure*}

\begin{figure}[htbp]
    \centering
    \includegraphics[width=\columnwidth]{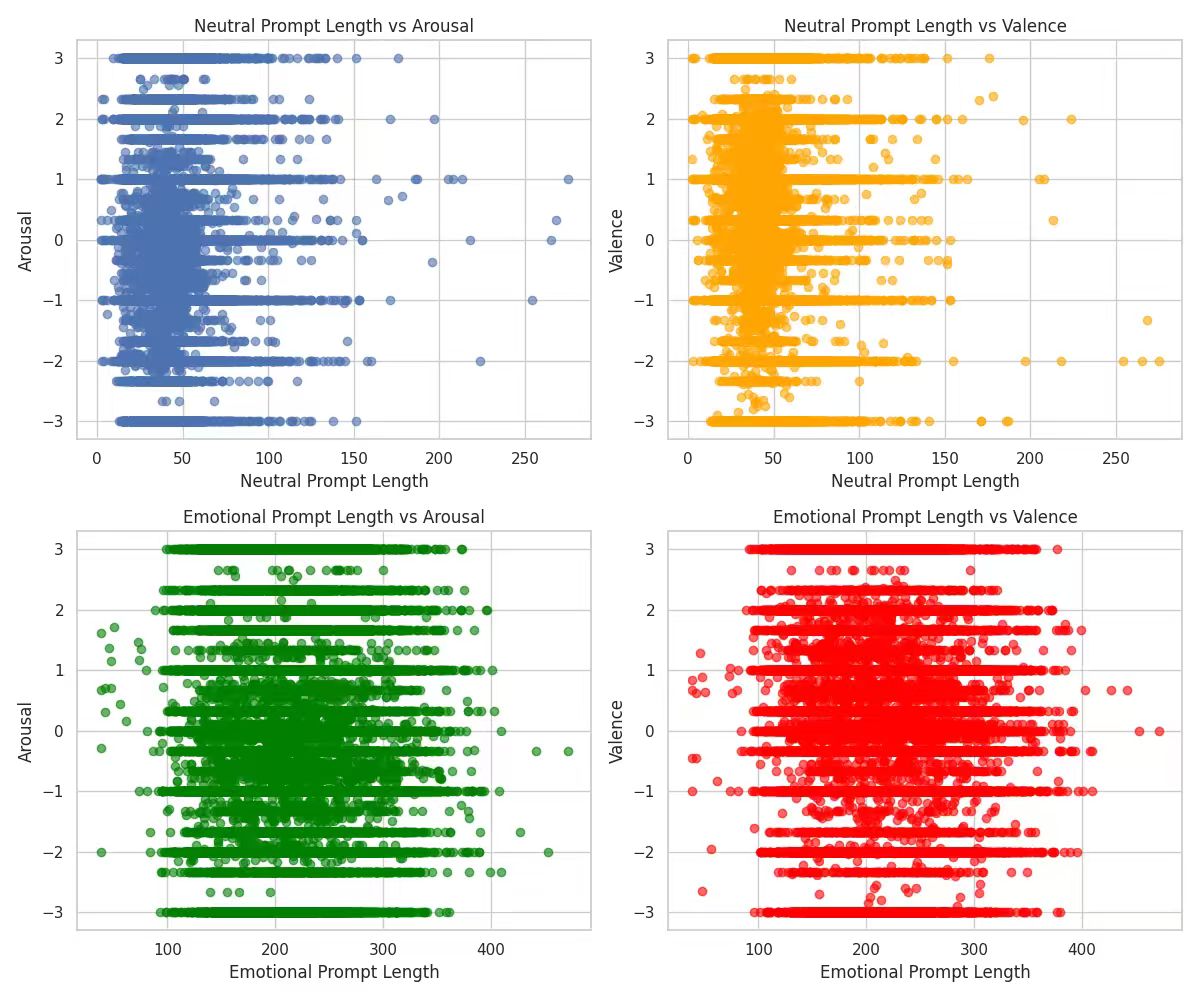}
    \caption{Overview of the dataset statistics. The figures show neutral prompt length, arousal and valence values, and emotional prompt length for different data splits (Blue, Green, Red).}
    \label{fig:dataset_stats}
\end{figure}

\subsection{Implementation Details}
Following standard practice in emotional image generation~\cite{dang2025emoticrafter_iccv}, we employ a dataset comprising paired neutral and emotional prompts annotated with continuous valence--arousal (VA) values. Fig.~\ref{fig:dataset_stats} summarizes the prompt lengths, VA annotations, and data splits. Evaluation covers emotional accuracy, prompt consistency, and reference-free image quality. For the VA reward and V/A-error computation, we use the frozen predictors released with EmotiCrafter: a CLIP image encoder followed by separate regression checkpoints for valence and arousal. Images are processed with the predictor's released CLIP preprocessing, and its outputs use the same $[-3,3]$ VA range as the target annotations. No predictor parameter is updated during GRPO training. Valence Error (V-Err) and Arousal Error (A-Err) are the mean absolute errors between these predictions and the target coordinates. Content preservation is evaluated using CLIPScore~\cite{hessel2021clipscore}, and perceptual image quality is assessed using CLIP-IQA~\cite{wang2023exploring}.

All metrics are reported on a held-out test set of 3,300 images (132 prompts × 5 valence values × 5 arousal values), consistent with prior work~\cite{dang2025emoticrafter_iccv}.
Our experiments are designed to isolate the impact of both the backbone model (SDXL vs. SD3.5M) and the optimization strategy, including supervised baseline, anchor-only, GRPO-only, and their combination. 
We adopt SDXL~\cite{podell2023sdxl} and SD3.5M~\cite{esser2024sdxl}, both fine-tuned with LoRA (rank \(r=4\)).

As shown in Fig.~\ref{fig:backbone}, we compare SDXL~\cite{podell2023sdxl} and SD3.5M~\cite{esser2024sdxl} under the same EIT injection settings. While SDXL has been widely adopted in prior conditional generation works, we observe that SD3.5M exhibits a stronger semantic prior that better preserves prompt-relevant content under emotional conditioning. Quantitatively, although SD3.5M shows a slightly higher initial VA error after EIT injection due to its more "rigid" semantic representation, it achieves substantially lower errors and higher CLIPScore after our GRPO+Anchor optimization. This can be attributed to SD3.5M's advanced MMDiT architecture and its richer text conditioning space (incorporating T5-XXL), which together provide a more reliable foundation for the anchor-based semantic constraint. Consequently, SD3.5M enables the model to better disentangle emotional expression from prompt semantics, motivating its use as the backbone for all subsequent experiments in this paper.

For reinforcement learning-based configurations, we adopt Flow-GRPO with group size $G=24$, clipping range $\epsilon=1\times10^{-4}$, and group-wise advantage normalization. In the SDE schedule of Eq.~\eqref{eq:flow_sde}, the scalar noise coefficient is $\alpha=0.3$.
When the anchor constraint is applied, the anchor loss is defined as the squared Euclidean distance between the CLIP embeddings of the generated image and a neutral reference image with VA = 0, with balancing coefficient \(\lambda=0.5\).
The ``Scale'' settings reported in the ablation table refer to the residual-amplification factor inherited from the EmotiCrafter EIT objective, rather than classifier-free guidance. It controls the magnitude of the target displacement from the neutral prompt feature toward the emotional prompt feature. All models are trained on two NVIDIA H100 GPUs for approximately 12 hours. The reward function is the negative Euclidean distance in Eq.~\eqref{eq:va_reward}.

\subsection{Qualitative Results}

\begin{figure*}[t]
    \centering
    \begin{minipage}[c]{0.05\textwidth}
        \rotatebox{90}{\textbf{Baseline}}
    \end{minipage}
    \begin{minipage}[c]{0.9\textwidth}
        \begin{subfigure}[b]{0.19\textwidth}
            \centering
            \includegraphics[width=\textwidth]{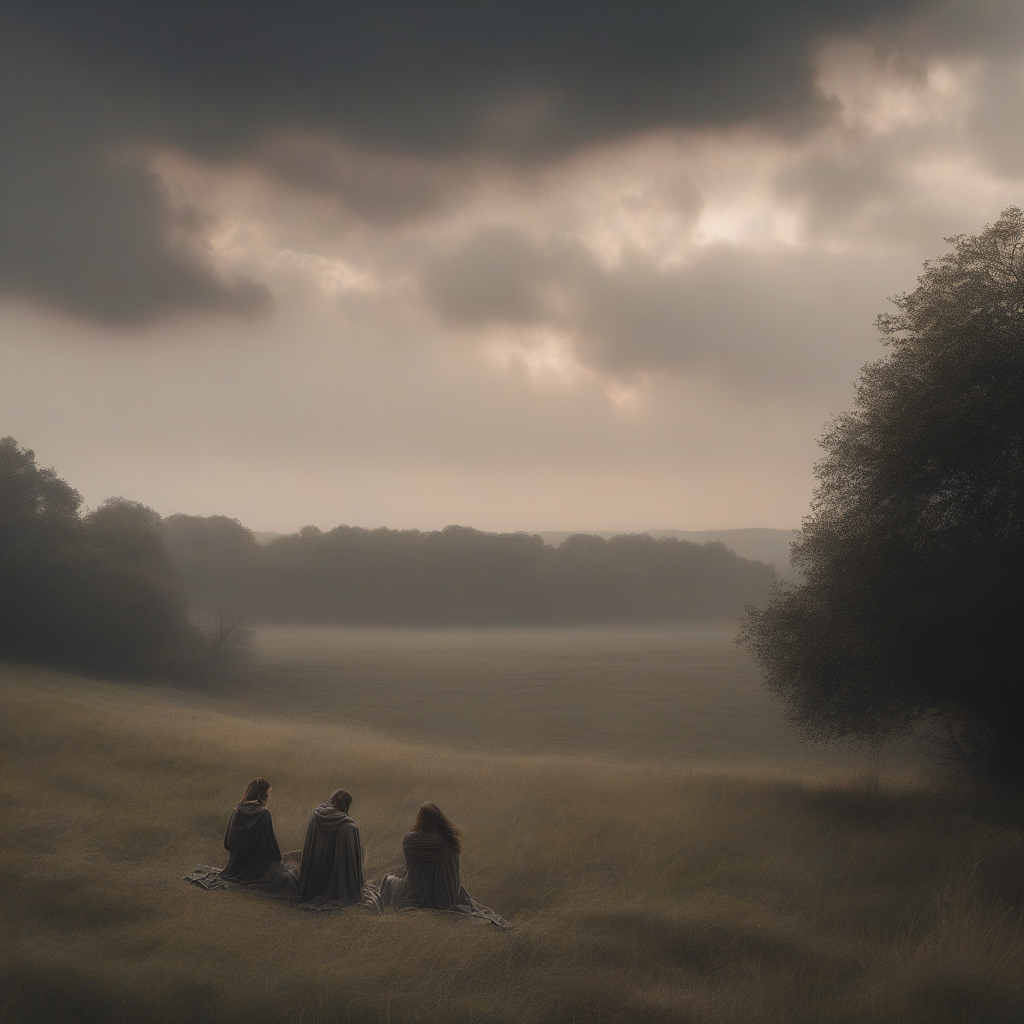}
            \caption{meadow}
            \label{fig:baseline_meadow}
        \end{subfigure}
        \hfill
        \begin{subfigure}[b]{0.19\textwidth}
            \centering
            \includegraphics[width=\textwidth]{fig_sdxl_prompt1.pdf}
            \caption{mountain}
            \label{fig:baseline_mountain}
        \end{subfigure}
        \hfill
        \begin{subfigure}[b]{0.19\textwidth}
            \centering
            \includegraphics[width=\textwidth]{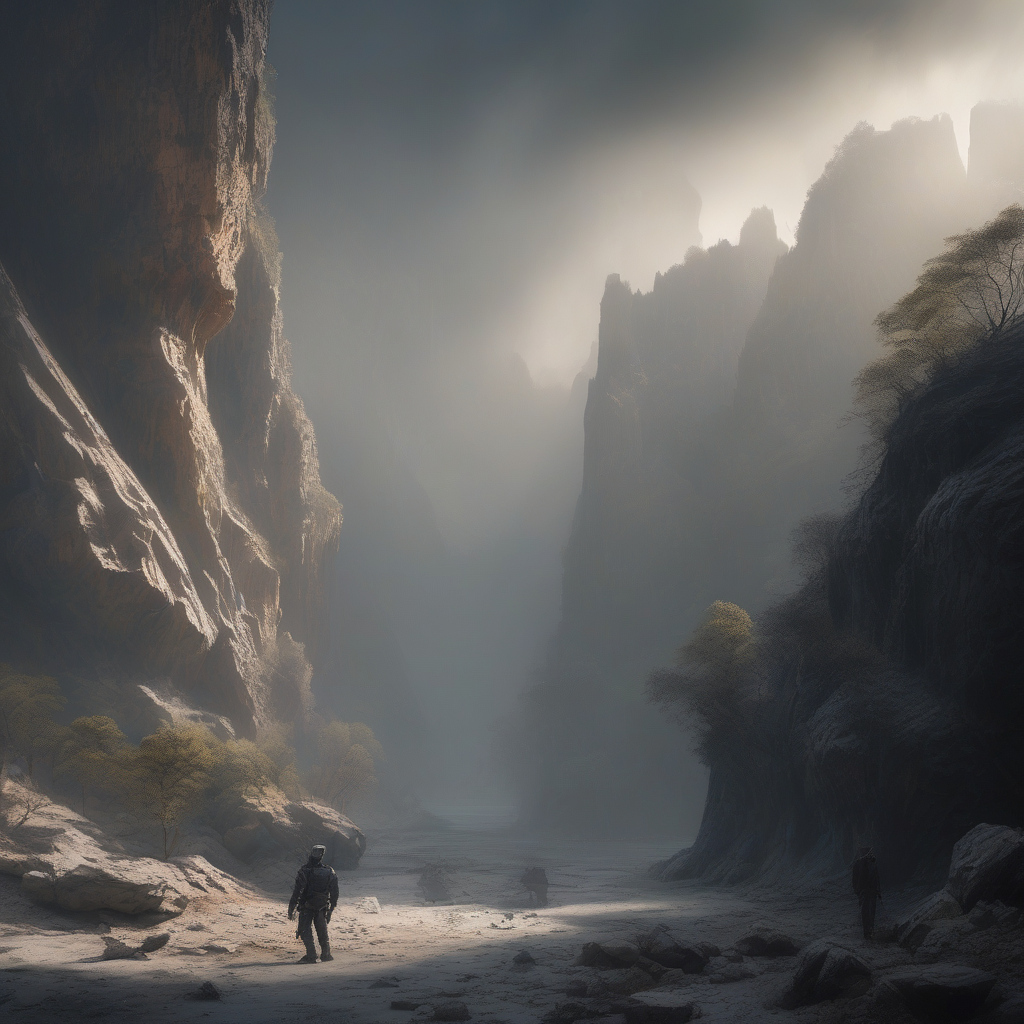}
            \caption{river}
            \label{fig:baseline_river}
        \end{subfigure}
        \hfill
        \begin{subfigure}[b]{0.19\textwidth}
            \centering
            \includegraphics[width=\textwidth]{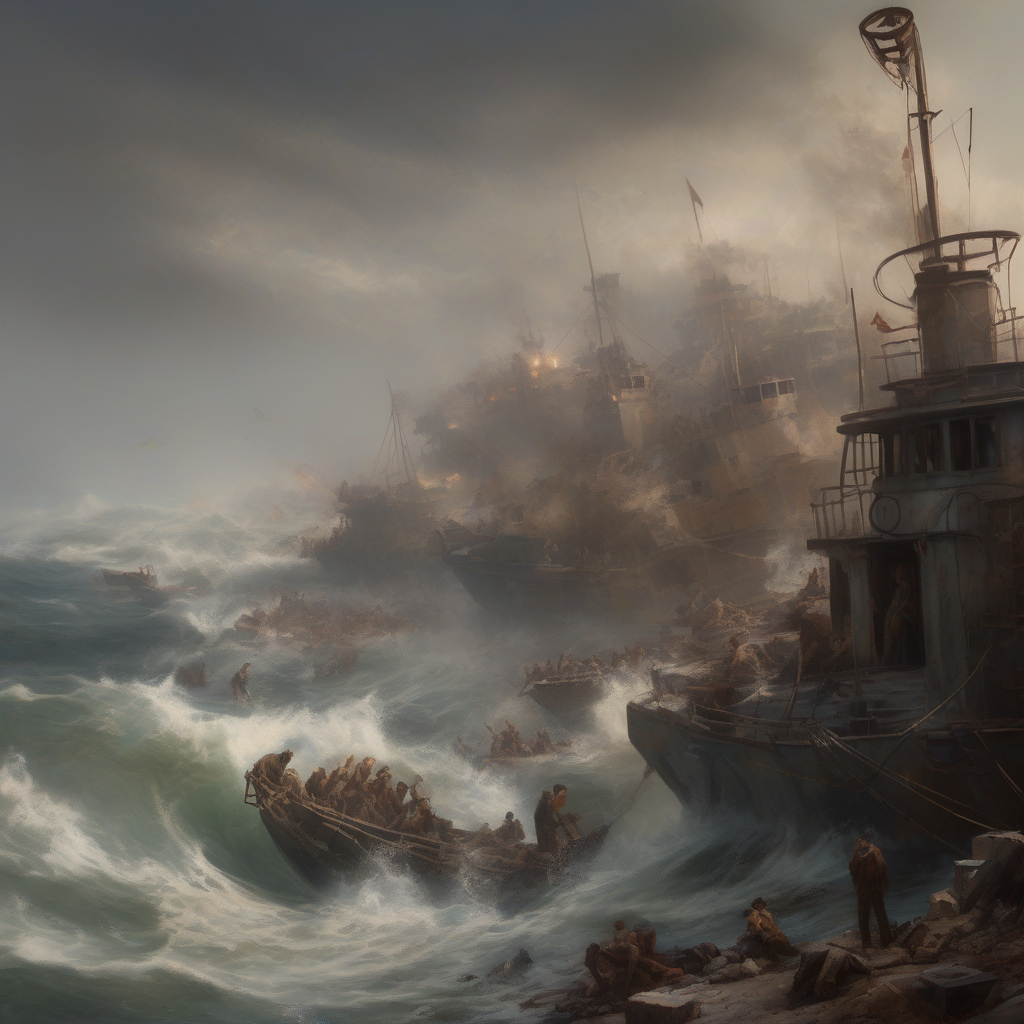}
            \caption{beach}
            \label{fig:baseline_beach}
        \end{subfigure}
        \hfill
        \begin{subfigure}[b]{0.19\textwidth}
            \centering
            \includegraphics[width=\textwidth]{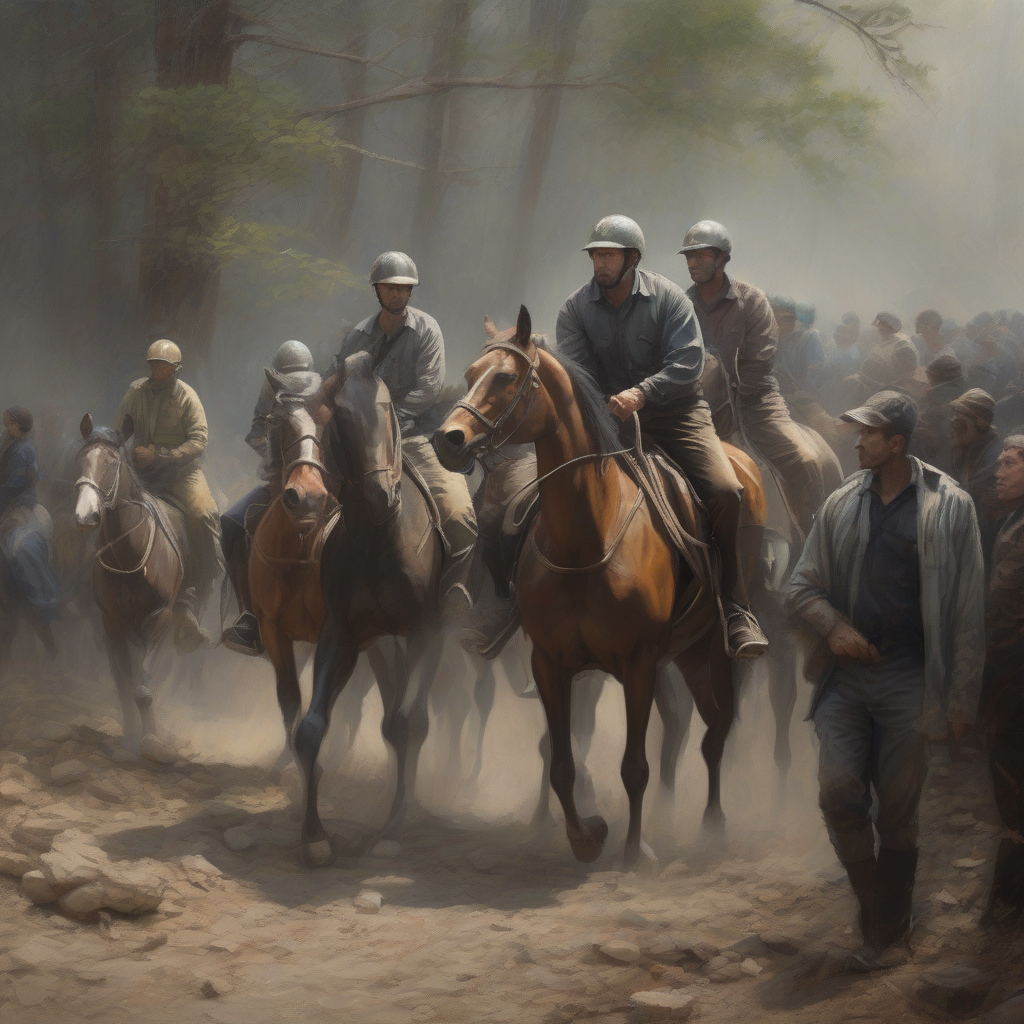}
            \caption{horses}
            \label{fig:baseline_horses}
        \end{subfigure}
    \end{minipage}
    
    \vspace{2ex}
    
    \begin{minipage}[c]{0.05\textwidth}
        \rotatebox{90}{\textbf{Ours}}
    \end{minipage}
    \begin{minipage}[c]{0.9\textwidth}
        \begin{subfigure}[b]{0.19\textwidth}
            \centering
            \includegraphics[width=\textwidth]{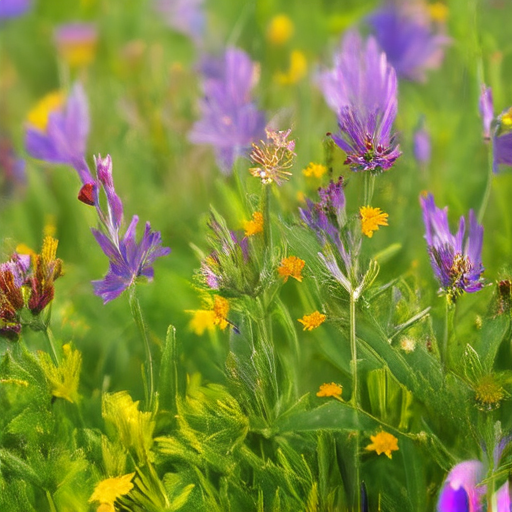}
            \caption{meadow}
            \label{fig:ours_meadow}
        \end{subfigure}
        \hfill
        \begin{subfigure}[b]{0.19\textwidth}
            \centering
            \includegraphics[width=\textwidth]{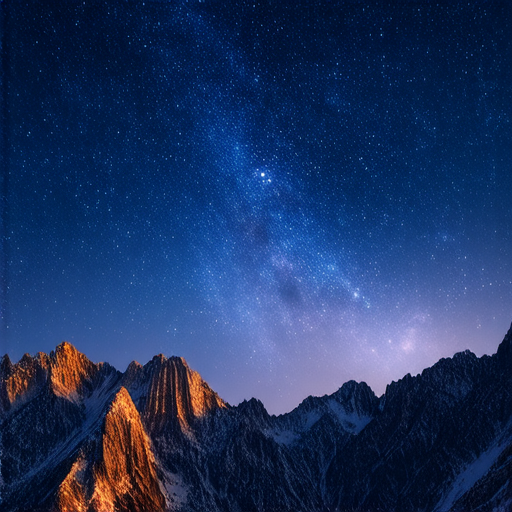}
            \caption{mountain}
            \label{fig:ours_mountain}
        \end{subfigure}
        \hfill
        \begin{subfigure}[b]{0.19\textwidth}
            \centering
            \includegraphics[width=\textwidth]{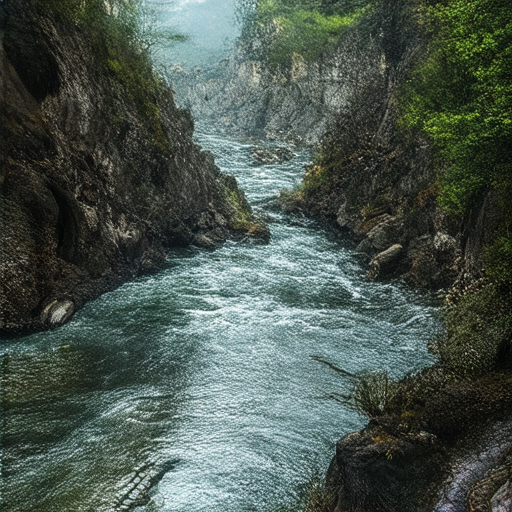}
            \caption{river}
            \label{fig:ours_river}
        \end{subfigure}
        \hfill
        \begin{subfigure}[b]{0.19\textwidth}
            \centering
            \includegraphics[width=\textwidth]{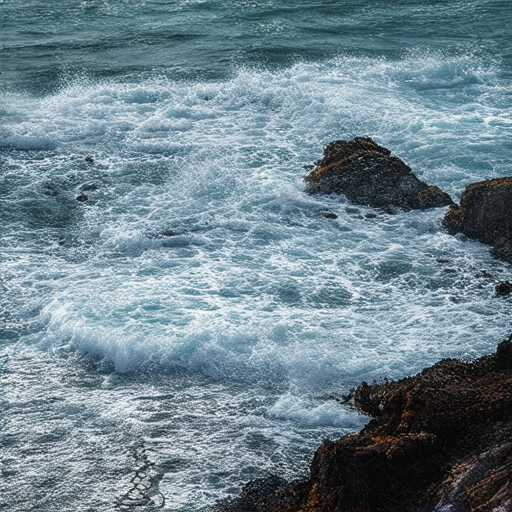}
            \caption{beach}
            \label{fig:ours_beach}
        \end{subfigure}
        \hfill
        \begin{subfigure}[b]{0.19\textwidth}
            \centering
            \includegraphics[width=\textwidth]{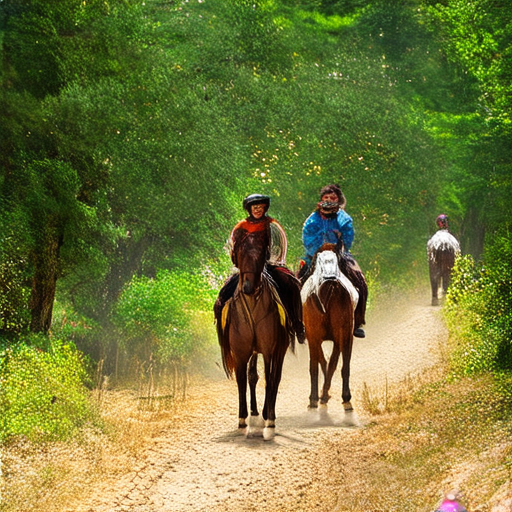}
            \caption{horses}
            \label{fig:ours_horses}
        \end{subfigure}
    \end{minipage}
    
    \caption{Qualitative comparison across five prompts. For each prompt, we compare the baseline model (top row) and our method (bottom row). The prompts are (from left to right): (1) ``A meadow filled with wildflowers'', (2) ``A mountain range under a starry night sky'', (3) ``A river running through a canyon'', (4) ``A rocky beach with crashing waves'', (5) ``People riding horses on a trail''. The baseline often alters the scene structure, while our method preserves the original content more consistently.}
    \label{fig:qualitative_multi_prompt}
\end{figure*}
\begin{figure*}[t]
    \centering
    \begin{subfigure}[t]{0.32\textwidth}
        \centering
        \includegraphics[width=\linewidth]{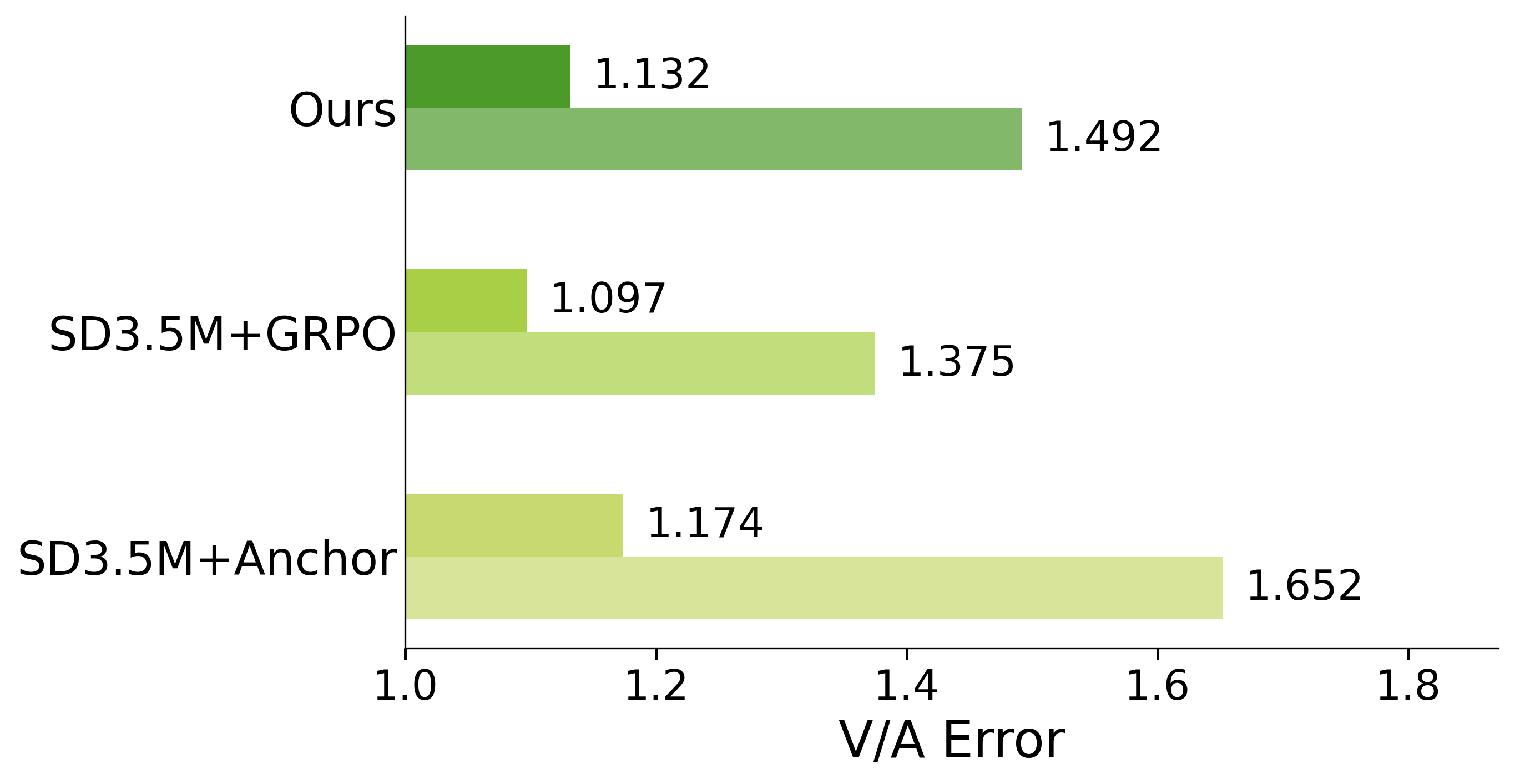}
        \vspace{2mm}
        \includegraphics[width=\linewidth]{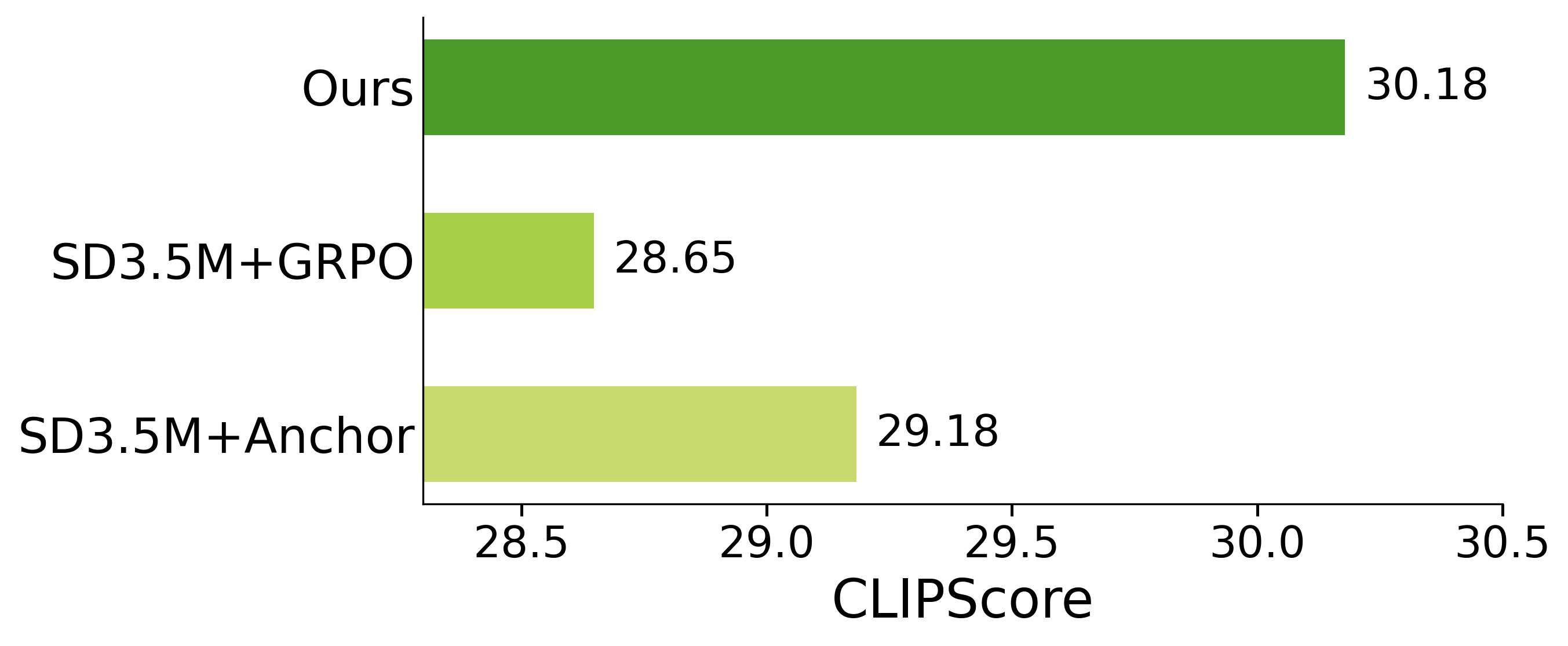}
        \caption{Variants on SD3.5M}
        \label{fig:ablation_group1}
    \end{subfigure}
    \hfill
    \begin{subfigure}[t]{0.32\textwidth}
        \centering
        \includegraphics[width=\linewidth]{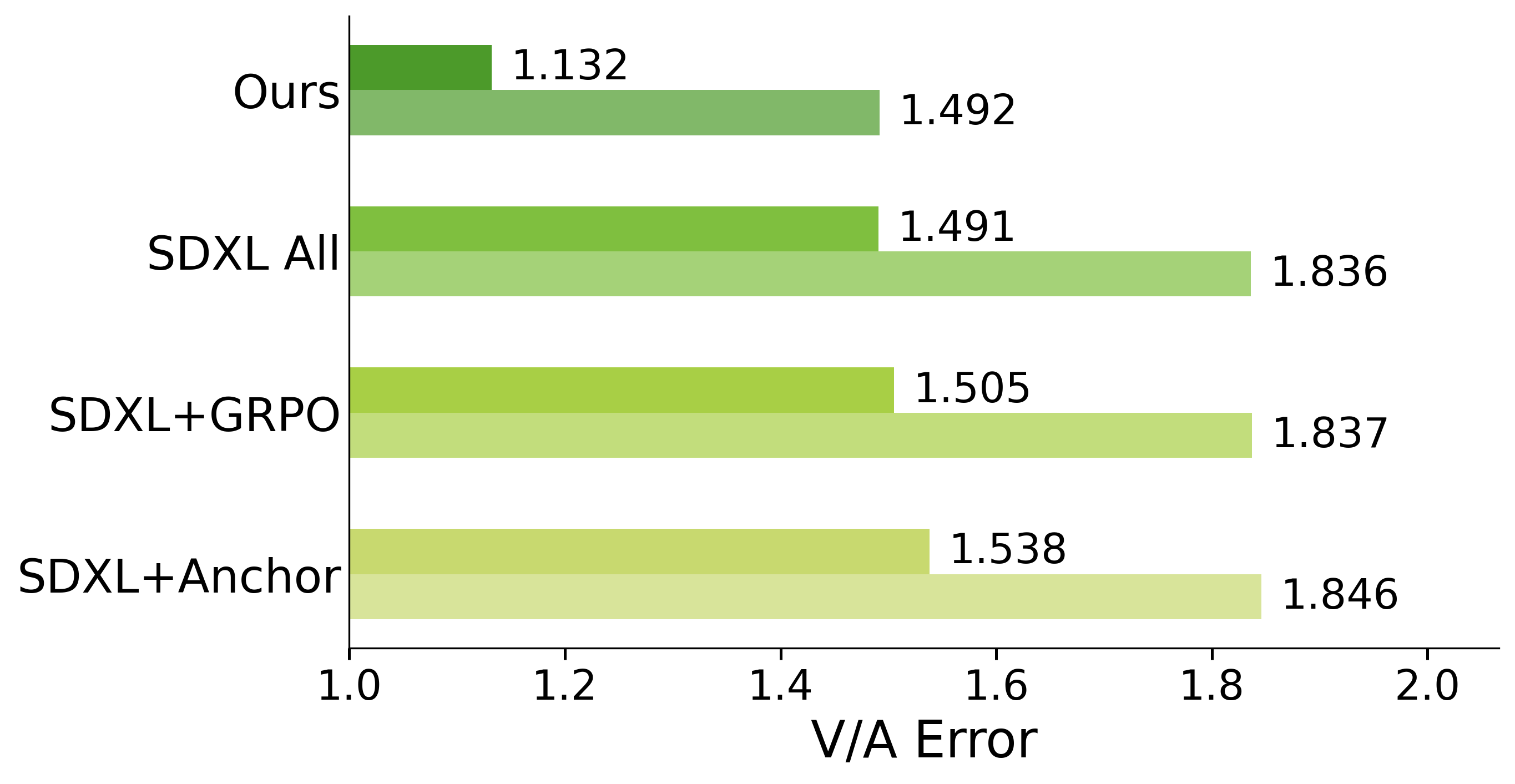}
        \vspace{2mm}
        \includegraphics[width=\linewidth]{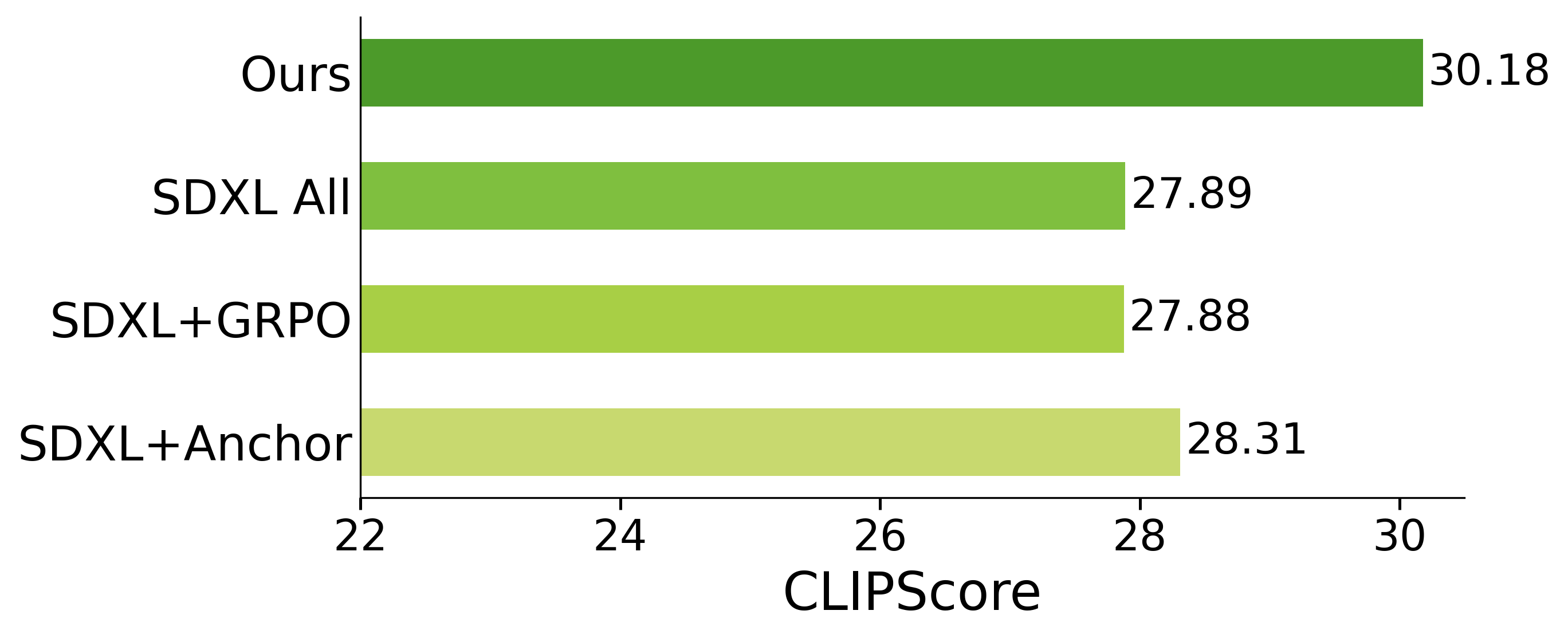}
        \caption{Variants on SDXL}
        \label{fig:ablation_group2}
    \end{subfigure}
    \hfill
    \begin{subfigure}[t]{0.32\textwidth}
        \centering
        \includegraphics[width=\linewidth]{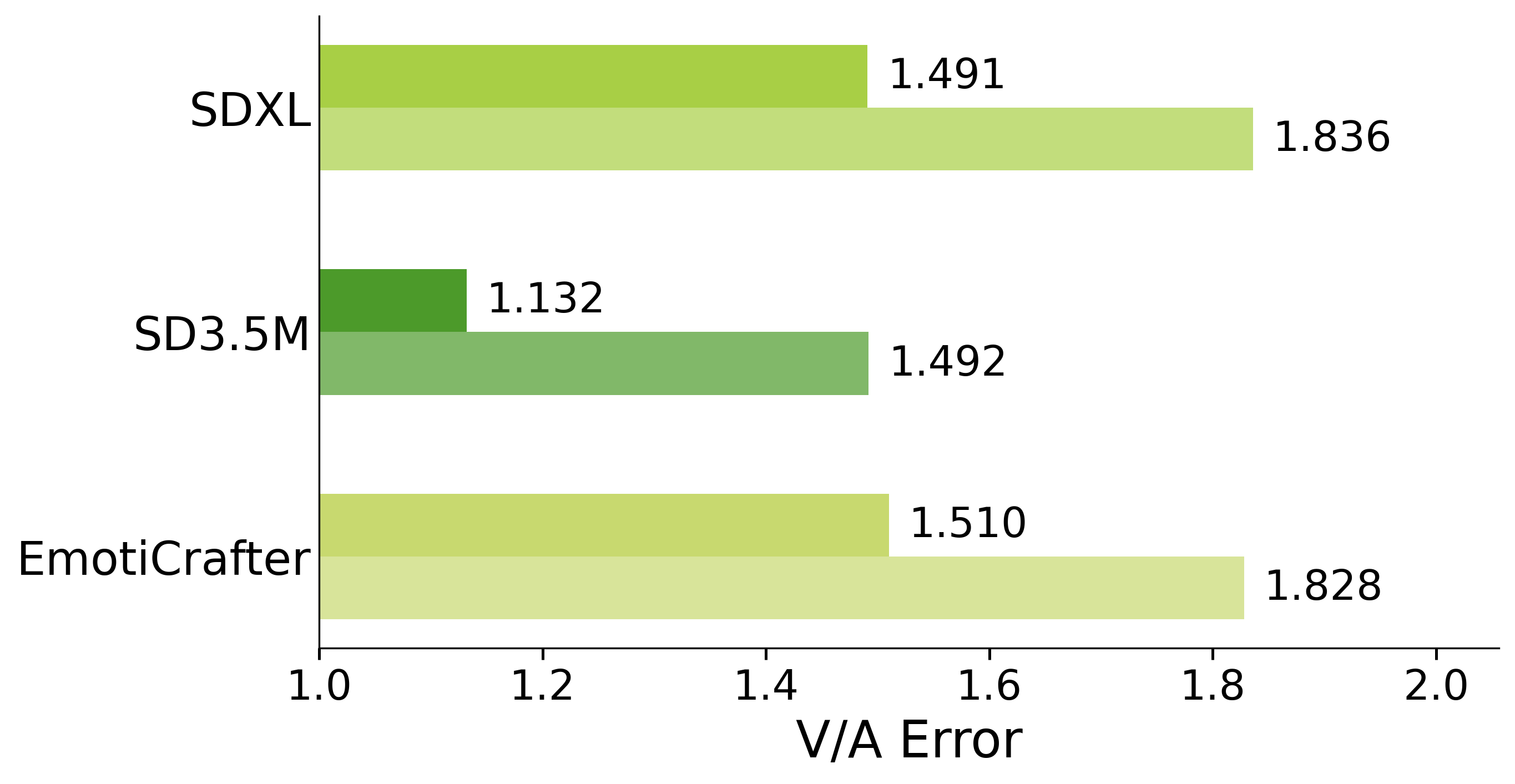}
        \vspace{2mm}
        \includegraphics[width=\linewidth]{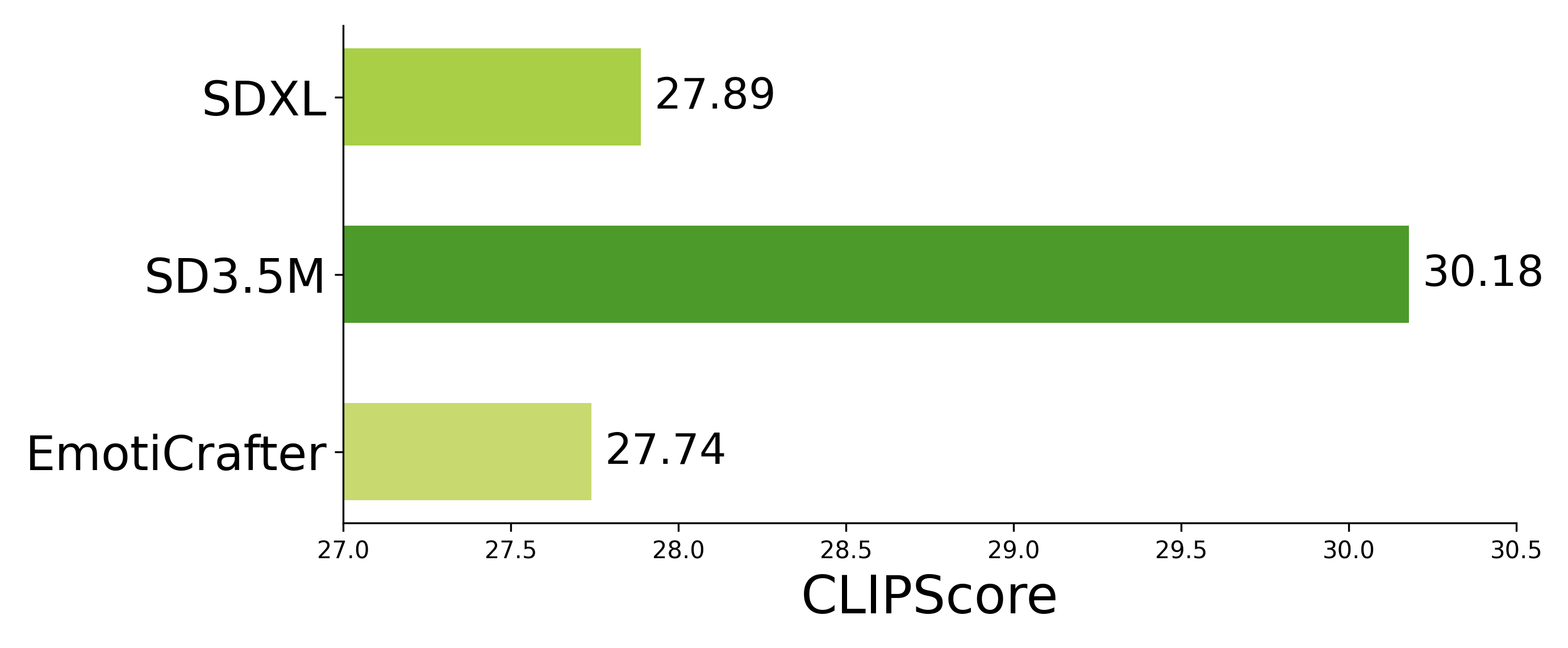}
        \caption{Baseline comparison}
        \label{fig:ablation_group3}
    \end{subfigure}
    \caption{Ablation study on diverse backbones. Each group shows V-Err and A-Err on the top, and CLIPScore on the bottom. (a) On SD3.5M, Anchor only and GRPO only each improve one aspect but not both; their combination (Ours) achieves the best balance. (b) All SDXL variants underperform our full method. (c) Our method substantially outperforms both EmotiCrafter and SDXL All.}
    \label{fig:ablation_all}
\end{figure*}

\begin{figure*}[t]
    \centering
    \begin{minipage}{0.48\textwidth}
        \centering
        \subcaption{Baseline (EmotiCrafter)}
        \label{fig:baseline_grid}
        \begin{tabular}{ccc}
            \includegraphics[width=0.3\linewidth]{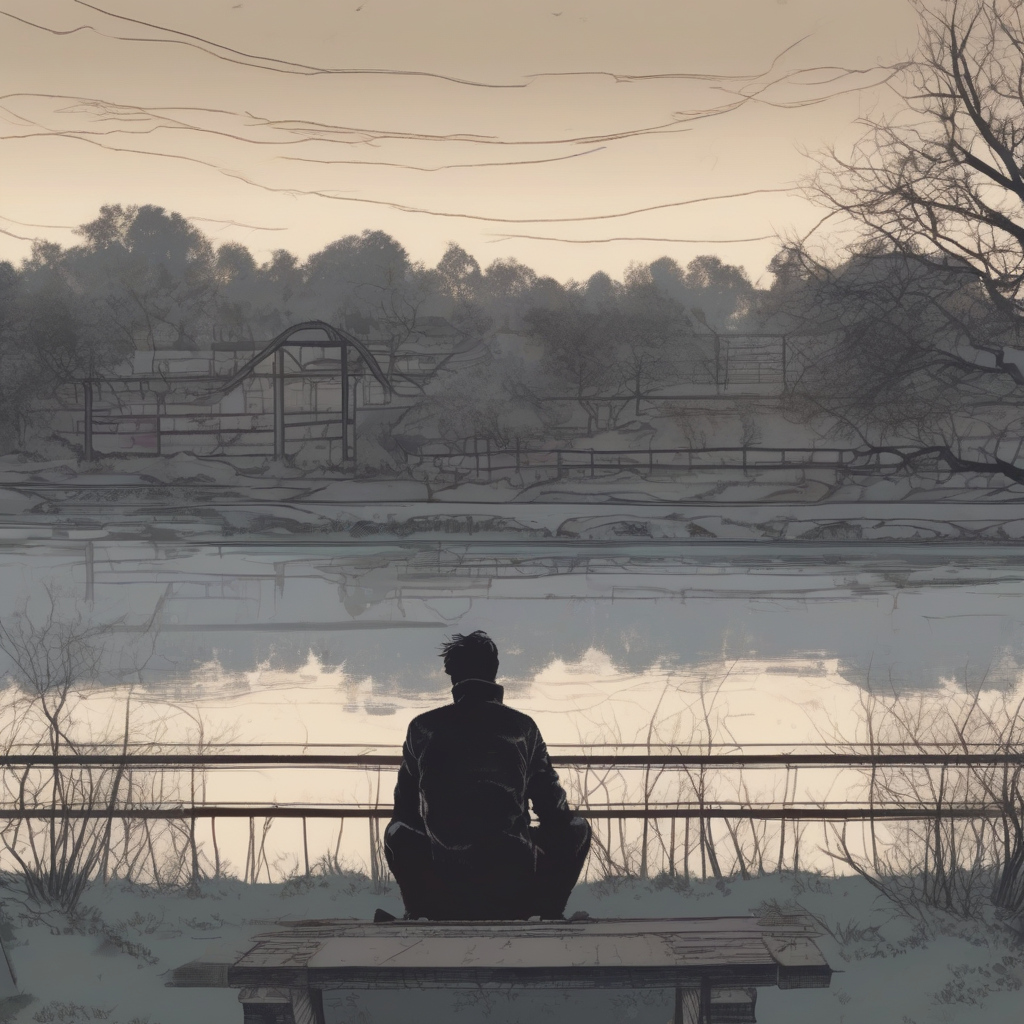} &
            \includegraphics[width=0.3\linewidth]{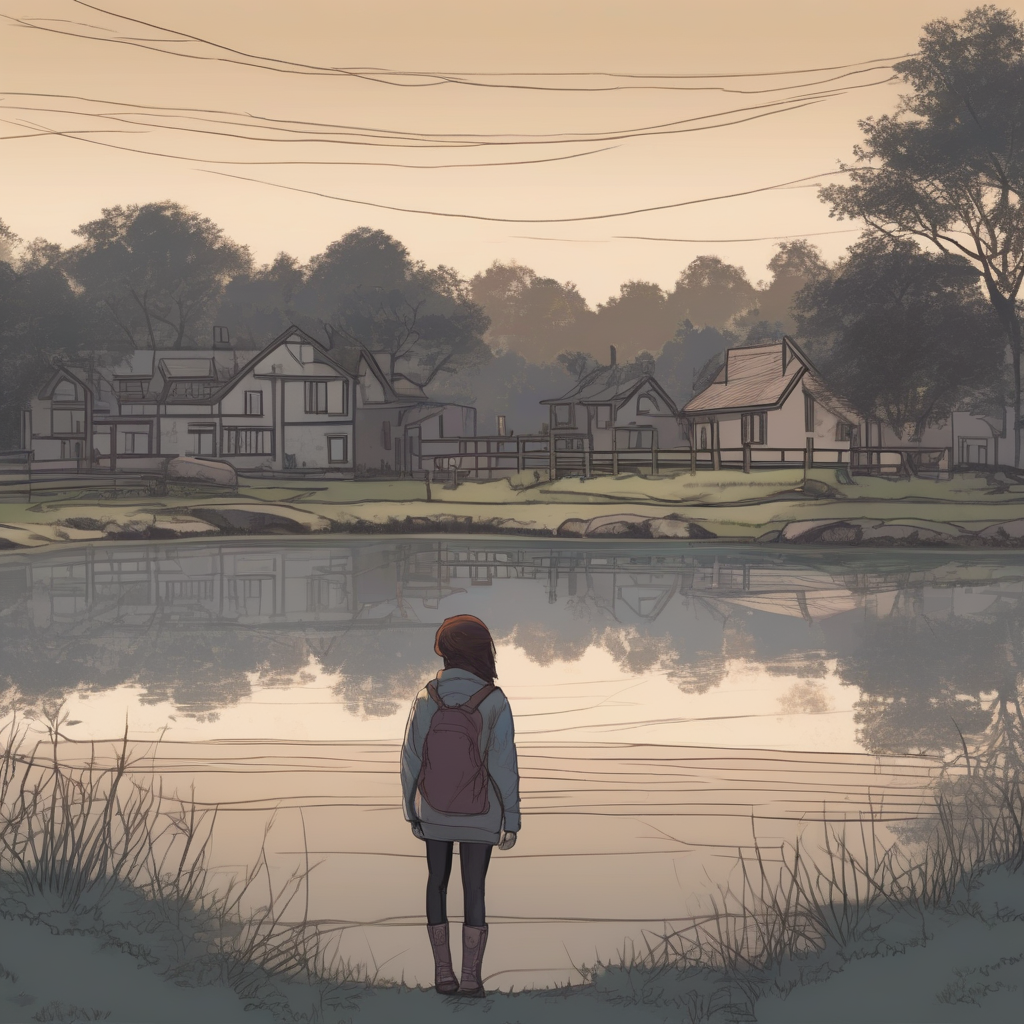} &
            \includegraphics[width=0.3\linewidth]{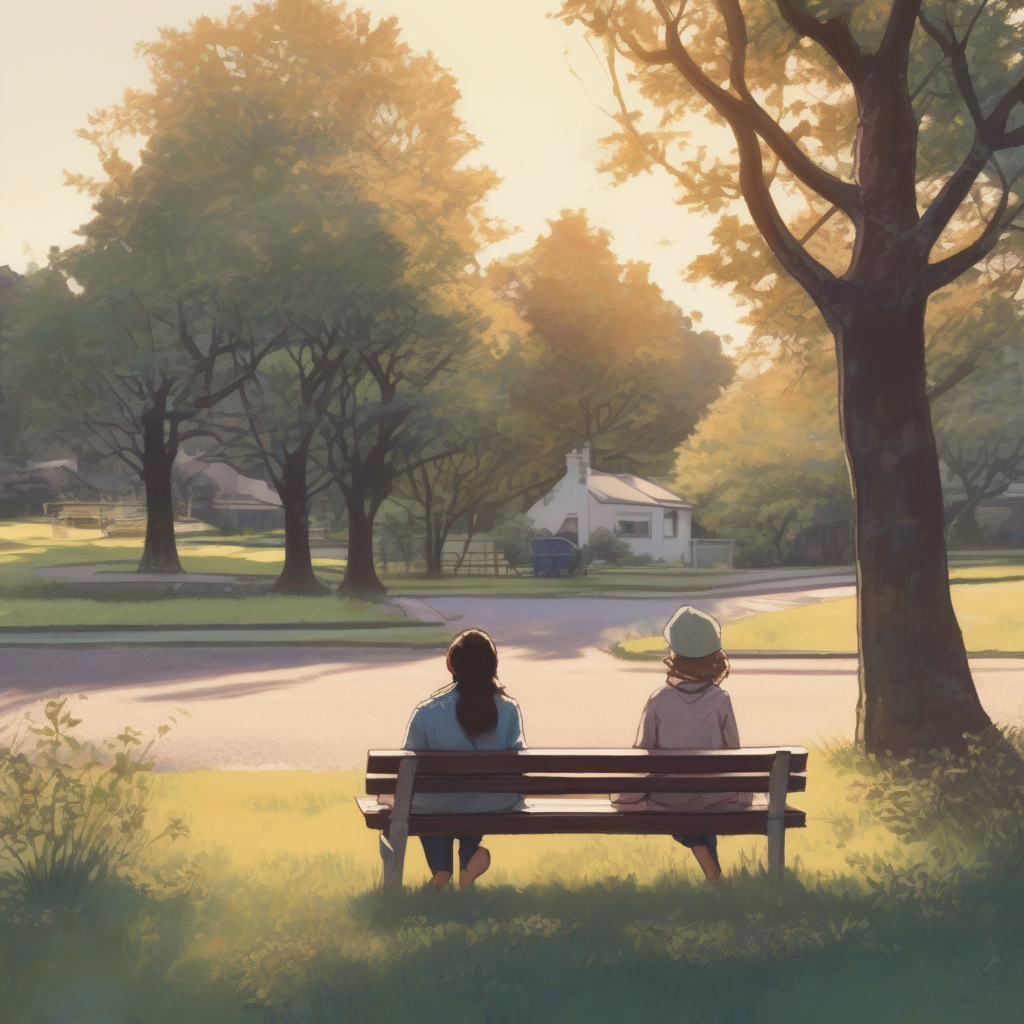} \\
            \includegraphics[width=0.3\linewidth]{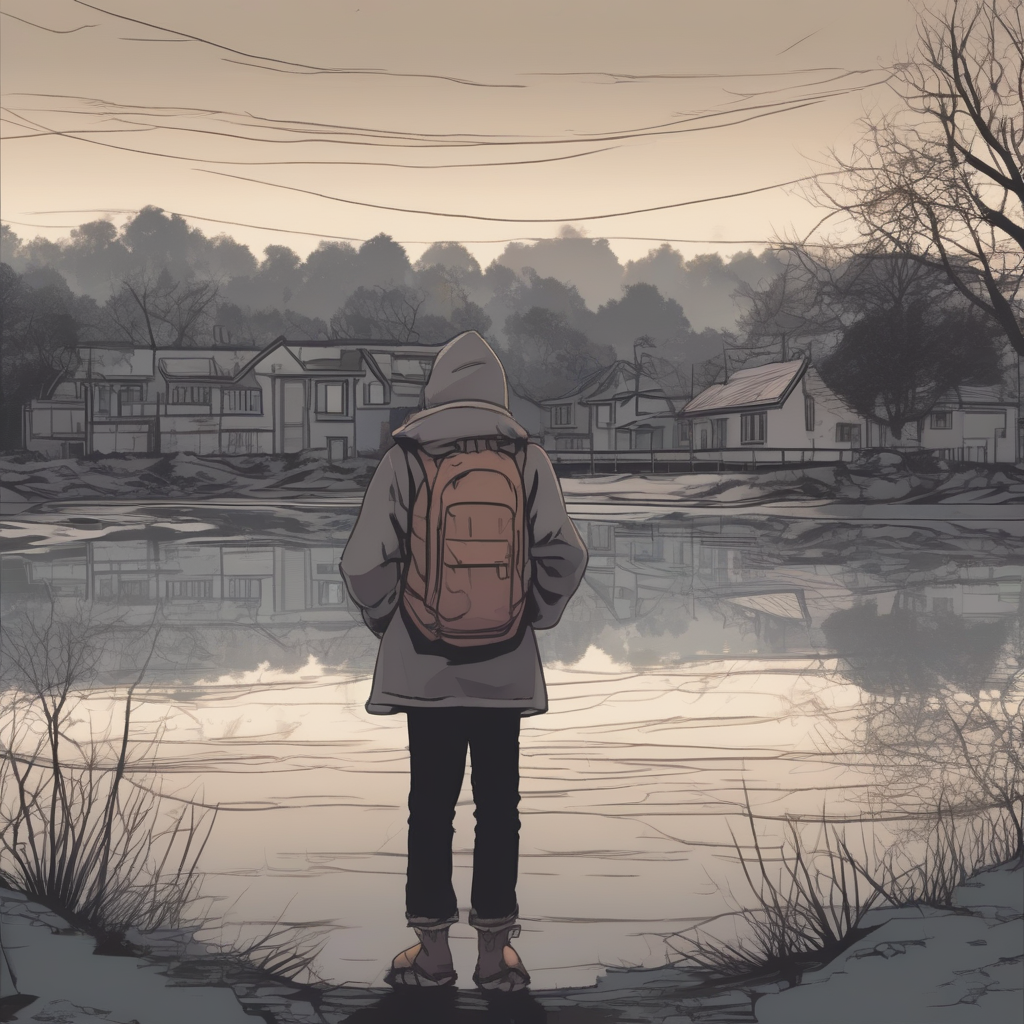} &
            \includegraphics[width=0.3\linewidth]{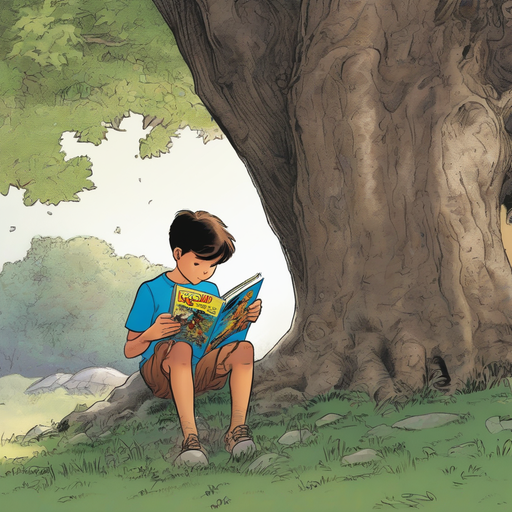} &
            \includegraphics[width=0.3\linewidth]{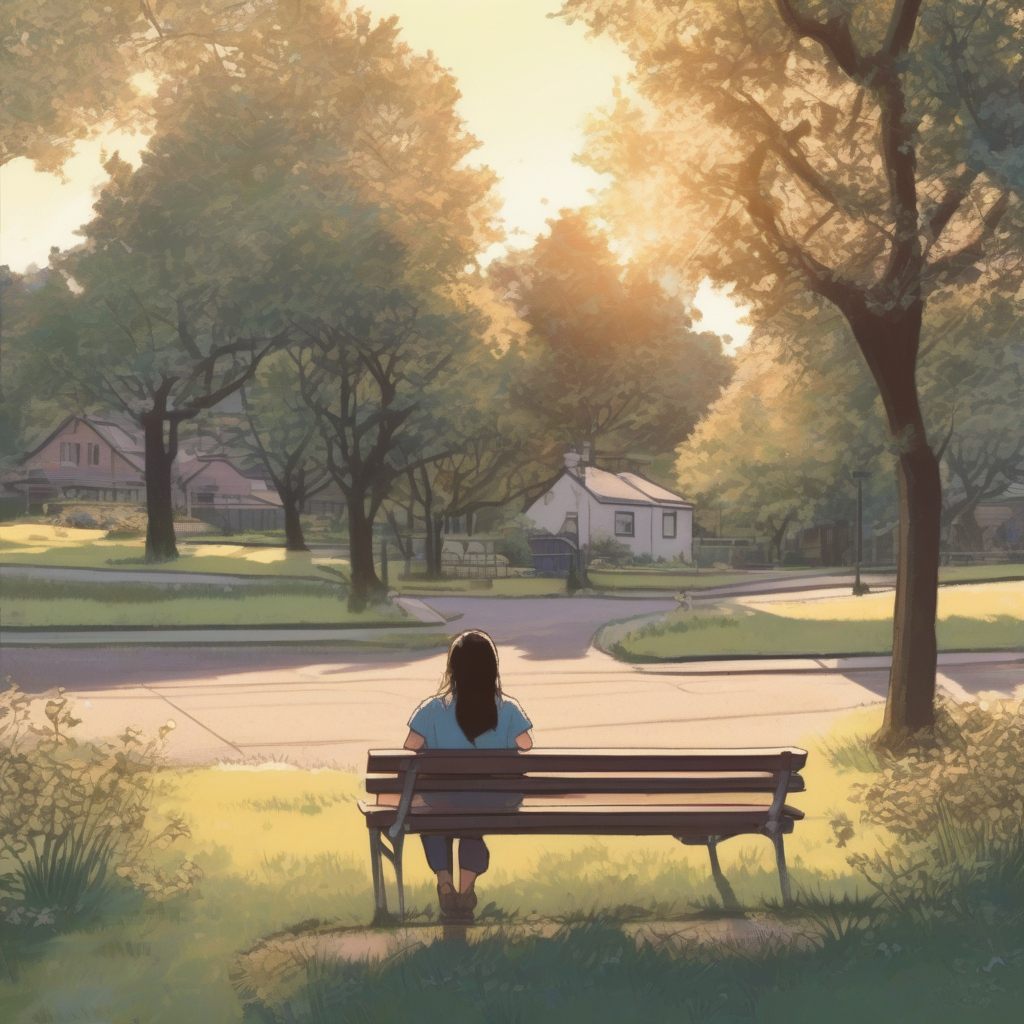} \\
            \includegraphics[width=0.3\linewidth]{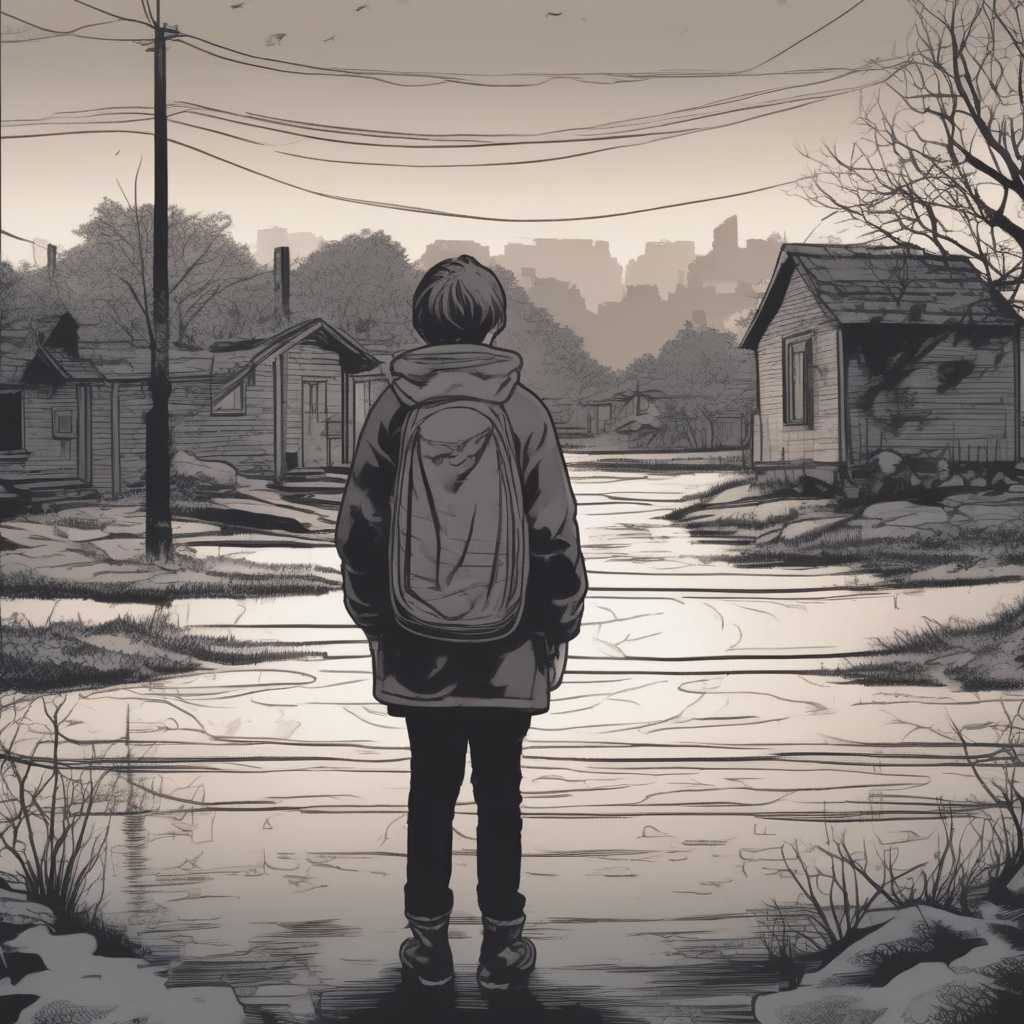} &
            \includegraphics[width=0.3\linewidth]{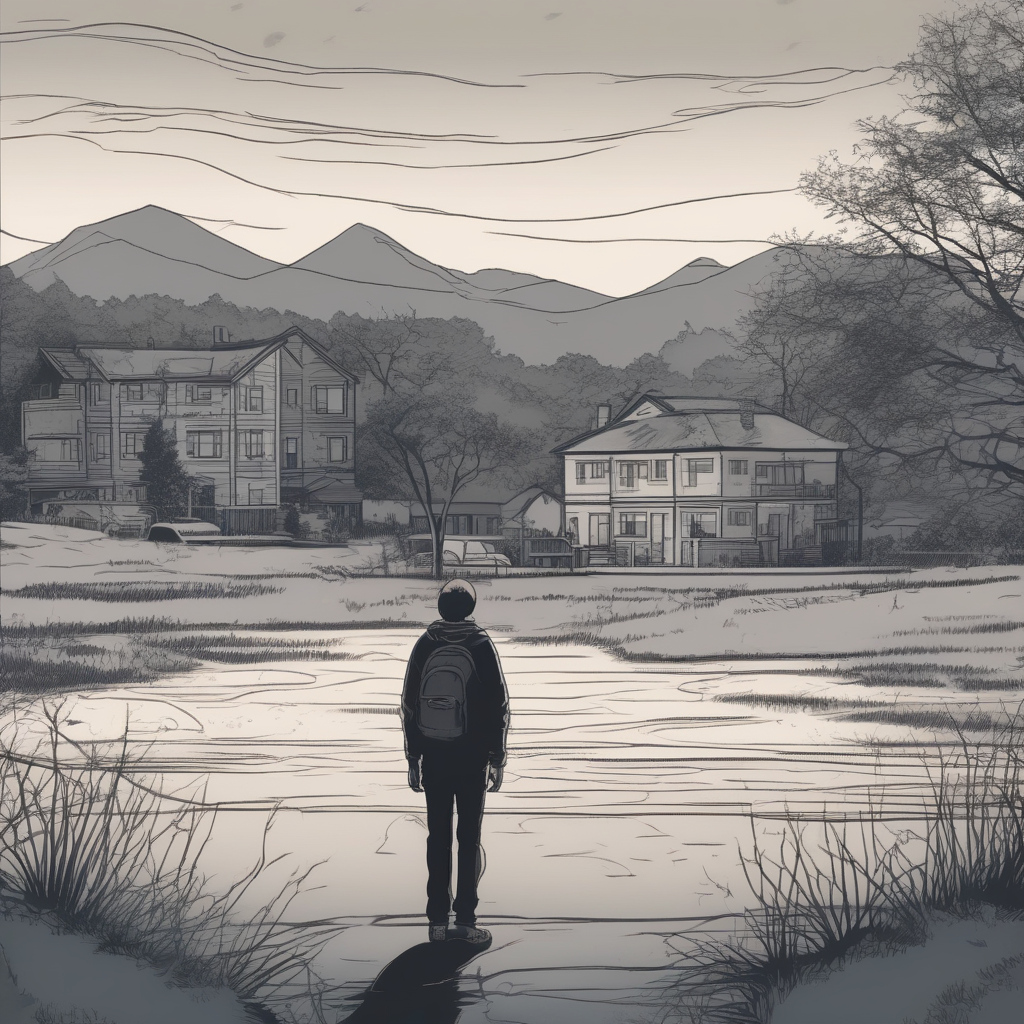} &
            \includegraphics[width=0.3\linewidth]{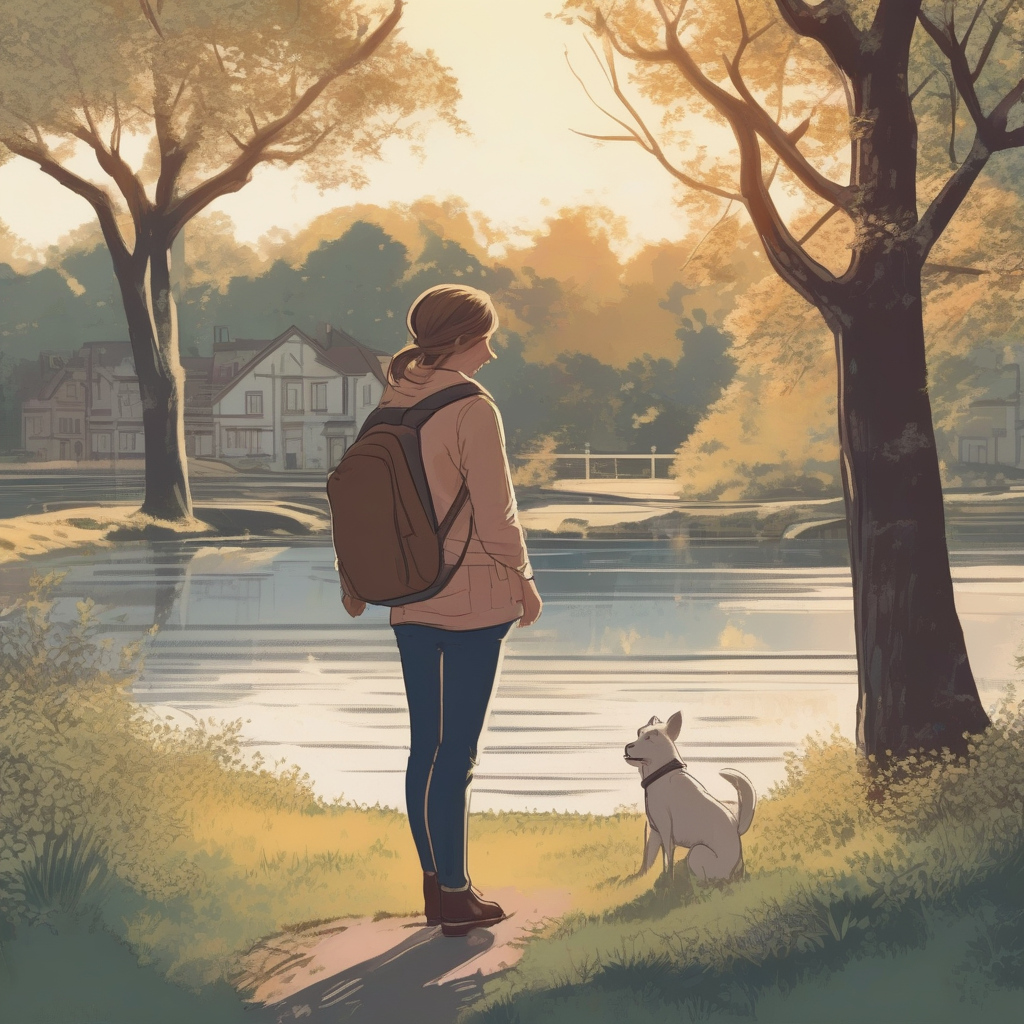} \\
        \end{tabular}
    \end{minipage}
    \hfill
    \begin{minipage}{0.48\textwidth}
        \centering
        \subcaption{Ours (Anchor-Constrained GRPO)}
        \label{fig:ours_grid}
        \begin{tabular}{ccc}
            \includegraphics[width=0.3\linewidth]{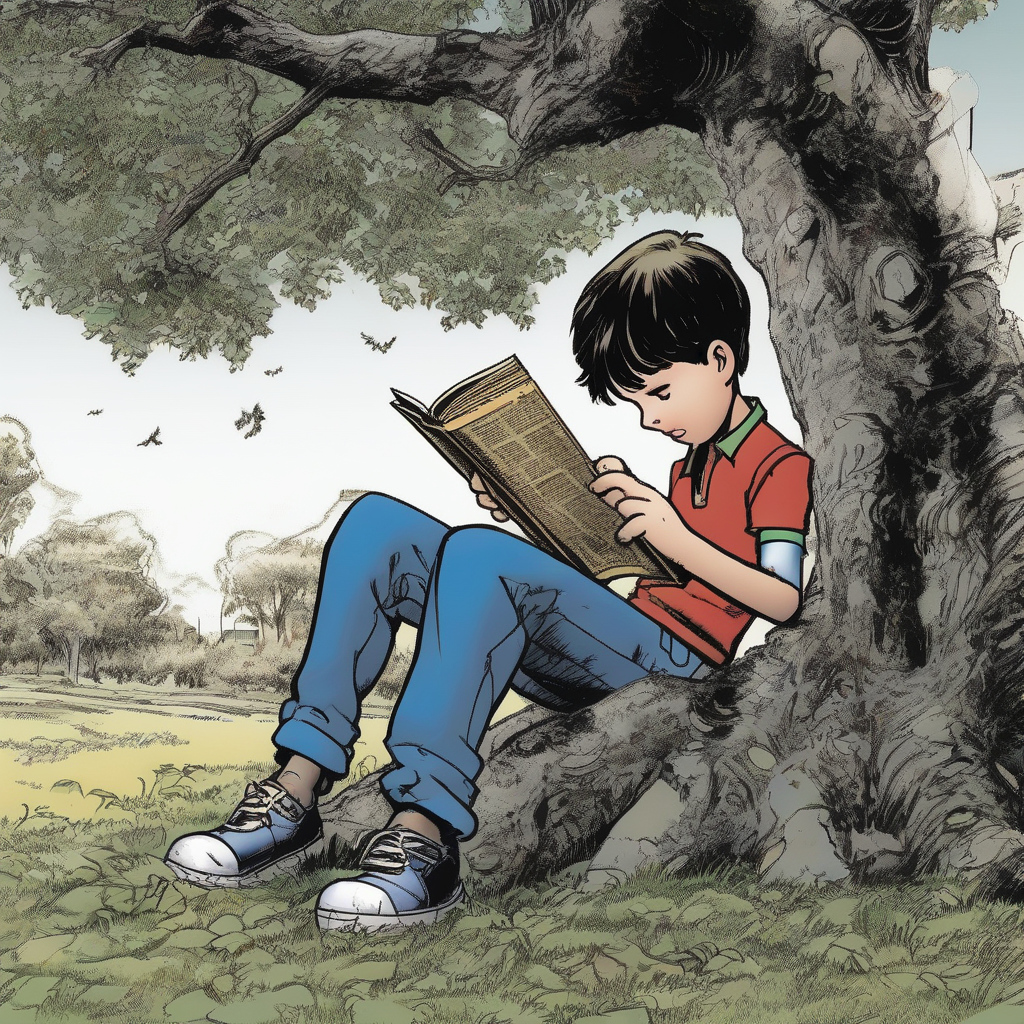}& \includegraphics[width=0.3\linewidth]{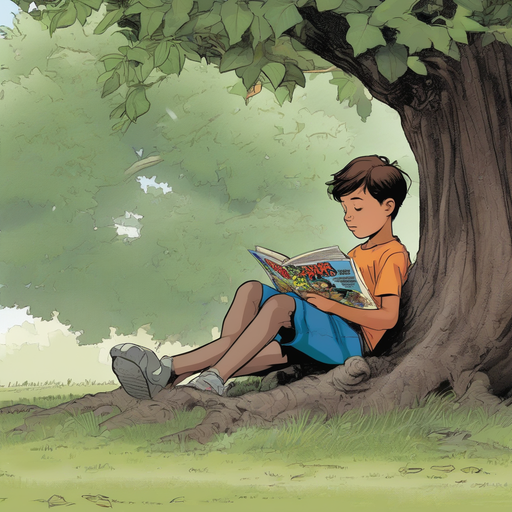}& \includegraphics[width=0.3\linewidth]{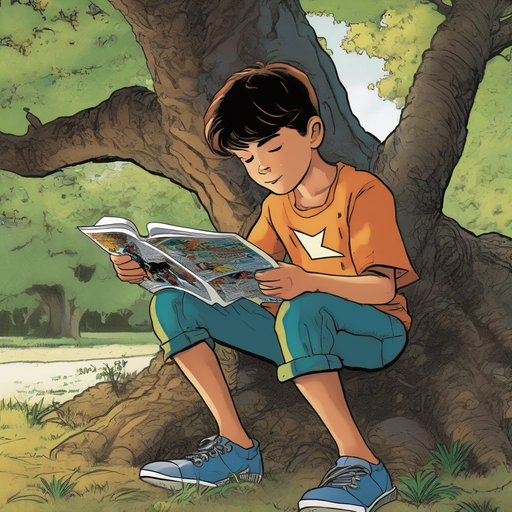}\\
            \includegraphics[width=0.3\linewidth]{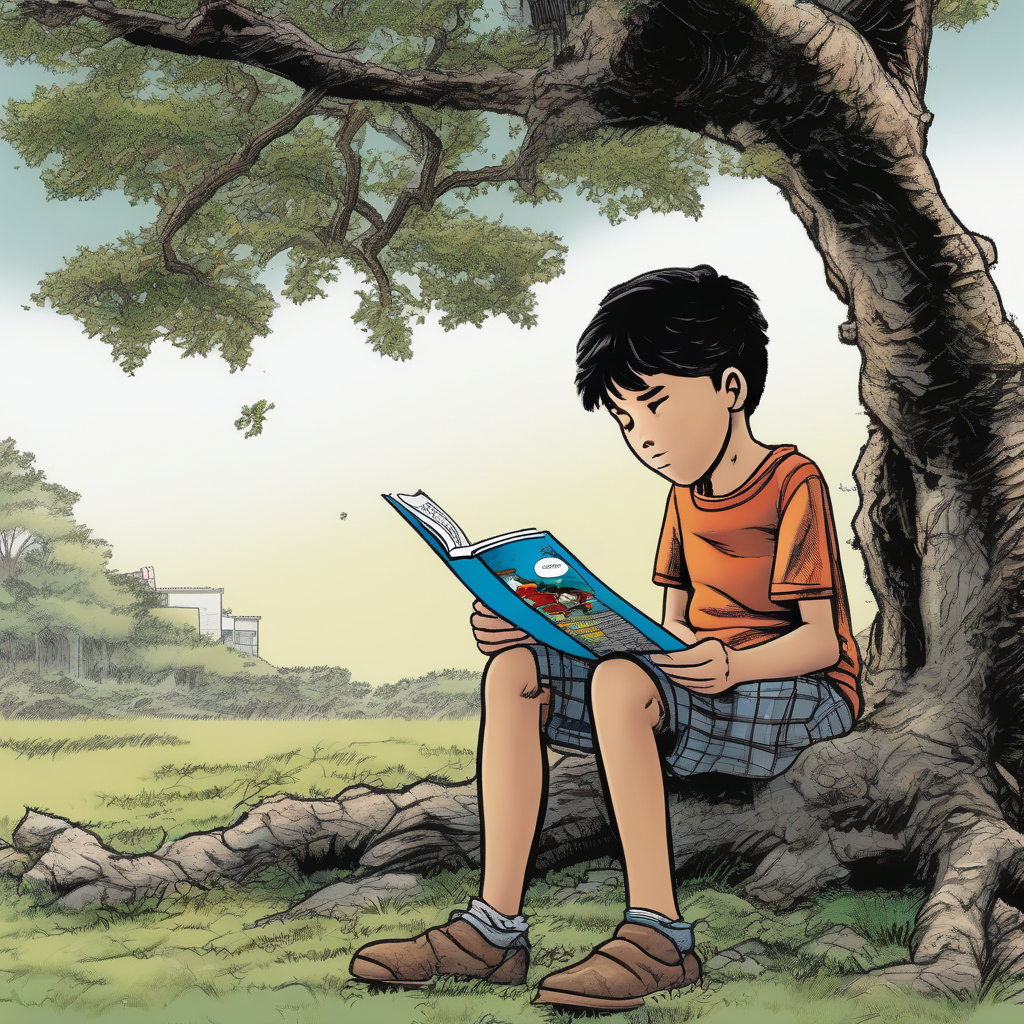}& \includegraphics[width=0.3\linewidth]{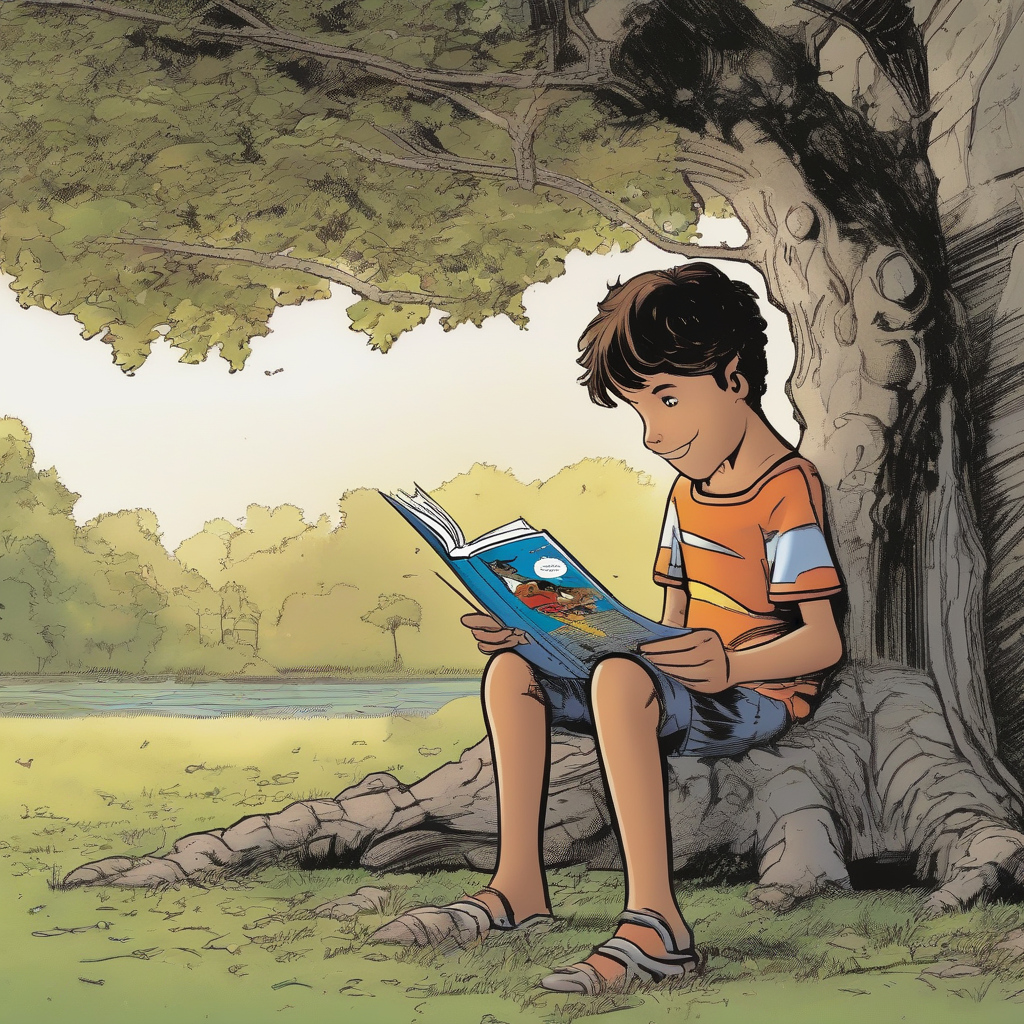}& \includegraphics[width=0.3\linewidth]{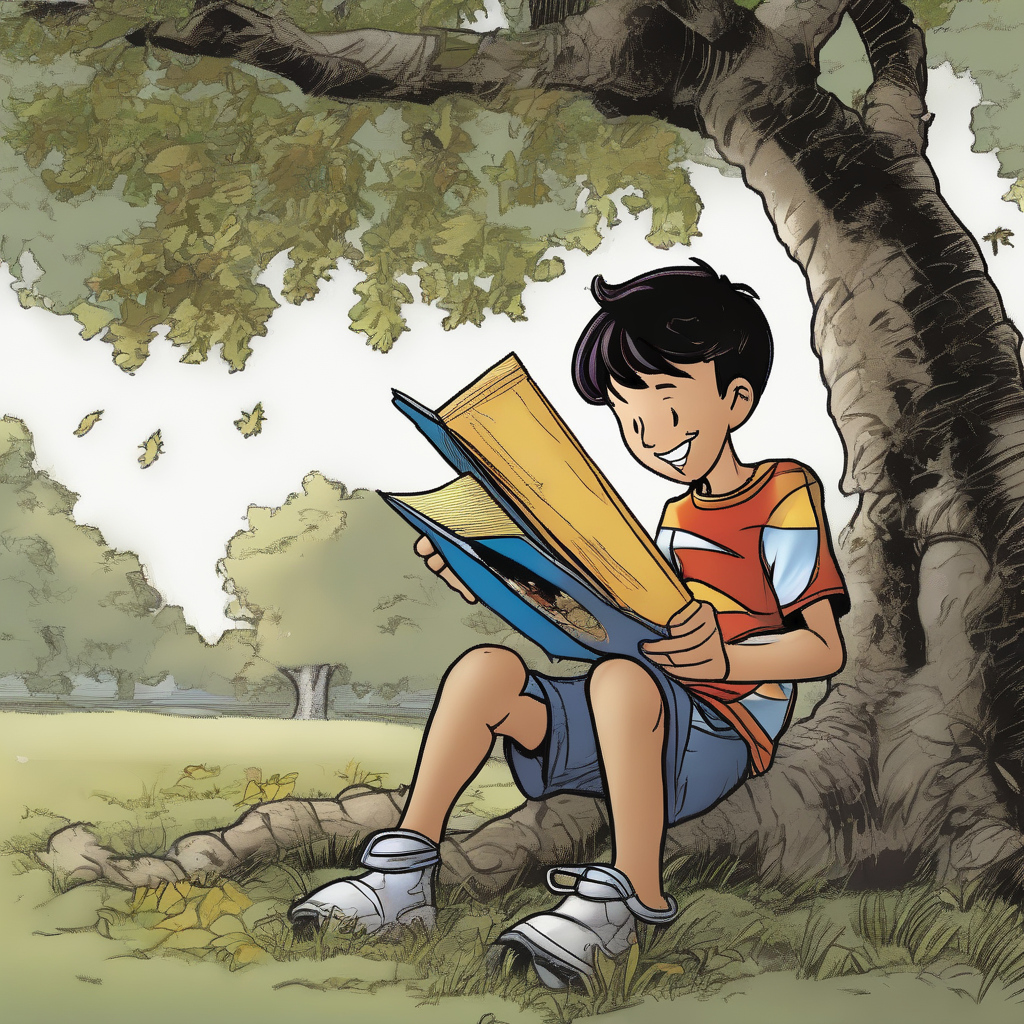}\\
            \includegraphics[width=0.3\linewidth]{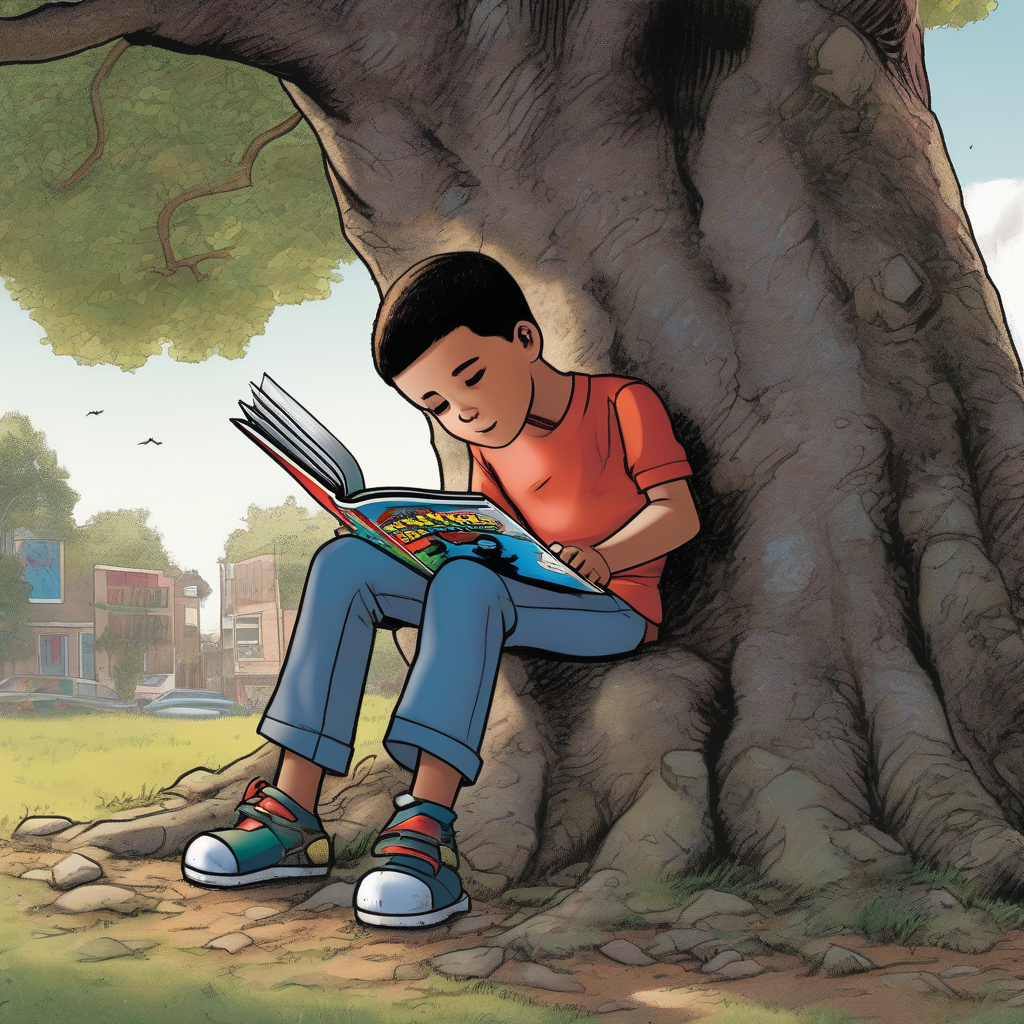}& \includegraphics[width=0.3\linewidth]{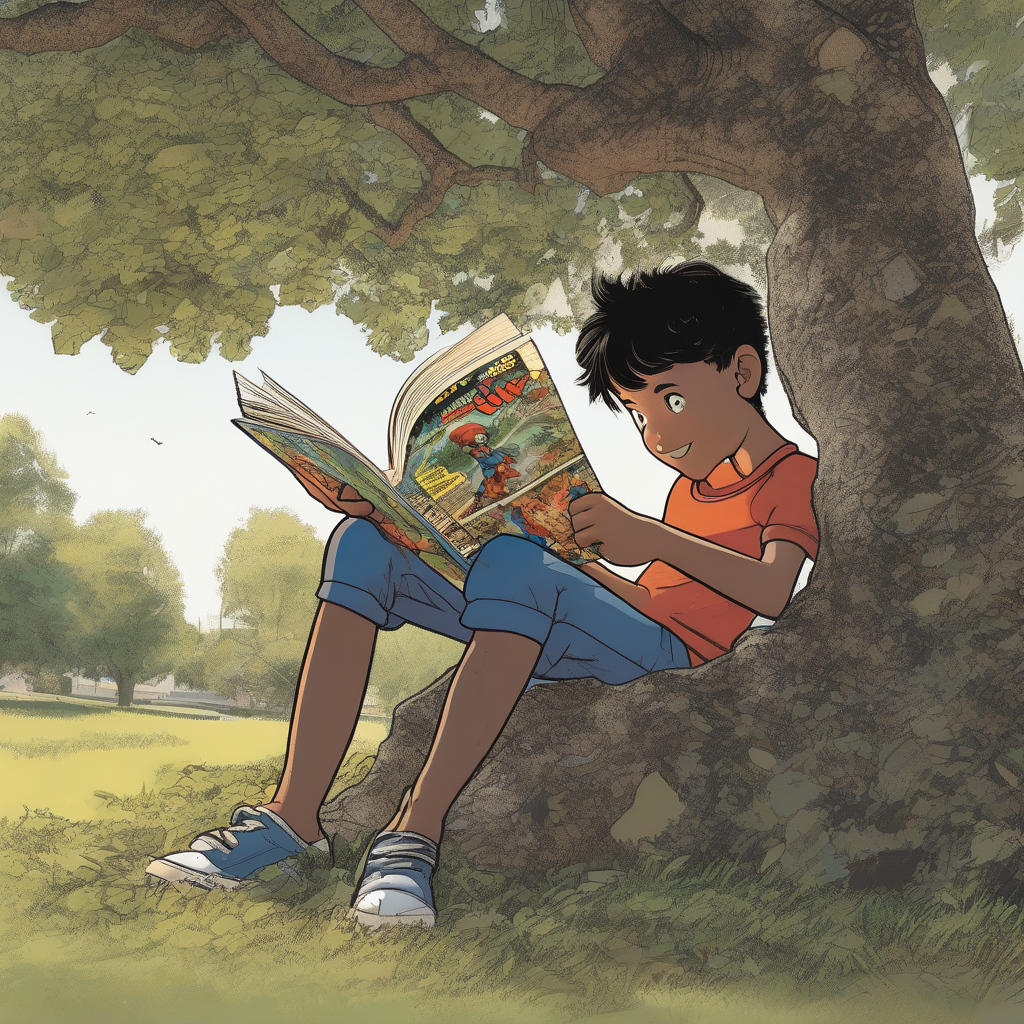}& \includegraphics[width=0.3\linewidth]{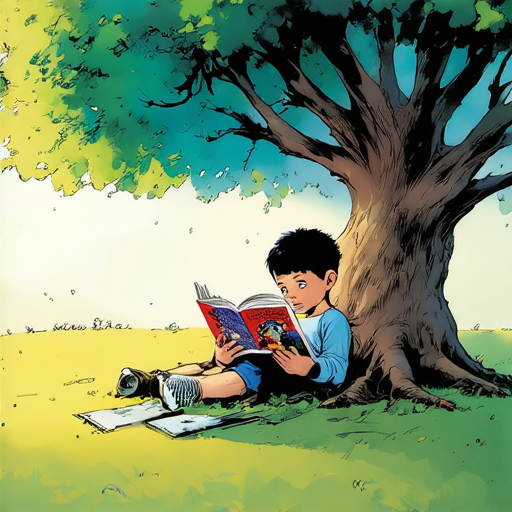}\\
        \end{tabular}
    \end{minipage}
    \caption{Emotion variation across the valence-arousal (VA) space for the prompt ``A boy reading a comic book under a tree.'' Left: baseline method (EmotiCrafter) suffers from severe emotion-semantic drift. 
    Right: our anchor-constrained GRPO method preserves all key elements (boy, tree, comic book) while smoothly adjusting emotional attributes such as lighting, color saturation, and facial expression. The book cover remains legible, and the background stays consistent across VA values, demonstrating effective content preservation.}
    \label{fig:emotion_grid}
\end{figure*}

\begin{figure*}[t]
    \centering
    \begin{minipage}{0.48\textwidth}
        \centering
        \subcaption{Baseline (EmotiCrafter)}
        \label{fig:baseline_grid_landscape}
        \begin{tabular}{ccc}
            \includegraphics[width=0.3\linewidth]{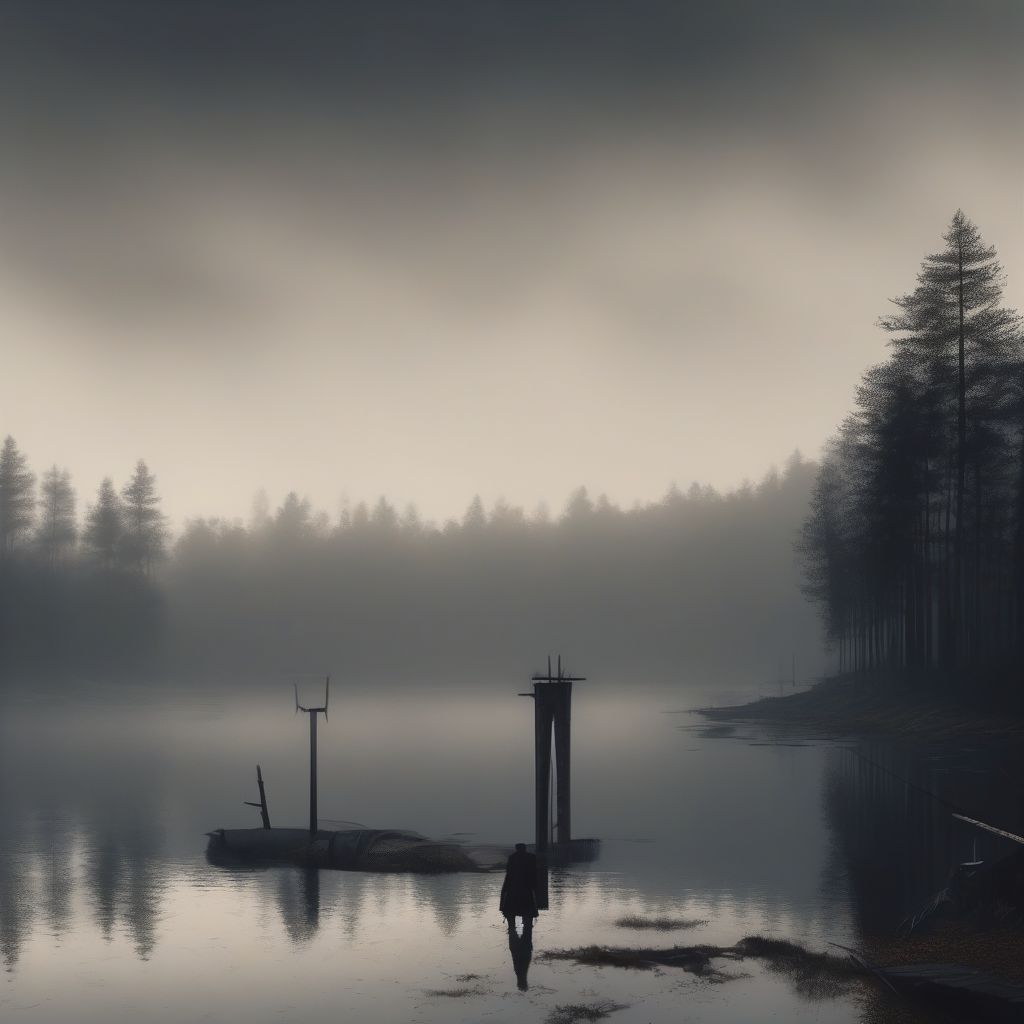} &
            \includegraphics[width=0.3\linewidth]{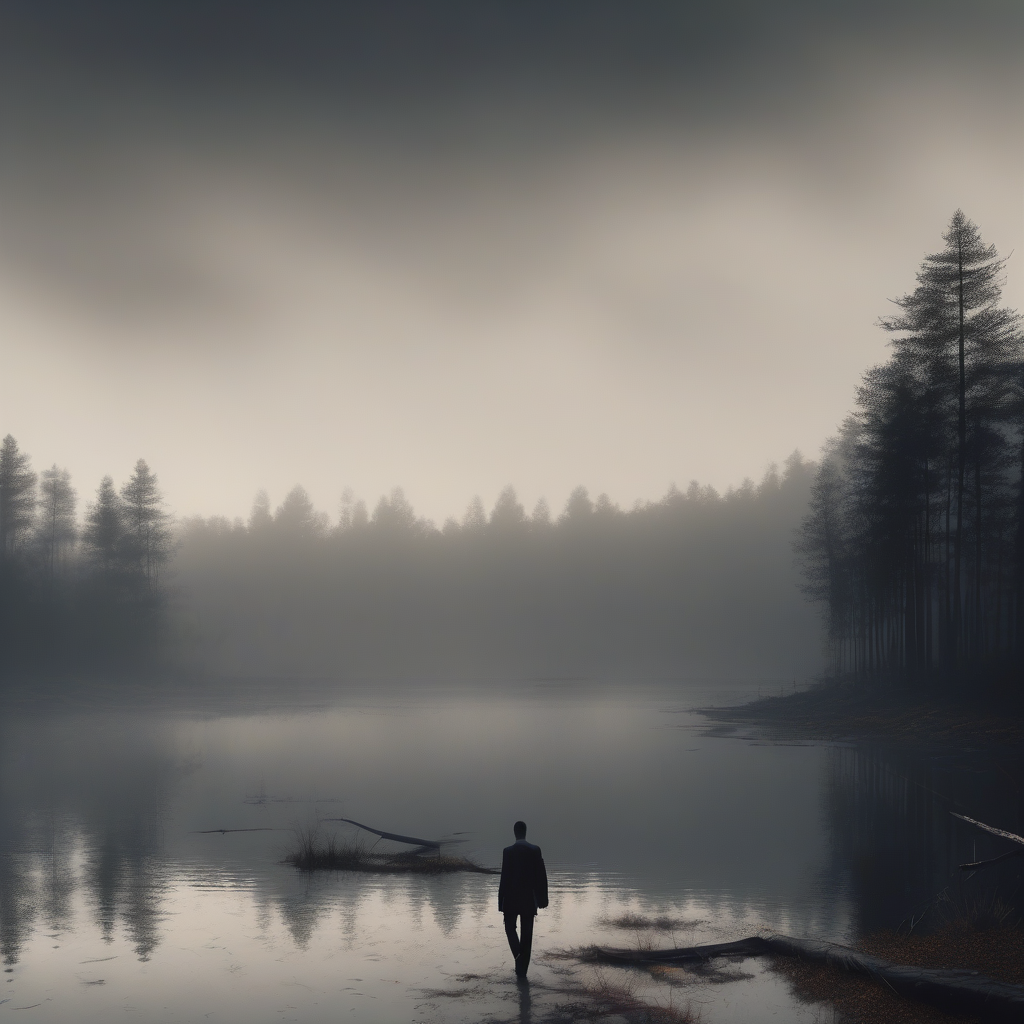} &
            \includegraphics[width=0.3\linewidth]{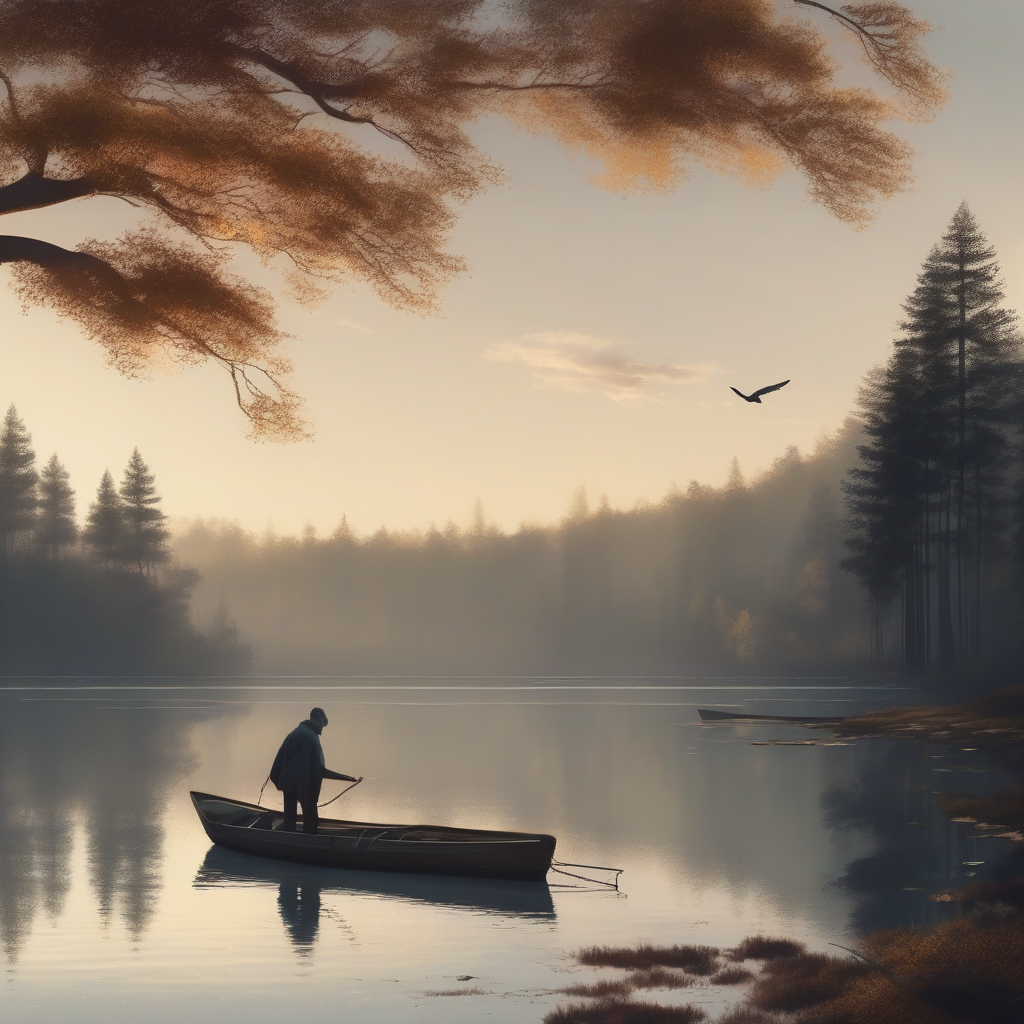} \\
            \includegraphics[width=0.3\linewidth]{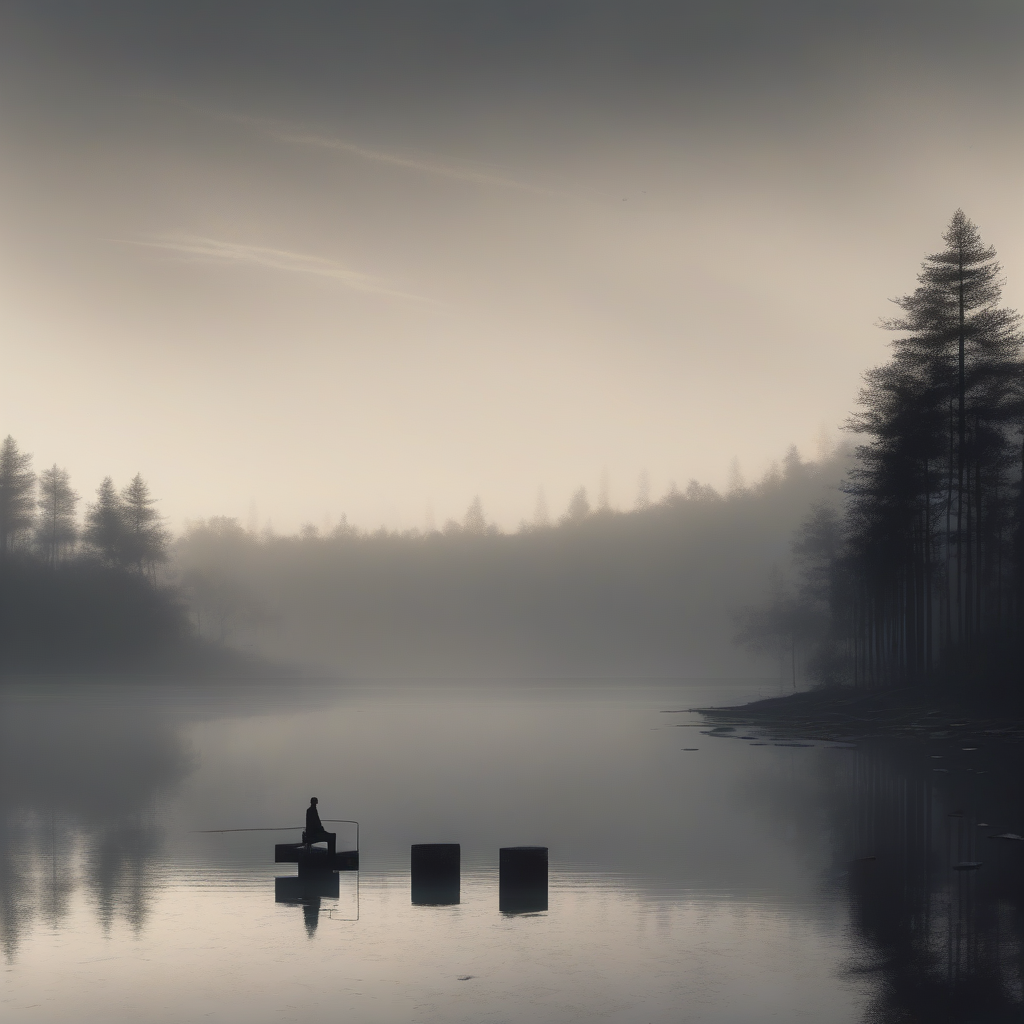} &
            \includegraphics[width=0.3\linewidth]{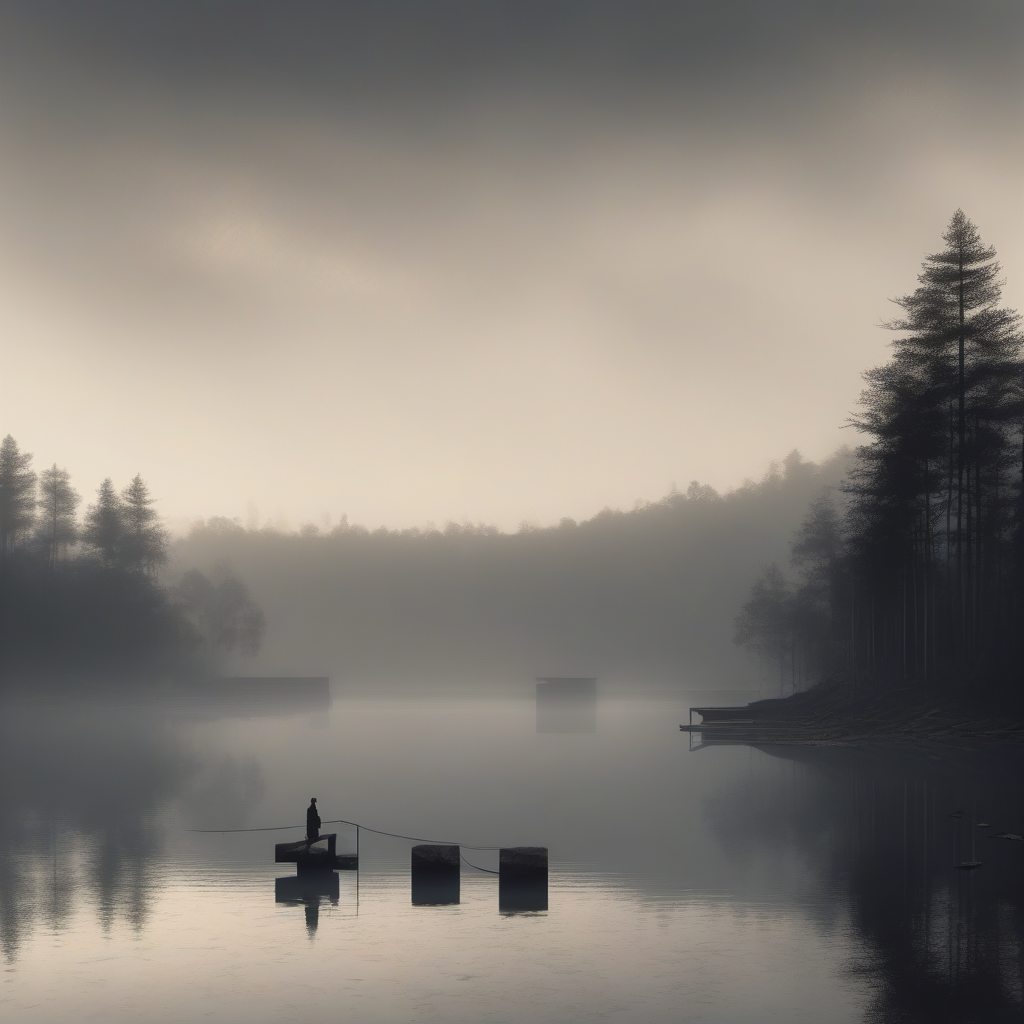} &
            \includegraphics[width=0.3\linewidth]{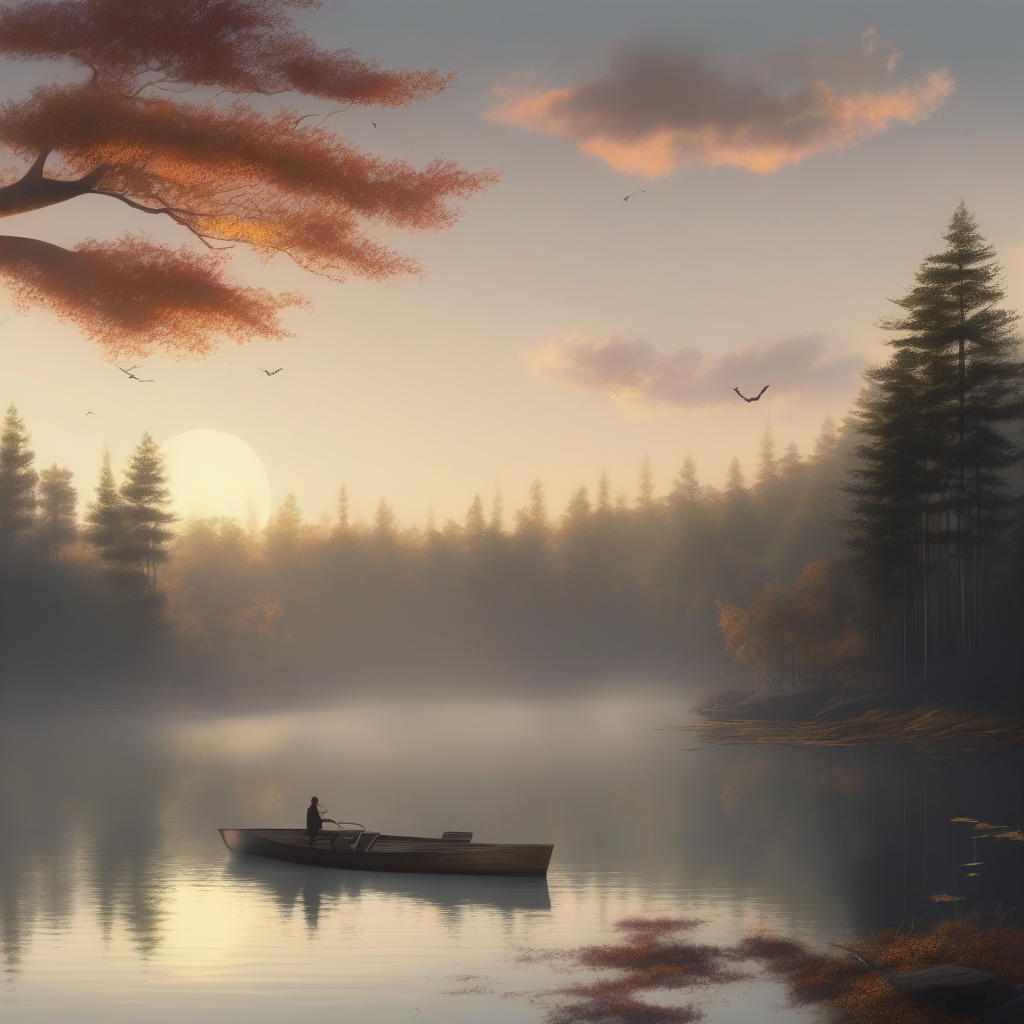} \\
            \includegraphics[width=0.3\linewidth]{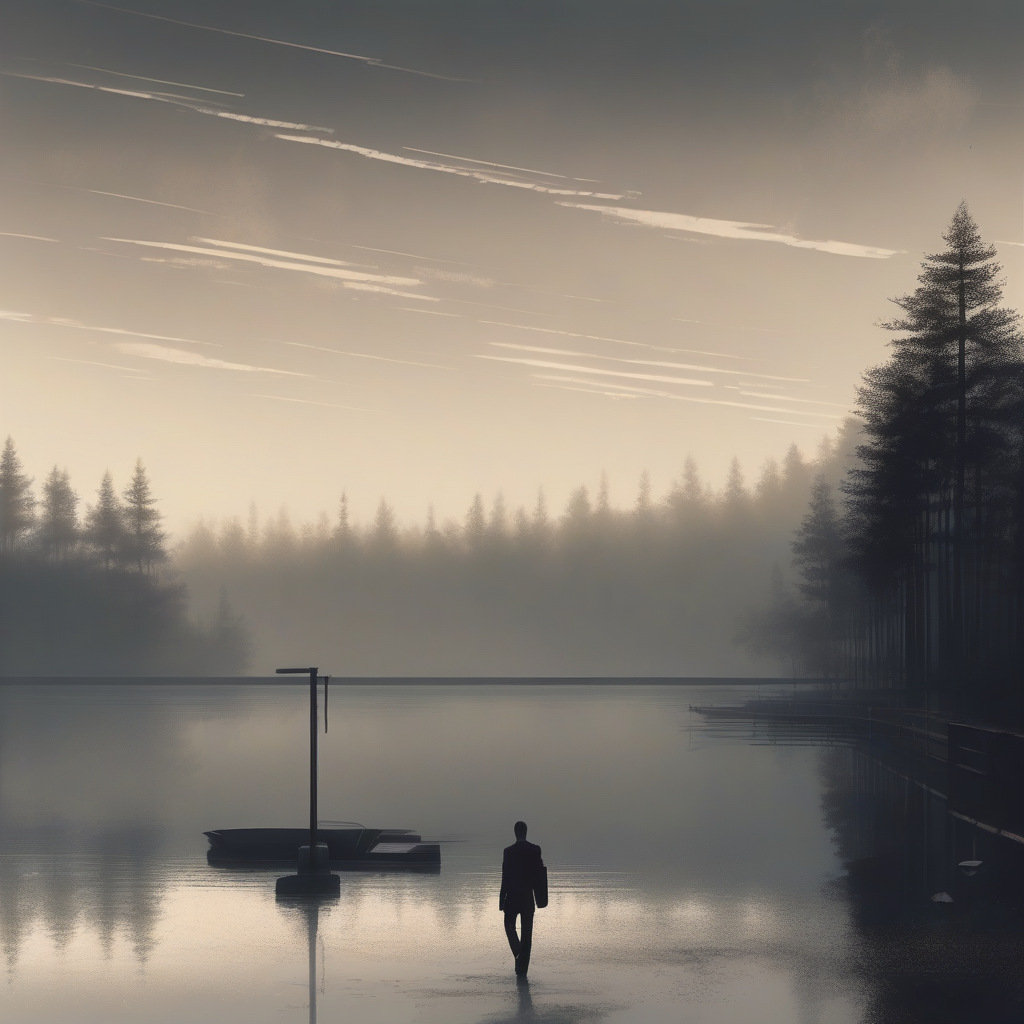} &
            \includegraphics[width=0.3\linewidth]{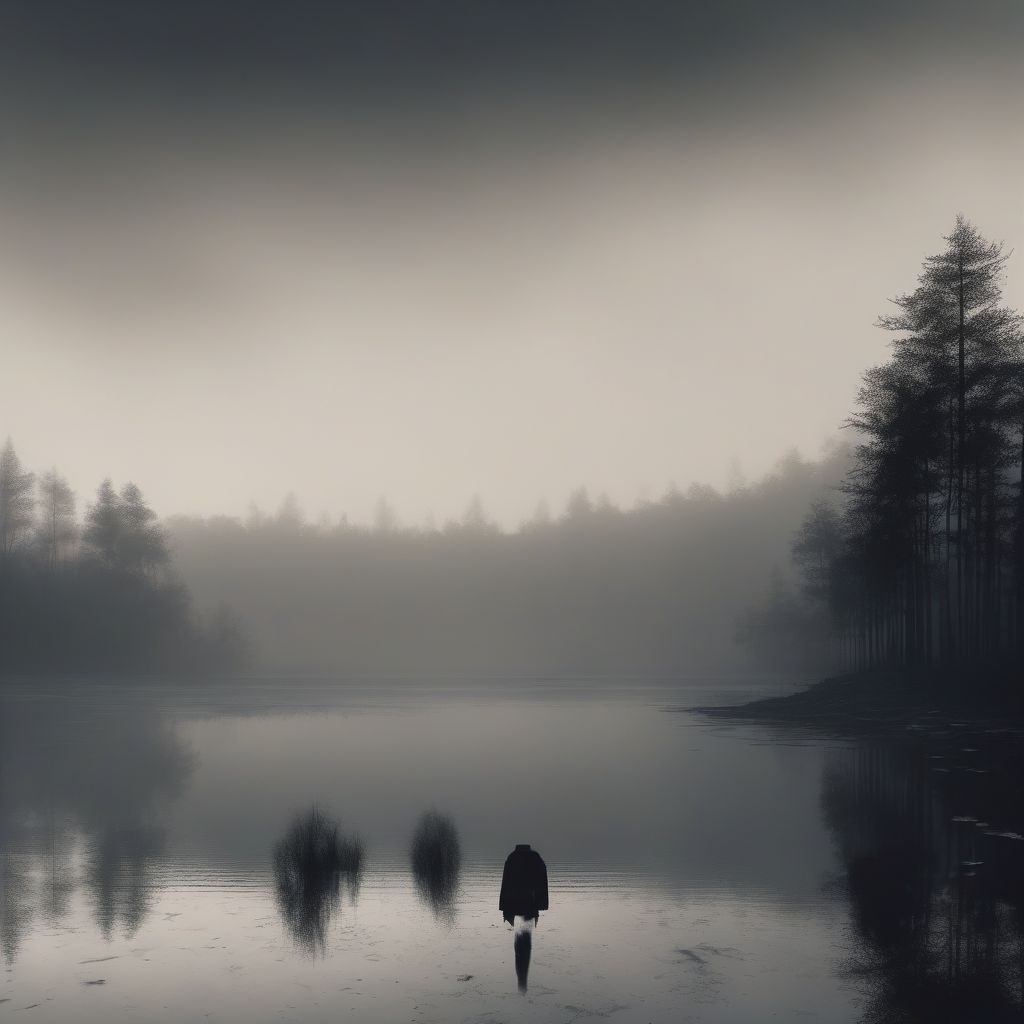} &
            \includegraphics[width=0.3\linewidth]{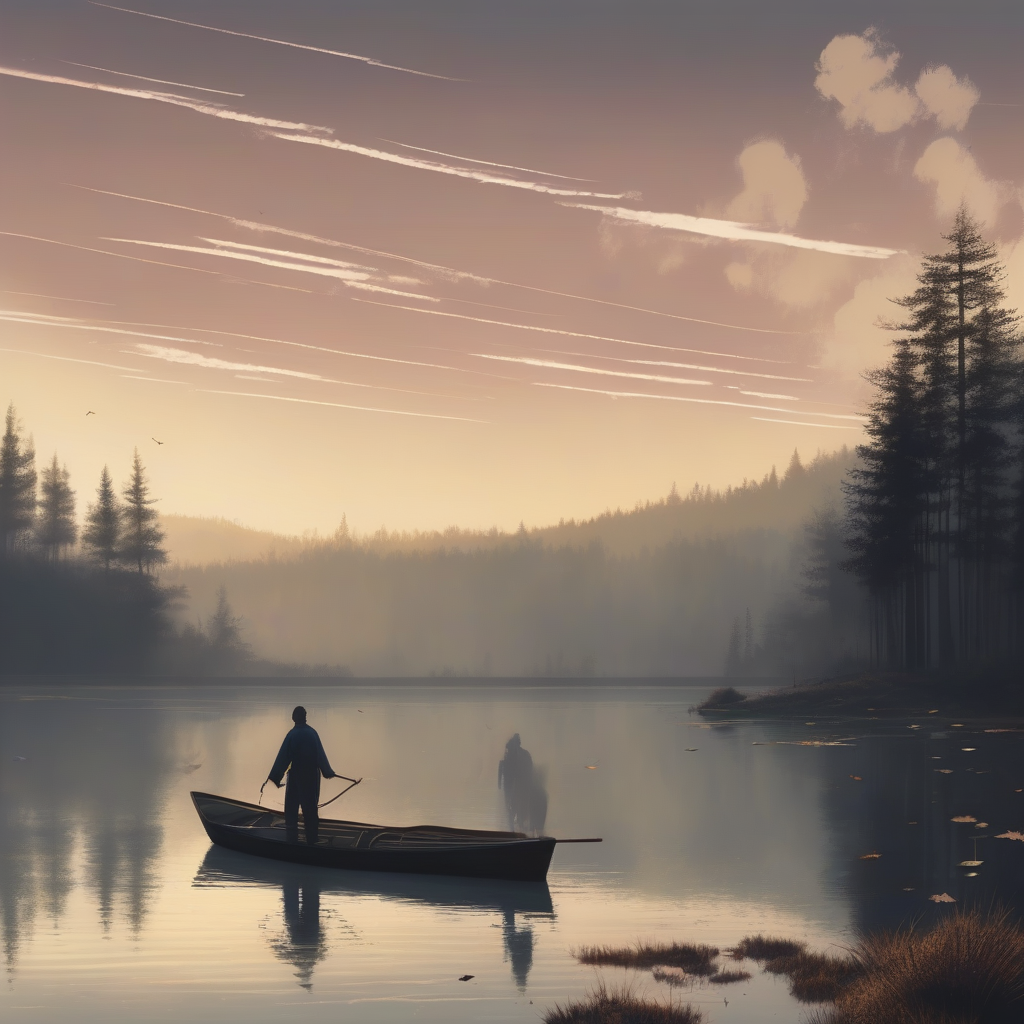} \\
        \end{tabular}
    \end{minipage}
    \hfill
    \begin{minipage}{0.48\textwidth}
        \centering
        \subcaption{Ours (Anchor-Constrained GRPO)}
        \label{fig:ours_grid_landscape}
        \begin{tabular}{ccc}
            \includegraphics[width=0.3\linewidth]{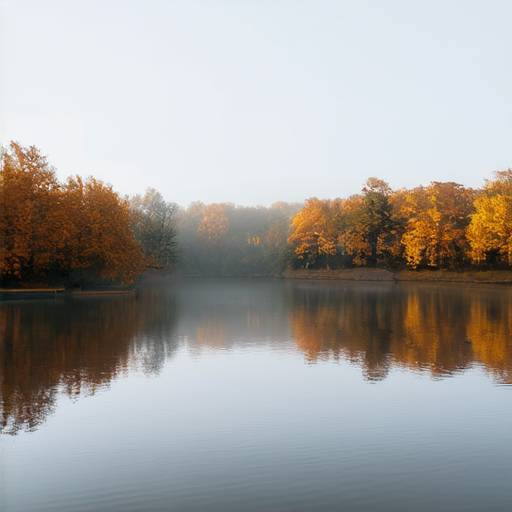} &
            \includegraphics[width=0.3\linewidth]{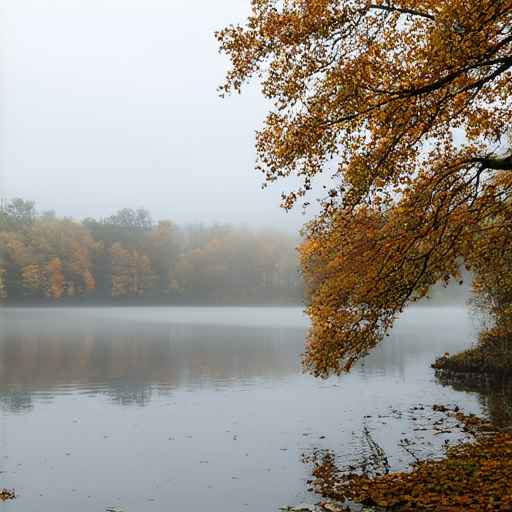} &
            \includegraphics[width=0.3\linewidth]{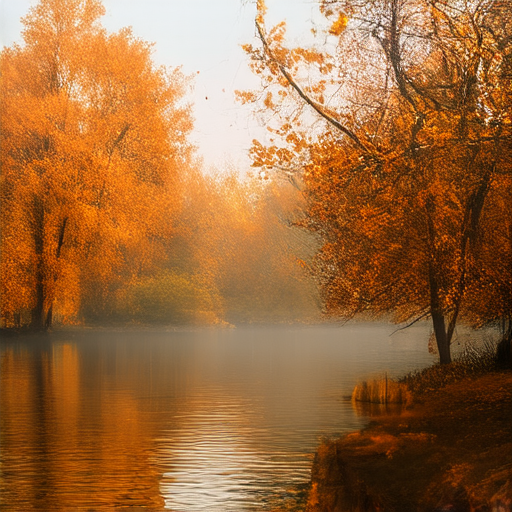} \\
            \includegraphics[width=0.3\linewidth]{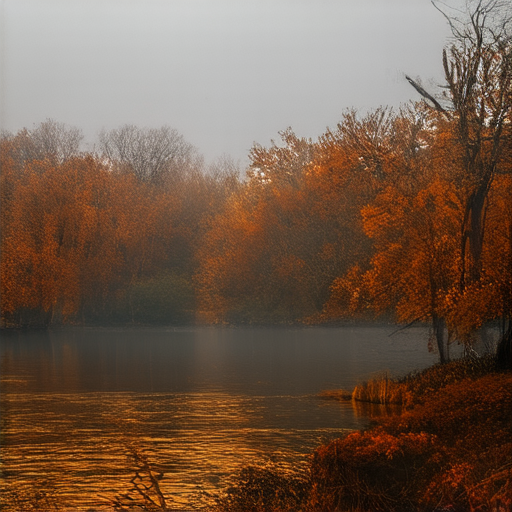} &
            \includegraphics[width=0.3\linewidth]{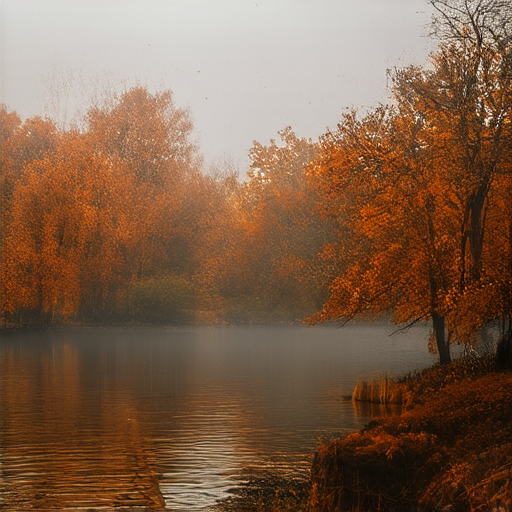} &
            \includegraphics[width=0.3\linewidth]{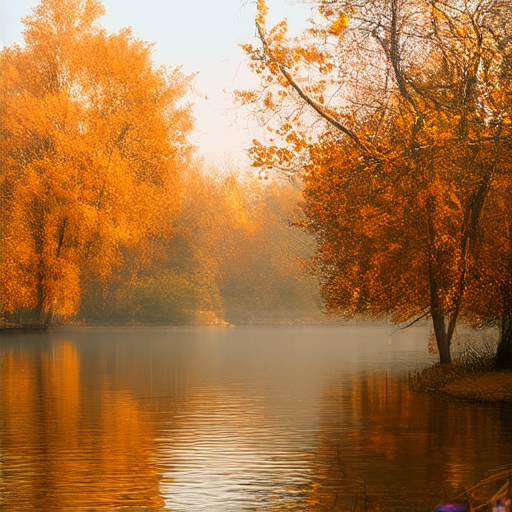} \\
            \includegraphics[width=0.3\linewidth]{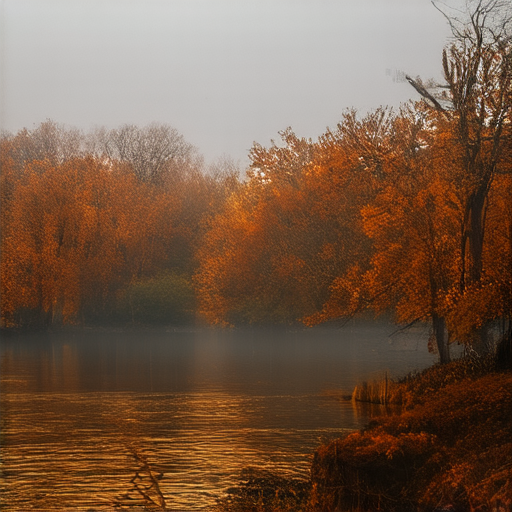} &
            \includegraphics[width=0.3\linewidth]{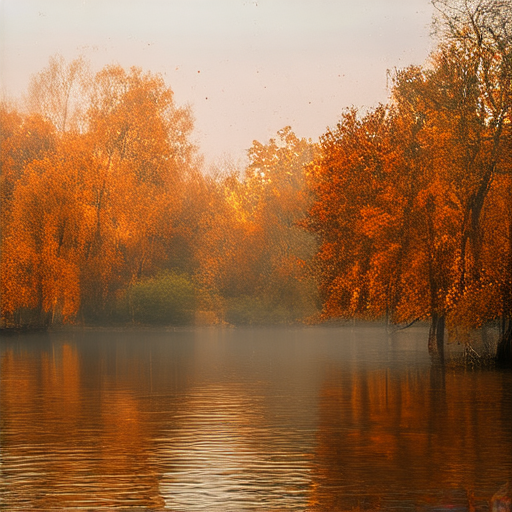} &
            \includegraphics[width=0.3\linewidth]{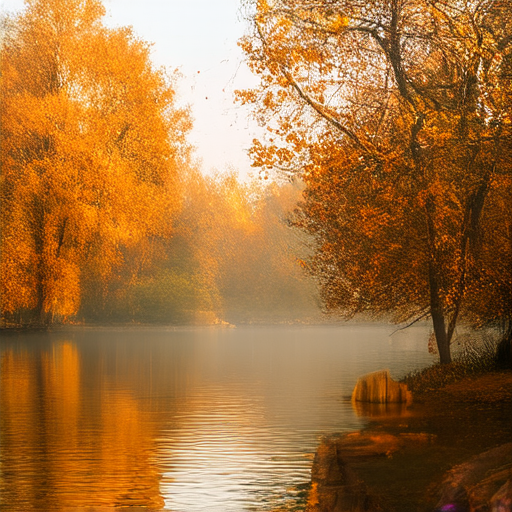} \\
        \end{tabular}
    \end{minipage}
    \caption{Emotion variation across the valence–arousal (VA) space for a landscape prompt ``A tranquil lake surrounded by golden autumn trees, misty morning.''
Left: the baseline (EmotiCrafter) exhibits severe emotion–semantic drift. 
Right: our anchor-constrained GRPO method maintains consistent scene structure (lake, autumn trees, mist) across all VA conditions, while expressing emotion through controlled changes in color, illumination, and atmosphere. 
}
    \label{fig:emotion_grid_landscape}
\end{figure*}

Figure~\ref{fig:qualitative_multi_prompt} presents a qualitative comparison across five diverse prompts, highlighting the differences in semantic preservation between the baseline and our method. For each prompt, the baseline model tends to alter the underlying scene structure when generating variations, often introducing unintended changes such as shifts in lighting, composition, or object layout. 
\begin{table*}[t]
\centering
\caption{Ablation study under different settings, including scale variations, GRPO-only, and anchor-only configurations on both SDXL and SD3.5M. Results are reported as mean $\pm$ std over 3,300 samples. Lower values are better for V-Err and A-Err, while higher values are better for CLIPScore and IQA. Bold indicates the best in each column.}
\label{tab:ablation}
\footnotesize
\setlength{\tabcolsep}{2.5pt}
\begin{tabular}{lcccccccc}
\toprule
\cellcolor{white} &
\multicolumn{4}{c}{\cellcolor{lightorange}\textbf{SDXL}} &
\multicolumn{4}{c}{\cellcolor{lightorange}\textbf{SD3.5M}} \\
\cmidrule(lr){2-5} \cmidrule(lr){6-9}
\cellcolor{white}\textbf{Setting} &
\cellcolor{lightblue}\textbf{V-Err $\downarrow$} &
\cellcolor{lightblue}\textbf{A-Err $\downarrow$} &
\cellcolor{lightblue}\textbf{CLIP $\uparrow$} &
\cellcolor{lightblue}\textbf{IQA $\uparrow$} &
\cellcolor{lightblue}\textbf{V-Err $\downarrow$} &
\cellcolor{lightblue}\textbf{A-Err $\downarrow$} &
\cellcolor{lightblue}\textbf{CLIP $\uparrow$} &
\cellcolor{lightblue}\textbf{IQA $\uparrow$} \\
\midrule
Scale=1.0 & \(1.901 \pm 1.299\) & \(1.879 \pm 1.104\) & \(\mathbf{30.556 \pm 2.830}\)& \(0.897 \pm 0.107\)
& \(1.896 \pm 1.315\) & \(1.888 \pm 1.111\) & \(30.265 \pm 2.662\) & \(0.891 \pm 0.081\) \\
Scale=1.5 & \(1.505 \pm 1.067\) & \(1.835 \pm 1.082\) & \(27.942 \pm 3.457\) & \(0.919 \pm 0.069\)
& \(1.609 \pm 1.117\) & \(1.846 \pm 1.073\) & \(26.822 \pm 3.554\) & \(0.895 \pm 0.074\) \\
GRPO only & \(\mathbf{1.505 \pm 0.304}\) & \(1.837 \pm 0.083\) & \(27.877 \pm 2.687\) & \(0.916 \pm 0.052\)
& \(\mathbf{1.097 \pm 0.784}\) & \(\mathbf{1.375 \pm 0.884}\) & \(28.648 \pm 3.647\) & \(0.816 \pm 0.165\) \\
Anchor only & \(1.538 \pm 0.914\) & \(1.846 \pm 0.083\) & \(28.312 \pm 3.187\) & \(\mathbf{0.974 \pm 0.102}\)& \(1.174 \pm 0.807\) & \(1.652 \pm 1.007\) & \(29.183 \pm 3.212\) & \(0.834 \pm 0.127\) \\
\bottomrule
\end{tabular}
\end{table*}
\begin{table*}[t]
\centering
\caption{Comparison of the proposed method  on backbones. Results are reported as mean $\pm$ std over 3,300 samples. }
\label{tab:backbone}
\setlength{\tabcolsep}{4pt}
\begin{tabular}{lcccc}
\toprule
\textbf{Method} &
\cellcolor{lightblue}\textbf{V-Err $\downarrow$} &
\cellcolor{lightblue}\textbf{A-Err $\downarrow$} &
\cellcolor{lightblue}\textbf{CLIPScore $\uparrow$} &
\cellcolor{lightblue}\textbf{IQA $\uparrow$} \\
\midrule
EmotiCrafter \cite{dang2025emoticrafter_iccv} & \(1.510 \pm 1.074\) & \(1.828 \pm 1.085\) & \(27.742 \pm 2.655\)& \(0.91 7\pm 0.069\)\\
SDXL All (GRPO+Anchor) & \(1.491 \pm 1.062\) & \(1.836 \pm 1.093\) & \(27.888 \pm 3.385\) & \(0.912 \pm 0.079\) \\
\rowcolor{lightgray} \textbf{Ours (SD3.5M + GRPO + Anchor)} & \(\mathbf{1.132 \pm 0.707}\)& \(\mathbf{1.492 \pm 1.080}\)& \(\mathbf{30.179 \pm 3.312}\) & \(0.861 \pm 0.137\) \\
\bottomrule
\end{tabular}
\end{table*}
For example, in the meadow and mountain scenes, the baseline outputs exhibit noticeable deviations from the original content, including changes in atmosphere and scene composition. Similarly, in the river and beach cases, structural elements such as terrain and water flow become less consistent. In the horses example, the baseline introduces additional variations in background and subject arrangement.

In contrast, the selected outputs from our method retain more of the prompt-level content while producing emotionally varied images. Scene layout, object identity, and spatial relationships are generally more stable than in the baseline examples, indicating an overall reduction rather than complete elimination of emotion--semantic drift. Figures~\ref{fig:emotion_grid} and~\ref{fig:emotion_grid_landscape} further illustrate this tendency across the valence-arousal space for a person-centric prompt and a landscape prompt, respectively.

\paragraph{Baseline Analysis.}
The baseline results (left grid) reveal a clear case of \textit{emotion-semantic drift}. When the target emotion is near neutral \(((0,0))\), the generated image largely follows the prompt, where a boy sits under a tree reading a book. However, as the valence or arousal deviates from zero, the generated content rapidly diverges from the intended semantics.
This degradation manifests in two primary forms. First, key semantic elements are gradually lost. For example, under positive valence conditions (e.g., \((3,-3)\)), the tree disappears, while under high arousal (e.g., \((0,3)\)), the comic book is replaced by a smartphone; in several cases, the boy is even replaced by a girl or an animal. Second, the model exhibits drastic scene-level changes under extreme emotional conditions. In the most extreme case \((3,3)\), the generated image contains a cat as the main subject, completely deviating from the original prompt.

\paragraph{Our Method Analysis.}
Compared with the baseline, our anchor-constrained GRPO method (right grid) more often retains the prompt's central subject and key objects across the VA conditions. The boy, tree, and comic book remain identifiable in most samples, although local details and the background may still vary. The grid therefore suggests that the proposed method reduces emotion--semantic drift overall, rather than guaranteeing exact semantic preservation at every VA coordinate.
Moreover, emotional variation is expressed mainly through attributes such as facial expression, color temperature, illumination, and atmosphere. This behavior supports smoother affective transitions with fewer large scene changes, while still allowing residual content variation and generation artifacts.

\subsection{Quantitative Results}

\begin{table*}[t]
\centering
\caption{Quantitative comparison with representative state-of-the-art baselines, including general text-to-image diffusion models, large multimodal models, and emotion-aware generation methods. 
}
\label{tab:main_results}
\setlength{\tabcolsep}{3.5pt}
\begin{tabular}{lcccc}
\toprule
\textbf{Method} & \textbf{V-Err} $\downarrow$ & \textbf{A-Err} $\downarrow$ & \textbf{CLIPScore} $\uparrow$ & \textbf{IQA} $\uparrow$ \\
\midrule
Cross Attention \cite{rombach2022highresolution}& $2.080 \pm 1.438$ & $1.923 \pm 1.153$ & $31.082 \pm 2.389$ & $\mathbf{0.942 \pm 0.066}$ \\
Time Embedding \cite{ho2020denoising} & $2.031 \pm 1.348$ & $1.941 \pm 1.168$ & $\mathbf{31.936 \pm 2.125}$ & $0.626 \pm 0.104$ \\
Textual Inversion \cite{gal2023textual} & $1.923 \pm 1.170$ & $1.958 \pm 1.188$ & $26.989 \pm 3.594$ & $0.323 \pm 0.131$ \\
GPT-4+SDXL \cite{openai2023gpt4,podell2023sdxl} & $1.517 \pm 1.060$ & $1.860 \pm 1.090$ & $31.242 \pm 2.705$ & $0.911 \pm 0.088$ \\
GPT-4+SD3.5M \cite{openai2023gpt4} & $1.622 \pm 1.179$ & $1.826 \pm 1.089$ & $31.211 \pm 2.565$ & $0.891 \pm 0.090$ \\
OmniGen-v2 \cite{wu2025omnigen} & $1.540 \pm 1.136$ & $1.845 \pm 1.096$ & $31.017 \pm 2.943$ & $0.790 \pm 0.177$ \\
FLUX.1-dev \cite{blackforest2024flux} & $1.602 \pm 1.158$ & $1.844 \pm 1.088$ & $30.809 \pm 2.564$ & $0.898 \pm 0.085$ \\
EmotiCrafter \cite{dang2025emoticrafter_iccv} & $1.510 \pm 1.074$ & $1.828 \pm 1.085$ & $27.742 \pm 2.655$ & $0.917 \pm 0.069$ \\
\midrule
SDXL All (GRPO+Anchor) & $1.491 \pm 1.062$ & $1.836 \pm 1.093$ & $27.888 \pm 3.385$ & $0.912 \pm 0.079$ \\
\textbf{Ours (SD3.5M + GRPO + Anchor)} & $\mathbf{1.132 \pm 0.707}$ & $\mathbf{1.492 \pm 1.080}$ & $30.179 \pm 3.312$ & $0.861 \pm 0.137$ \\
$\Delta$& \textcolor{ForestGreen}{$+ 25.0\%$} & \textcolor{ForestGreen}{$+ 18.4\%$} & \textcolor{ForestGreen}{$+ 8.78\%$} & \textcolor{red}{$- 6.11\%$} \\
\bottomrule
\end{tabular}
\end{table*}

Table~\ref{tab:main_results} compares the proposed method with general and emotion-conditioned generators. Our full method obtains the lowest emotional errors, with V-Err 1.132 and A-Err 1.492. Its CLIPScore of 30.179 improves on EmotiCrafter (27.742), although it remains below several general prompt-conditioning baselines. Its IQA score of 0.861 likewise indicates that the gain in emotion control is accompanied by a perceptual-quality trade-off rather than a uniform improvement on every metric.
Cross Attention, Time Embedding, and Textual Inversion produce V-Err values above 1.92, suggesting that feature-space injection alone provides only indirect supervision for final-image emotion. Prompt rewriting improves the errors but does not match the direct VA-reward optimization of the proposed method.

The SDXL All configuration jointly applies GRPO and the anchor on the SDXL backbone. It remains close to EmotiCrafter in VA error and CLIPScore, whereas replacing the backbone with SD3.5M yields a substantially larger improvement. The results should therefore be interpreted as evidence for the complete SD3.5M-based configuration, rather than attributing the full gain to GRPO alone.
The strongest evidence for the anchor mechanism comes from comparing our full method with the GRPO-only variant on SD3.5M, as shown in Table~\ref{tab:ablation}. GRPO-only achieves slightly lower VA errors (1.097/1.375) but sacrifices CLIPScore (28.648). In contrast, adding the anchor loss boosts CLIPScore to 30.179 (+1.531) with only a minor increase in VA error. This trade-off is precisely the intended behavior, where the anchor constraint prevents the model from exploiting semantic shortcuts (e.g., changing scene layout or objects) to game the VA reward, thereby preserving the prompt's original content while maintaining strong emotional expressiveness.

Moreover, relative to the SDXL-based counterpart in Table~\ref{tab:backbone}, the SD3.5M configuration reduces V-Err by 0.359 (24.1\%) and A-Err by 0.344 (18.7\%), while increasing CLIPScore by 2.291. IQA decreases from 0.912 to 0.861, so the comparison supports improved emotional alignment and prompt consistency but not an across-the-board quality gain.

\subsection{Ablation Study}

To systematically evaluate the contribution of each component, we conduct ablation experiments across two backbone architectures. Fig.~\ref{fig:ablation_all} summarizes the complete ablation, with panels~\ref{fig:ablation_group1}--\ref{fig:ablation_group3} reporting Valence/Arousal Errors in the upper plots and CLIPScore in the lower plots. Lower VA errors and higher CLIPScore indicate better performance.

On SD3.5M (Group~1), using only the anchor loss yields a CLIPScore of 29.183, compared with 27.742 for EmotiCrafter, but its VA errors remain 1.174/1.652. Using only GRPO produces the lowest VA errors (1.097/1.375) among the SD3.5M variants but a lower CLIPScore of 28.648. Their combination achieves the highest CLIPScore among the SD3.5M variants (30.179) while retaining low VA errors (1.132/1.492). The complementary pattern is consistent with the intended roles of the two terms: GRPO emphasizes emotional alignment, whereas the anchor penalizes semantic deviation.

All SDXL-based configurations, including Anchor only, GRPO only, and their combination (SDXL All), produce higher VA errors (V-Err $\ge 1.491$, A-Err $\ge 1.836$) and lower CLIPScore ($\le 28.312$) than our SD3.5M-based method. This demonstrates that the foundation model (SD3.5M) is critical for achieving both high emotional accuracy and content fidelity.
Our method reduces V-Err by 0.378 (25.0\%) and A-Err by 0.336 (18.4\%) compared to EmotiCrafter, while improving CLIPScore by 2.437(from 27.742 to 30.179). Compared to SDXL All (the best SDXL variant), our method improves V-Err by 0.359 (24.1\%), A-Err by 0.344 (18.7\%), and CLIPScore by 2.291. These gains provide the substantial improvement raised by the proposed emotional reward.

The anchor constraint restricts semantic variation, making VA optimization harder: on SD3.5M, adding the anchor increases V-Err by 0.035 and A-Err by 0.117 compared to GRPO-only, but raises CLIPScore by 1.531. This confirms reduced semantic shortcutting, though CLIPScore alone does not fully measure scene-structure preservation.

GRPO-only achieves the lowest VA errors (1.097/1.375), but its CLIPScore (28.648) is 1.531 below that of the full method. The combination of GRPO and the anchor therefore provides the best observed balance among the SD3.5M variants. The comparison with SDXL All further shows that the stronger SD3.5M backbone contributes materially to the reported VA-error and CLIPScore gains; this conclusion does not extend to IQA, on which SDXL All remains higher.

\subsection{User Study}
To complement the automatic metrics and directly assess human perception of emotional expressiveness and text–image alignment, we conduct a user study comparing the proposed method with EmotiCrafter~\cite{dang2025emoticrafter_iccv} and an ablation variant without the anchor constraint (Ours w/o Anchor).

\paragraph{Stimuli}
A set of 10 diverse text prompts is selected, covering natural scenes, human activities, and objects (e.g., ``a boy reading a comic book under a tree'', ``a sunny meadow'', ``a cat sleeping on a sofa''). For each prompt, images are generated under five target VA conditions: neutral (0,0), high valence (3,0), high arousal (0,3), high valence with high arousal (3,3), and low valence with low arousal (-3,-3). Three methods are evaluated: EmotiCrafter, Ours w/o Anchor, and Ours (Full), resulting in a total of \(10 \times 5 \times 3 = 150\) images.

\paragraph{Participants}
A total of 30 volunteers (18 female, 12 male; age 18–41) are recruited via a university mailing list. All participants have normal or corrected-to-normal vision and are fluent in English. Each participant evaluates a random subset of 50 images to reduce fatigue.

\paragraph{Evaluation Protocol}
For each image, participants are presented with the original text prompt and the corresponding target VA values (e.g., ``Valence = 3, Arousal = 3: very happy and excited''). They then rate the generated images along two independent dimensions using a 5-point Likert scale.
The first dimension measures emotional expressiveness, indicating how well the image reflects the target emotion (1 = not at all, 5 = perfectly). The second dimension evaluates text–image alignment, reflecting how well the image matches the content described in the prompt (1 = not at all, 5 = completely).

\paragraph{Results}
As summarized in Table~\ref{tab:user_study}, the proposed method receives higher mean ratings than both baselines on the two evaluation dimensions. Text--image alignment increases from 2.72 for EmotiCrafter to 4.45 for the proposed method, while emotional expressiveness increases from 3.67 to 4.32. These human ratings complement the model-based VA errors and support the intended emotion--semantic balance.

\begin{figure}[t]   
    \centering
    \includegraphics[width=0.8\linewidth]{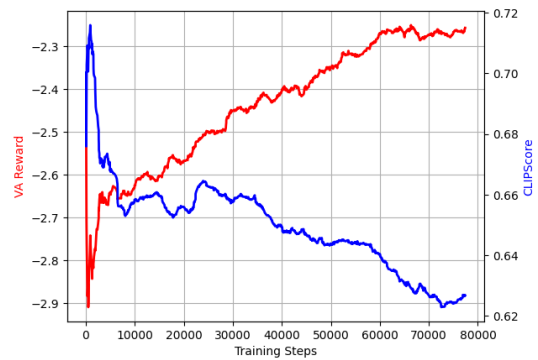}
    \caption{Training dynamics of GRPO on SD3.5M. The red curve (VA reward, defined as negative VA error) increases steadily throughout training, indicating progressively improved emotional alignment. In contrast, the blue curve (CLIPScore) exhibits a limited and mild decrease, reflecting the inherent trade-off between emotional expressiveness and semantic fidelity, where the anchor constraint effectively stabilizes semantic consistency during RL optimization.}
    \label{fig:training_curve}
\end{figure}
\begin{table}[htbp]
    \centering
    \footnotesize
    \setlength{\tabcolsep}{4pt}
    \caption{User study results (mean \(\pm\) std, N=30). Higher is better. Percentages below our variants indicate relative change compared to EmotiCrafter baseline, where $\Delta$ denotes the improvement compared with EmotiCrafter \cite{dang2025emoticrafter_iccv}. All differences between Ours (Full) and baselines are statistically significant (\(p < 0.05\) by paired t-test).}
    \label{tab:user_study}
    \scalebox{0.9}{
    \begin{tabular}{lcc}
        \toprule
        \textbf{Method} & \textbf{Emotional Expressiveness \(\uparrow\)} & \textbf{Text–Image Alignment \(\uparrow\)} \\
        \midrule
        EmotiCrafter~\cite{dang2025emoticrafter_iccv} & \(3.67 \pm 0.72\) & \(2.72 \pm 0.89\)\\
        \midrule
        Ours w/o Anchor & \(4.15 \pm 0.58\) & \(3.58 \pm 0.71\) \\
        $\Delta$ & \textcolor{ForestGreen}{+13.1\%} & \textcolor{ForestGreen}{+31.6\%} \\
        \textbf{Ours (Full)} & \(\mathbf{4.32 \pm 0.51}\) & \(\mathbf{4.45 \pm 0.48}\) \\
        $\Delta$ & \textcolor{ForestGreen}{+17.7\%} & \textcolor{ForestGreen}{+63.6\%} \\
        \bottomrule
    \end{tabular}}
\end{table}

The human evaluation results show higher mean ratings for both text--image alignment and emotional expressiveness in the evaluated samples. These results are consistent with an improved emotion--semantic balance, but they do not imply that the anchor prevents semantic drift in every generated image.

\subsection{Discussion in Emotional Image Generation}
A direct quantitative comparison with EmoGen~\cite{yang2024emogen}, EmoFeedback2~\cite{jia2025emofeedback2}, and EmoSENSE~\cite{emosense} is not available under the current protocol. EmoGen and EmoSENSE adopt categorical emotion interfaces that cannot be uniquely mapped to our continuous $5\times5$ VA grid, while EmoFeedback2 employs a different backbone and vision--language-model-based feedback procedure. However, A controlled cross-framework evaluation would require a common prompt set, matched backbone, and predefined category-to-VA mapping. We therefore limit our empirical claims to the methods evaluated in Tables~\ref{tab:main_results}--\ref{tab:backbone} and do not infer superiority over these systems from the Related Work comparison. 

Within our evaluation setting, Flow-GRPO reduces the continuous VA error, while the anchor improves CLIPScore over GRPO-only; the SD3.5M backbone further contributes to both improvements. These results suggest that emotional alignment and semantic preservation require complementary optimization objectives. In particular, the anchor helps mitigate the semantic degradation that may arise when optimizing primarily for emotional alignment.

Fig.~\ref{fig:training_curve} further illustrates this trade-off: the VA reward steadily increases during training, whereas CLIPScore shows a mild decline. This divergence indicates that stronger optimization of the emotional objective can gradually affect semantic consistency, supporting the need for explicit semantic constraints to achieve a better balance between emotional fidelity and semantic preservation.

\section{Conclusion}
In this paper, we presented an anchor-regularized Flow-GRPO framework for continuous emotion-conditioned image generation. The marginal-preserving ODE-to-SDE conversion provides stochastic transition densities for online policy optimization, the frozen VA regressor supplies a terminal emotion reward, and the zero-VA anchor penalizes semantic deviation in the CLIP visual space. On the evaluated benchmark, the full SD3.5M configuration obtains the lowest valence and arousal errors and improves CLIPScore over EmotiCrafter and the GRPO-only variant. The lower IQA score relative to several baselines shows that stronger emotional alignment does not automatically imply uniformly higher perceptual quality. Overall, the results support joint reward optimization and semantic regularization as a practical direction for continuous-affect image synthesis.



\bibliographystyle{IEEEtran}%

\thispagestyle{empty}

\section*{Author Biography}

\authorbibliography[scale=0.09,overhang=0pt,wraplines=9,imagewidth=3cm,imagepos=l]{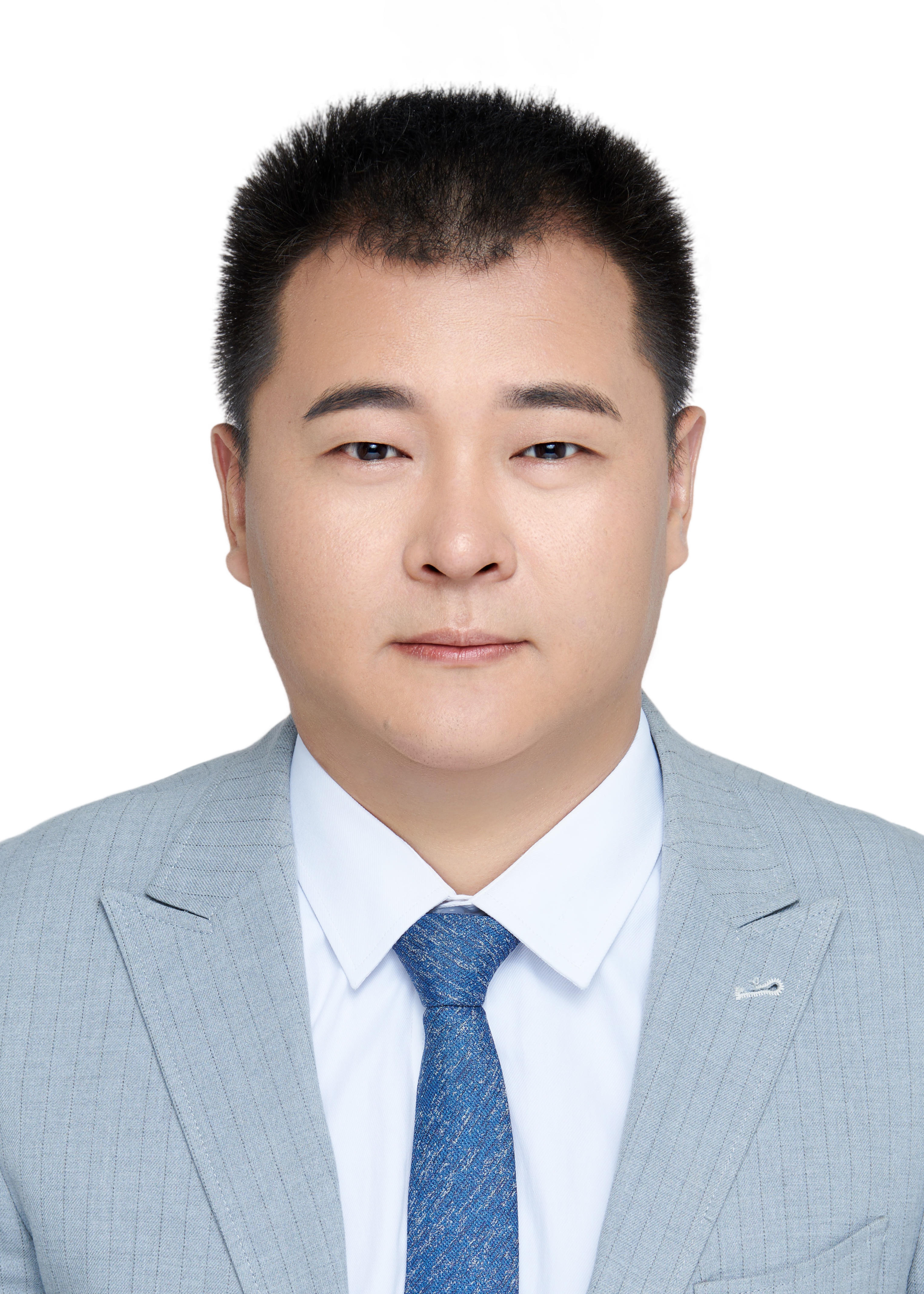}{Jisheng Dang \rm{
received the Ph.D. degree from Sun Yat-sen University, China, advised by Prof. Jianhuang Lai and Prof. Huicheng Zheng.
He worked as a research fellow at the NExT++ laboratory of the National University of Singapore, advised by Prof. Tat-Seng Chua. 
He is now a tenured associate professor at the School of Information Science and Engineering, Lanzhou University. 
His research interests include multimodal learning, video understanding, and embodied intelligence. 
He has published several papers as the first author or corresponding author in major journals and conferences including IEEE TPAMI/TIP/TNNLS/TITS/IJCAI/AAAI/ICLR/NeurIPS/PR.
}}

\authorbibliography[scale=0.115,overhang=0pt,wraplines=11,imagewidth=3cm,imagepos=l]{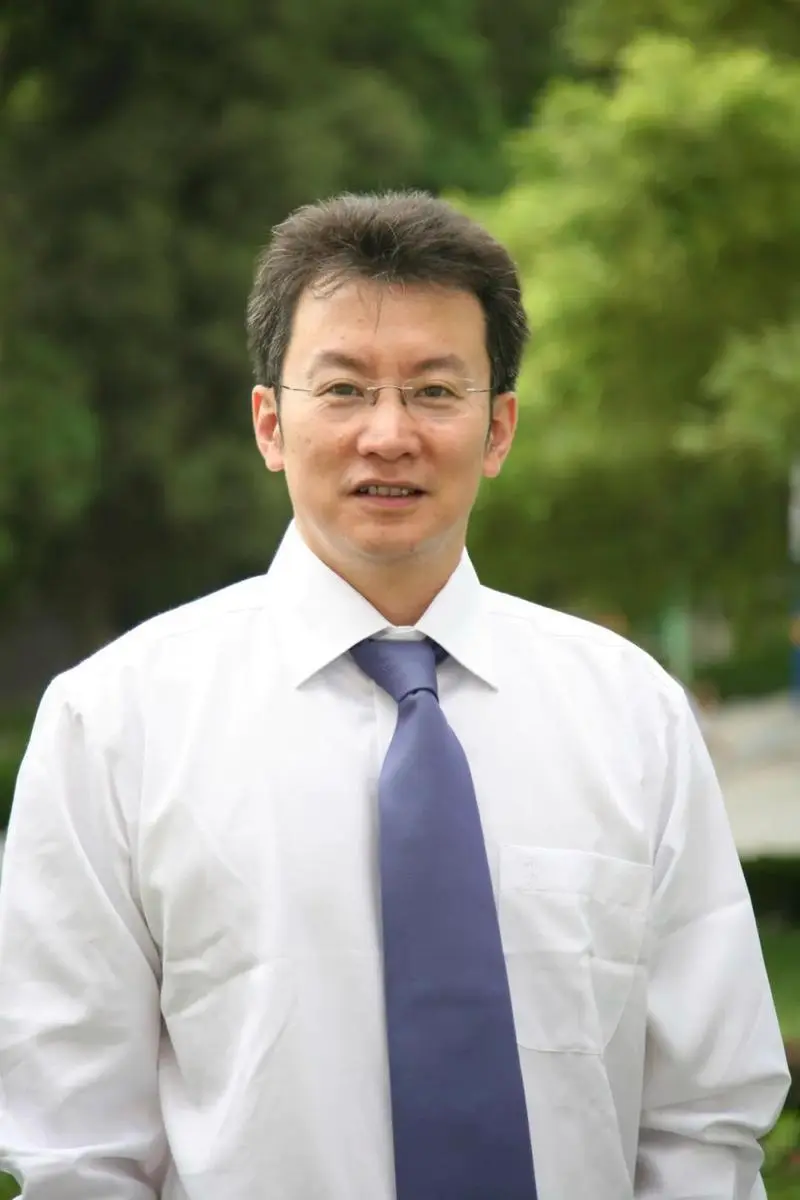}{Bin Hu (Fellow, IEEE) \rm{
received the Ph.D. degree in computer science from the Institute of Computing Technology, Chinese Academy of Sciences, China, in 1998. 
Since 2008, he has been a Professor and Dean of the School of Information Science and Engineering, Lanzhou University. 
He has also held a guest professorship at ETH Zurich. 
He serves as Editor-in-Chief of IEEE Transactions on Computational Social Systems and is a Fellow of IET and AAIA. 
His research interests include pervasive computing, computational psychophysiology, data modeling, and artificial intelligence.
}}

 
\authorbibliography[scale=0.12,overhang=0pt,wraplines=11,imagewidth=3cm,imagepos=l]{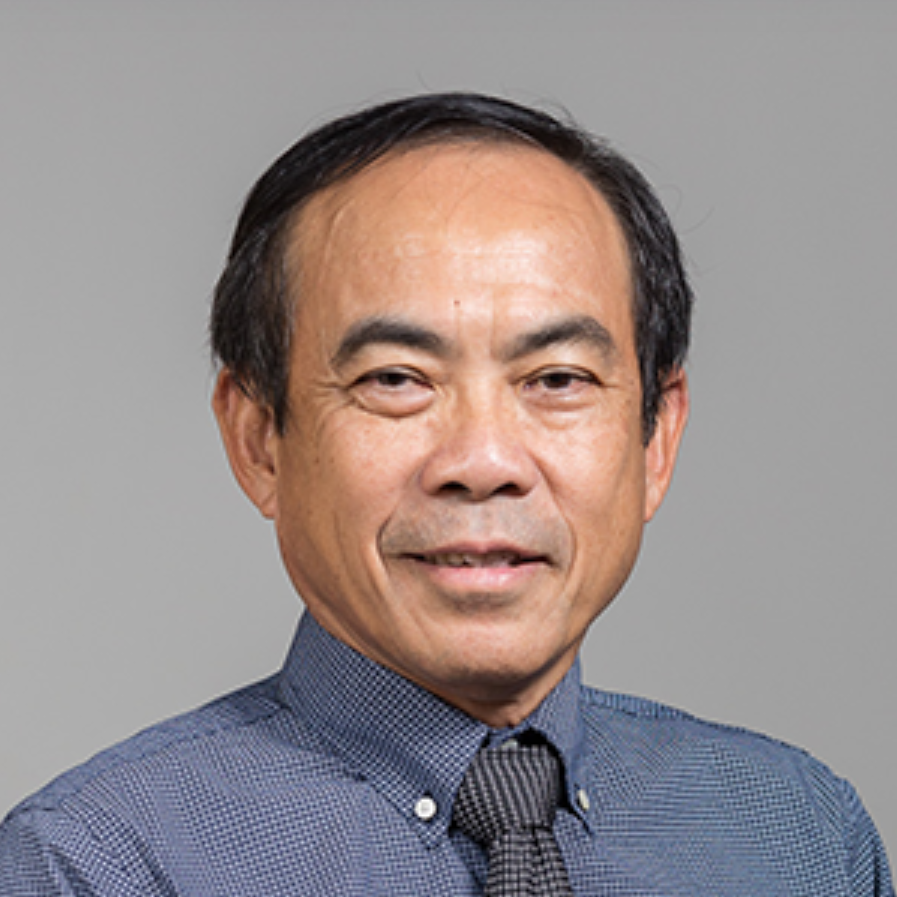}{Tat-Seng Chua \rm{received the Ph.D. degree from the University of Leeds, U.K. He is the KITHCT chair professor with the School of Computing, National University of Singapore, where he was the acting and founding dean of the School from 1998 to 2000. He is the co-director of NExT, a joint center between NUS and Tsinghua University, to develop technologies for live social media search. He is the 2015 winner of the prestigious ACM SIGMM Award. He is the chair of the Steering Committee of the ACM International Conference on Multimedia Retrieval (ICMR) and the Multimedia Modeling (MMM) conference series.   He is also the general co-chair of ACM Multimedia 2005, ACM CIVR (now ACM ICMR) 2005, ACM SIGIR 2008, and ACM Web Science 2015.   He serves on the editorial boards of four international journals. He is the co-founder of two technology startups in Singapore and a Fellow of the Singapore Academy of Sciences, with   103,316 citations on Google Scholar.}}

\vfill
\end{document}